%% file: main.tex
\documentclass[review,english]{elsarticle}

\usepackage[T1]{fontenc}
\usepackage[latin9]{inputenc}
\usepackage{comment}
\usepackage{geometry}
\usepackage{float}
\usepackage{url}
\usepackage{enumitem}
\usepackage{dcolumn}
\usepackage{calc}
\usepackage{amsmath}
\usepackage{amssymb}
\usepackage{epsfig}
\usepackage{booktabs}
\usepackage{multirow}
\usepackage{appendix}
\usepackage{array}
\usepackage[export]{adjustbox}

\usepackage{graphicx}
\graphicspath{{../figs/}}
\usepackage[caption=false,font=scriptsize]{subfig}

\makeatletter
\newenvironment{subfigures}
 {\begin{minipage}{\columnwidth}\def\@captype{figure}\centering}
 {\end{minipage}}
\makeatother

\usepackage{algorithm}
\usepackage{algpseudocode} %[noend]
\algnewcommand\algorithmicforeach{\textbf{for each}}
\algnewcommand{\LeftComment}[1]{\State \hspace*{-0.3em} \(\triangleright\) #1}
\algnewcommand{\LeftCommentx}[1]{\State \(\triangleright\) #1}
\algdef{S}[FOR]{ForEach}[1]{\algorithmicforeach\ #1\ \algorithmicdo}

\makeatletter
\renewcommand{\ALG@beginalgorithmic}{\scriptsize}
\algrenewcommand\alglinenumber[1]{\scriptsize #1:}
\makeatother

\algrenewcommand\textproc{}

\let\oldReturn\Return
\renewcommand{\Return}{\State\oldReturn}

\usepackage{esvect}

\newcolumntype{P}[1]{>{\centering\arraybackslash}p{#1}}
\floatstyle{ruled}
\newfloat{algorithm}{tbp}{loa}

\providecommand{\algorithmname}{Algorithm}
\floatname{algorithm}{\protect\algorithmname}

\usepackage{babel}

\usepackage{lineno}
\modulolinenumbers[5]

\addto\captionsenglish{\renewcommand{\appendixname}{Appendix}} %to fix a bug in elsearticle

\begin{document}

\title{Distributed Embodied Evolution over Networks}
%\titlerunning{Distributed Embodied Evolution in Networks of Agents}

%\corref{}

\author{Anil Yaman}
\ead{anilyaman@kaist.ac.kr}
\address{Department of Bio and Brain Engineering\\
Korea Advanced Institute of Science and Technology\\
Daejeon, 34141, Republic of Korea}

\author{Giovanni Iacca}
\ead{giovanni.iacca@unitn.it}
\address{Department of Information Engineering and Computer Science\\
University of Trento\\
Via Sommarive 9, 38123 Povo (Trento), Italy}

%\orcidID{0000-0003-1379-3778} %Anil
%\orcidID{0000-0001-9723-1830} %Giovanni
%\authorrunning{A. Yaman and G. Iacca}

%%%%%%%%%%%%%%%%%%%%%%%

\begin{abstract}
In several network problems the optimum behavior of the agents (i.e., the nodes of the network) is not known before deployment. Furthermore, the agents might be required to adapt, i.e. change their behavior based on the environment conditions. In these scenarios, offline optimization is usually costly and inefficient, while online methods might be more suitable. In this work, we use a distributed Embodied Evolution approach to optimize spatially distributed, locally interacting agents by allowing them to exchange their behavior parameters and learn from each other to adapt to a certain task within a given environment. Our results on several test scenarios show that the local exchange of information, performed by means of crossover of behavior parameters with neighbors, allows the network to conduct the optimization process more efficiently than the cases where local interactions are not allowed, even when there are large differences on the optimal behavior parameters within each agent's neighborhood.
\end{abstract}
\maketitle

%%%%%%%%%%%%%%%%%%%%%%%

\input{1_introduction}
\input{2_methods}
\input{3_experimentalSetup}
\input{4_experimentalResults}
\input{5_conclusions}

%\section{*Acknowledgements}

%%%%%%%%%%%%%%%%%%%%%%%%%%%%%%%%%%%%%%%%%%%%%%%%%%%%%%%%%%%%

\clearpage

\bibliographystyle{elsarticle-num}

\clearpage

\appendix
%\input{6_appendix_imitation}
%\clearpage
\input{6_appendix_stats}

\end{document}

%% file: 1_introduction.tex
\section{Introduction}
\label{sec:intro}

Networks of agents, such as sensor networks, wireless networks, swarms of drones or terrestrial robots, etc. are used nowadays in many tasks in the context of environment exploration, monitoring, and Internet of Things (IoT) applications. Typically, the behavior of the network as a whole derives from the \emph{local behavior} of each single agent. In this work, we consider a \emph{local behavior} any form of agent-local decision based on the environment, be it an action depending on external stimuli (in the case of agents with actuation capabilities), or the update of a data-driven model used e.g. for classification or forecast tasks. Modelling the possible mutual interactions of such agent-local behaviors (which can be very different across different parts of the network), as well as the interactions of the agents with the environment, can be difficult. In fact, the environment conditions might not be known \emph{a priori}, i.e. before deployment, and may change dynamically and unexpectedly. As such, finding the optimal behavior of network agents offline, analytically, can be challenging: on the one hand, an analytical formulation is rarely available; on the other, collecting all the agents to tune their parameters when needed can be inefficient -and expensive- especially when the number of agents is large. Moreover, in some cases the agents might not even be accessible for collection or data extraction, e.g. if their position is not known or they are placed in hard-to-access environments such as underground tunnels or pipes~\citep{hallawa2020morphological}. 

For all these reasons, a large body of literature has investigated various concepts such as \emph{autonomic}, or \emph{self-adaptive} networking~\citep{bouabene2009autonomic,xiao2016bio}, in the quest for solutions to make networks capable of adapting automatically to the environment and thus optimizing autonomously, at runtime, their behavior. In this area, bio-inspired techniques -especially distributed Evolutionary Algorithms (EAs)- have attracted a great attention~\citep{nakano2010biologically,dressler2010survey} due to their intrinsic adaptivity. It is worth mentioning though that the idea of distributed EAs was initially introduced mainly in the context of numerical optimization, in the form of structured populations of solutions evolved in parallel over multiple cores or over computing networks~\citep{alba2002parallelism,arenas2002framework,maqbool2019scalable}. More recently, distributed versions of Differential Evolution have been presented for instance in~\citep{4135297,4135309}, and \citep{biazzini2013p2poem}, among which the latter work introduced a distributed optimization framework over P2P networks. A special flavour of distributed EAs is Cellular Evolution (CE), originally devised for Alife studies with explicit Cellular Automata (CA) models~\citep{sipper1994studying}, and later applied e.g. to evolvable hardware~\citep{sipper1997evolution} and other numerical optimization problems~\citep{alba2000cellular}, where a population of individuals is restricted to interact (i.e., mate) locally. This can help preserve the population diversity longer. Of note, in CE all individuals optimize the same fitness function.

As an alternative paradigm to using a network as a collection of computing nodes for solving an optimization problem, more recently distributed EAs have been used also in physical or simulated networks to evolve agent-local parameters. One field of application of this concept is Wireless Sensor Networks (WSNs), for which previous works have proposed e.g. CA~\citep{wang2012cellular,choudhury2012cellular}, distributed EAs~\citep{iacca2012introducing,iacca2013distributed} and distributed Genetic Programming (GP)~\citep{johnson2005genetic,valencia2010distributed} to evolve the sensor nodes' parameters and functioning logic. However, robotics is the area where distributed EAs have shown their greatest potential so far. In this context, Embodied Evolution (EE), i.e. the idea of running an EA distributedly on a group of agents, has been successfully applied to optimize the behavior of robotic swarms in various works, see e.g. ~\citep{watson2002embodied,4488332,eiben2010embodied,Bredeche2018}. In particular, \emph{environment-driven} EE, a form of artificial ecology where the environment conditions guide the evolutionary process of a collection of agents, has emerged as a promising way to obtain truly self-adaptive swarms~\citep{bredeche2012environment}. As such, this research area has been very active in recent years, advancing not only on the algorithmic aspects~\citep{bredeche2012environment,haasdijk2014combining,montanier2016behavioral,hart2015improving,zahadat2015evolving,perez2019influence}, but also on the application side: for instance, an interesting application on indoor surveillance and location has been proposed in~\citep{trueba2015embodied,prieto2016real}.

\begin{figure}[ht!]
\centering
  \includegraphics[width=0.8\columnwidth,clip,trim=0 0 0 0.5cm]{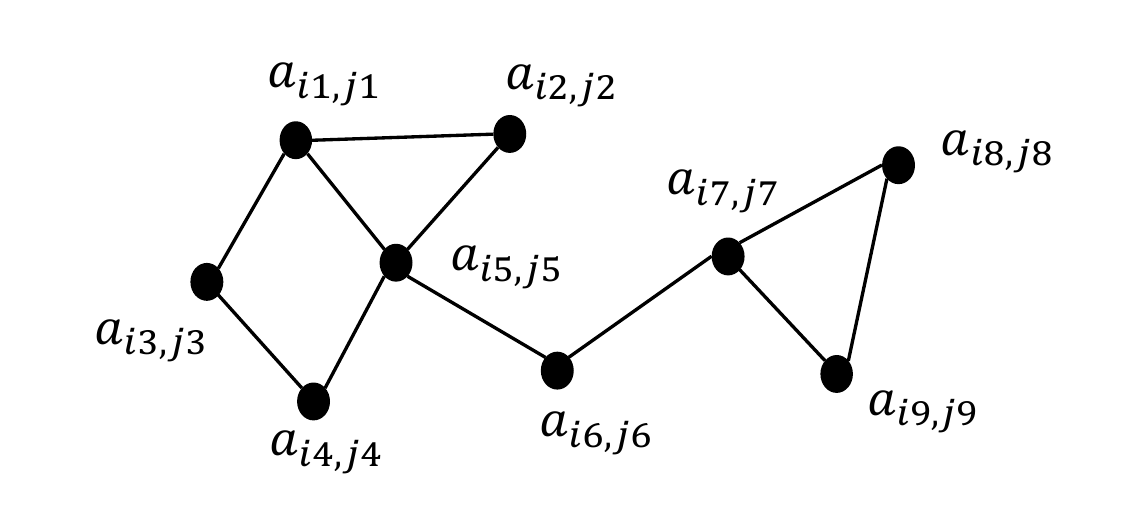}
  \caption{An illustration of a network of spatially distributed, locally connected agents.}
  \label{fig:networkIllustration}
\end{figure}

In this work, we continue this research stream by investigating the effect of \emph{local information exchange} on distributed Embodied Evolution. However, differently from most of the existing literature on EE focusing on robotics, here we focus specifically on networks. We consider networks of agents distributed at fixed locations within a certain environment, see Figure~\ref{fig:networkIllustration} for an example. We assume that each agent is required to perform a certain local behavior, that is described by some agent-local parameters, and that in general the optimal local behaviors are different across the agents as they depend on the local environmental conditions. Nevertheless, in some cases the behaviors of the agents in close proximity can be similar, since they may share similar environmental conditions. We also assume that the parameters for the optimal local behaviors are not known at the time of deployment. Therefore, in our setup the agents are required to optimize their local parameters autonomously while they operate within the environment. To achieve that, we use a distributed embodied EE, allowing each agent to run a population-less EA \emph{in situ}, to optimize its own behavior. In contrast to CE, we assume that the fitness function of each agent can be different.

We introduce local information exchange in terms of exchange of behavior parameters with local neighbors: we consider various kinds of parameter exchange, such as 1) copying the parameters from the best or a random agent in the neighborhood, with some mutation, or 2) exchange the parameters partially, i.e. by means of crossover of parameters with neighbors. We hypothesize that using parameter exchange as crossover and local perturbations for mutation leads to a social learning process through which it is possible to adapt a network of agents to a task at runtime, and in a distributed fashion. While it may be intuitive to exchange parameters with neighbors when their local conditions are similar, we are also interested in investigating the effect of parameter exchange in the cases where the optimal behaviors of the neighboring agents are different. Therefore, our main research questions are:
\begin{enumerate}
    \item What are the conditions under which it is more appropriate to exchange local parameters across nodes in an EE setting?
    \item What are the most appropriate parameter exchange mechanisms, and parameters thereof?
\end{enumerate}

To answer these questions, we devise three test problems --of which one is a real-world IoT application taken from the literature-- with different network sizes and different levels of similarities in terms of optimal behavior within a neighborhood. Additionally, these three problems show how the generic definition of behavior introduced above encompasses different kinds of applications, ranging from distributed actuation (in a synthetic illumination task) to a distributed data-driven model of human presence and activity (in a realistic IoT scenario).

Overall, our results show that local information exchange is beneficial in any case -even with large differences of the optimal parameters within a neighborhood- while the optimization performance can be largely affected by the frequency of the parameter exchange (i.e., the crossover probability) and the mutation rate. To summarize, our main contributions are the following:
\begin{itemize}[leftmargin=*]
    \item We apply EE to networks, with a particular emphasis on IoT systems such as sensor networks and alike: we believe that this represents a novel (and promising) application area of EE, which so far has been applied almost exclusively to swarm robotics.
    \item We specifically include in our experiments cases where the local fitness functions are different. This is especially important because sharing local parameters may not be necessarily advantageous in those cases. To the best of our knowledge, this aspect is not well-studied in previous works on EE in robotics, where usually the fitness function is assumed to be the same for all robots (e.g. their energy level, such as in~\citep{watson2002embodied}), but also in most of the literature on CE, where little or no attention has been given to possible differences on the local fitness functions.
    \item We perform extensive tests using several different variants of local information exchange strategies, also studying their parametrization, and assess their effect on the evolutionary process with a thorough statistical analysis (see the Appendix). We find that this in-depth analysis of the parameters' effect on EE schemes is also unprecedented in the previous related literature.
\end{itemize}

The remaining of this paper is structured as follows. In the next section, we introduce the general scheme for distributed Embodied Evolution over networks. Then, Section~\ref{sec:setup} describes the experimental setup, while Section~\ref{sec:results} presents the numerical results. Finally, in Section~\ref{sec:conclusions} we give the conclusions and highlight possible future research directions.

%% file: 2_methods.tex
\section{Methods}
\label{sec:methods}

Without loss of generality, we consider a system composed of a collection of $k$ agents, spatially distributed on a 2-dimensional plane at fixed locations indexed by ${i,j}$, with $i$ and $j$ $\in \mathbb{N}_0$. Each agent $a_{i,j}$ must show a behavior such that it optimizes a certain local function. The performance of each agent is measured by its local fitness value $f_{i,j}$, that measures how good is the agent's behavior w.r.t. the desired local function. Furthermore, each agent can communicate (bi-directionally) with its local neighbors defined by a neighborhood function $N$. An example of such a system is shown in Figure~\ref{fig:networkIllustration}: for instance, the neighborhood function of $a_{i5,j5}$ is defined by $N=\{(i1,j1),(i2,j2),(i3,j3),(i4,j4),(i6,j6)\}$, i.e. the set of indices of the agents that are connected to $a_{i5,j5}$.

We employ an Embodied Evolution approach~\citep{Bredeche2018} where each agent runs a (population-less) evolutionary algorithm to optimize its own behavior parameters, which represent the agent's genotype. This can be achieved with or without information sharing (i.e. sharing all or some of the elements of the genotypes) with the neighboring agents. In the case where there is no information sharing, each agent can iteratively test a randomly perturbed version of its genotype, and store the genotype that performs the best (in the following, we will refer to this approach as HillClimbing). In the case where information sharing is allowed, the agents can copy and/or perform crossover with the genotypes of their neighbors. As discussed earlier, the optimal behavior of each agent can be potentially different from that of its neighbors, and studying the possible advantages/disadvantages of sharing information in this condition is one of the main focuses of this work.

We define two helper functions, $best(\boldsymbol{X}^{i,j})$ and $rand(\boldsymbol{X}^{i,j})$, that return respectively the genotype of the best and the genotype of a (uniformly selected) random agent in the neighborhood of a given agent $\boldsymbol{X}^{i,j}$. In real-world scenarios, this may be implemented by allowing each agent to communicate its fitness and genotype with its neighbors. In this case, and assuming that the local fitness values are comparable, it would be possible for each agent to pick the best genotype, or a random genotype, to perform evolutionary operators.
%NOTE: this assumption is quite important actually!

%, $\boldsymbol{X}^{\prime}\gets xover(\boldsymbol{X}^{i,j},\boldsymbol{X}^{k,l}, cp, cr)$, 
Finally, we use two evolutionary operators, namely \textit{uniform crossover} and \textit{Gaussian mutation}, applied in this order. The crossover operator generates a new genotype $\boldsymbol{X}^{\prime}$ from two given parent genotypes, $\boldsymbol{X}^{i,j}$ and $\boldsymbol{X}^{k,l}$, as follows: first, $\boldsymbol{X}^{k,l}$ is copied into $\boldsymbol{X}^{\prime}$; then, each element of $\boldsymbol{X}^{i,j}$ is copied into the corresponding element of $\boldsymbol{X}^{\prime}$ with a crossover rate $cr \in [0,1]$. The crossover probability $cp \in [0,1]$ specifies the probability of performing the crossover operator. If the genotype of the agents consists of a single parameter, then we use \textit{arithmetic crossover}, that computes the mean of the parents' parameters, i.e. $x^{\prime} = (x^{i,j} + x^{k,l})/2$ (note that in this case $cr$ is not needed). After crossover, the mutation operator perturbs each element of $\boldsymbol{X}^{\prime}$ using a Gaussian mutation (sampled independently for each element) $\mathcal{N}(0,\sigma)$ with zero mean and standard deviation $\sigma$ (in the following, we will refer to $\sigma$ as the mutation rate, $mr$).

For illustration purposes, we provide in Algorithm~\ref{alg:embodiedNetworkEvolution} a pseudo-code of a version of the algorithm in which we apply crossover with a random neighbor. We dub this algorithm as XoverRand. The procedure randomInitialize() indicates a random initialization of the initial behavior parameters $\boldsymbol{X}^{i,j}$, where each element is uniformly sampled within its range $[lb,ub]$, being $lb$ and $ub$ the lower and upper bound respectively, while $g$ and $g_{max}$ indicate, respectively, the current and the maximum number of generations. 
\begin{algorithm}[ht!]
    \begin{algorithmic}[1]
	    \Procedure{Evolve}{}
            \State $\boldsymbol{X}^{i,j} \gets \text{randomInitialize}()$
            \State $f_{i,j} \gets \text{eval}(\boldsymbol{X}^{i,j})$
            \State $g \gets 0$
            \While {$g < g_{max}$}
                \State $\boldsymbol{X}^{\prime} \gets \text{xover}(\boldsymbol{X}^{i,j}, \text{rand}(\boldsymbol{X}^{i,j}), cp, cr)$ \label{ln:xover} \Comment{Crossover with random neighbor}
                \State $\boldsymbol{X}^{\prime} \gets \boldsymbol{X}^{\prime} + \mathcal{N}(0,\sigma)$\label{ln:mutation} \Comment{Gaussian mutation}
                \State $f^{\prime} \gets \text{eval}(\boldsymbol{X}^{\prime})$
                %\Comment{Function evaluation}
                \If {$f^{\prime} < f_{i,j}$}\label{ln:comparison}
                \Comment{Update (assuming minimization)}
                    \State $f_{i,j} \gets f^{\prime}$
                    \State $\boldsymbol{X}^{i,j}\gets \boldsymbol{X}^{\prime}$
                \EndIf
                \State $g \gets g+1$
            \EndWhile \label{ln:endWhile}
		\EndProcedure
	\end{algorithmic}
\caption{Embodied evolution of an agent $a_{i,j}$ (and its corresponding behavior $\boldsymbol{X}^{i,j}$) with crossover applied with a random neighbor (XoverRand).}
\label{alg:embodiedNetworkEvolution}
\end{algorithm}

%% file: 3_experimentalSetup.tex
\section{Experimental setup}
\label{sec:setup}

In this section, we provide the details of our experimental setup. We are mainly interested in investigating the performance of Embodied Evolution with and without exchanging information, i.e. exchanging each agent's genotype (behavior parameters) with its neighbors, as well as the effect of the frequency of information sharing. Therefore, we test different versions of the Embodied Evolution algorithm introduced in the previous section, including variants in which it is allowed to exchange information across agents, and variants in which this exchange does not occur.

In the following, we first define the three test problems we used in our experimentation (and, for each problem, the relative scenarios). Then, we describe the five versions of the algorithm we tested on these problems.

In the first two problems (dubbed as ``imitation'' and ``illumination'', respectively), we use a 2-dimensional environment represented as an $m \times n$ grid. Each cell ${i,j}$ on this grid is occupied by an agent $a_{i,j}$. Of note, in this setup the grid topology makes the EE model similar to Cellular Evolution, although as discussed earlier differently from CE here we also consider cases where the local fitness functions are different. We use the following neighborhood function: $N_{Moore} = \lbrace (i-1,j-1), (i,j-1), (i+1,j-1), (i-1,j), (i+1,j), (i-1,j+1), (i,j+1), (i+1,j+1) \rbrace$. This function, known as the Moore Neighborhood~\citep{gray2003mathematician}, allows each agent to communicate with its nearest horizontal, vertical and diagonal neighbors. The agents that are located at the border of the environment can communicate only with their existing neighbors.

In the third problem, concerning a distributed model of indoor human presence and activity, we instead use three different irregular (i.e., not grid-shaped) network topologies, shown in Figure~\ref{fig:roomClimate}, where each node in the network corresponds to an agent and the neighborhood function is the 1-hop neighborhood according to the given topology.

\begin{figure}[ht!]
\begin{subfigures}
\subfloat[$t=1$]{\includegraphics[width=0.32\columnwidth]{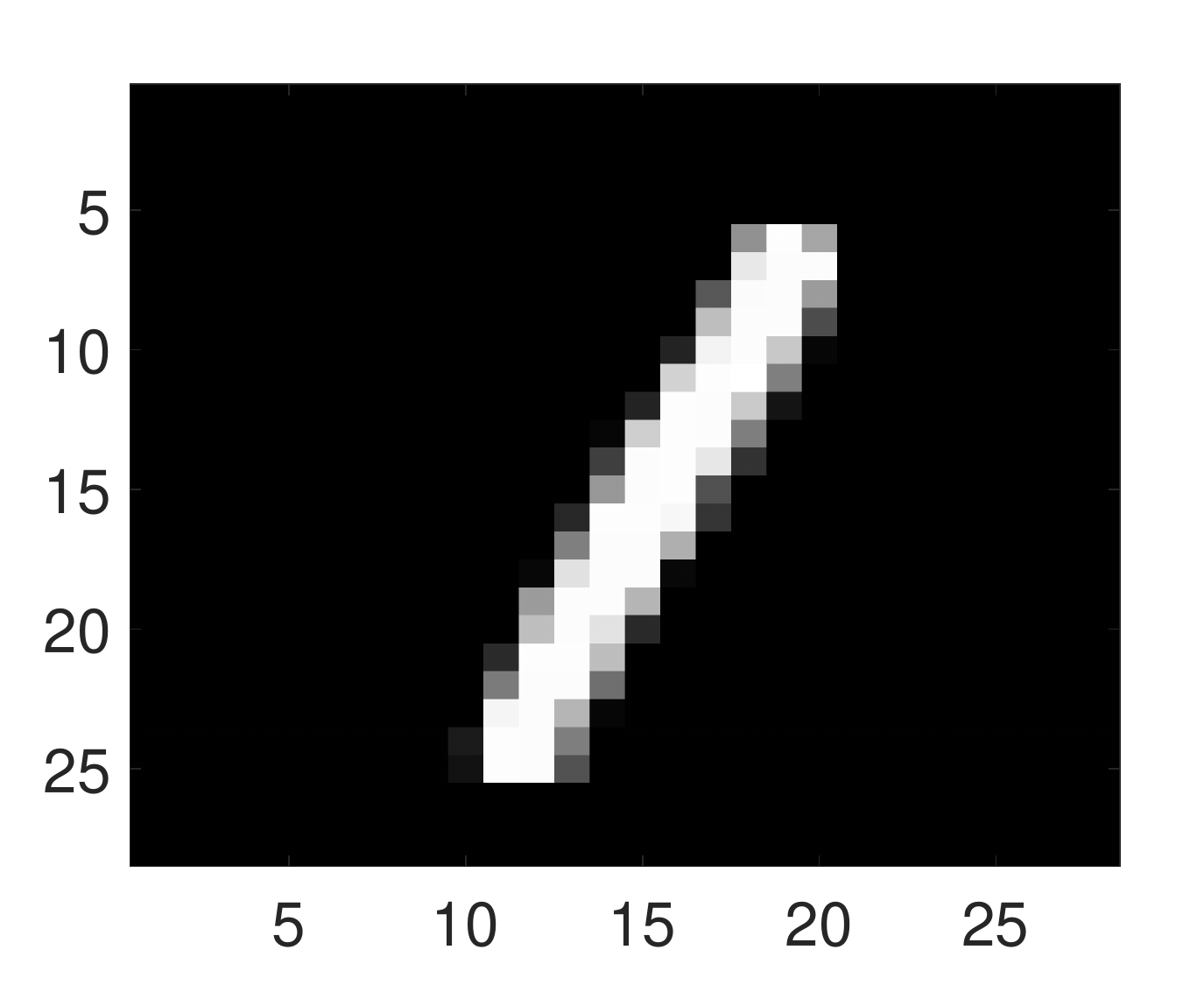}\label{fig:MNISTGroundTruth1}}
\subfloat[$t=50$]{\includegraphics[width=0.32\columnwidth]{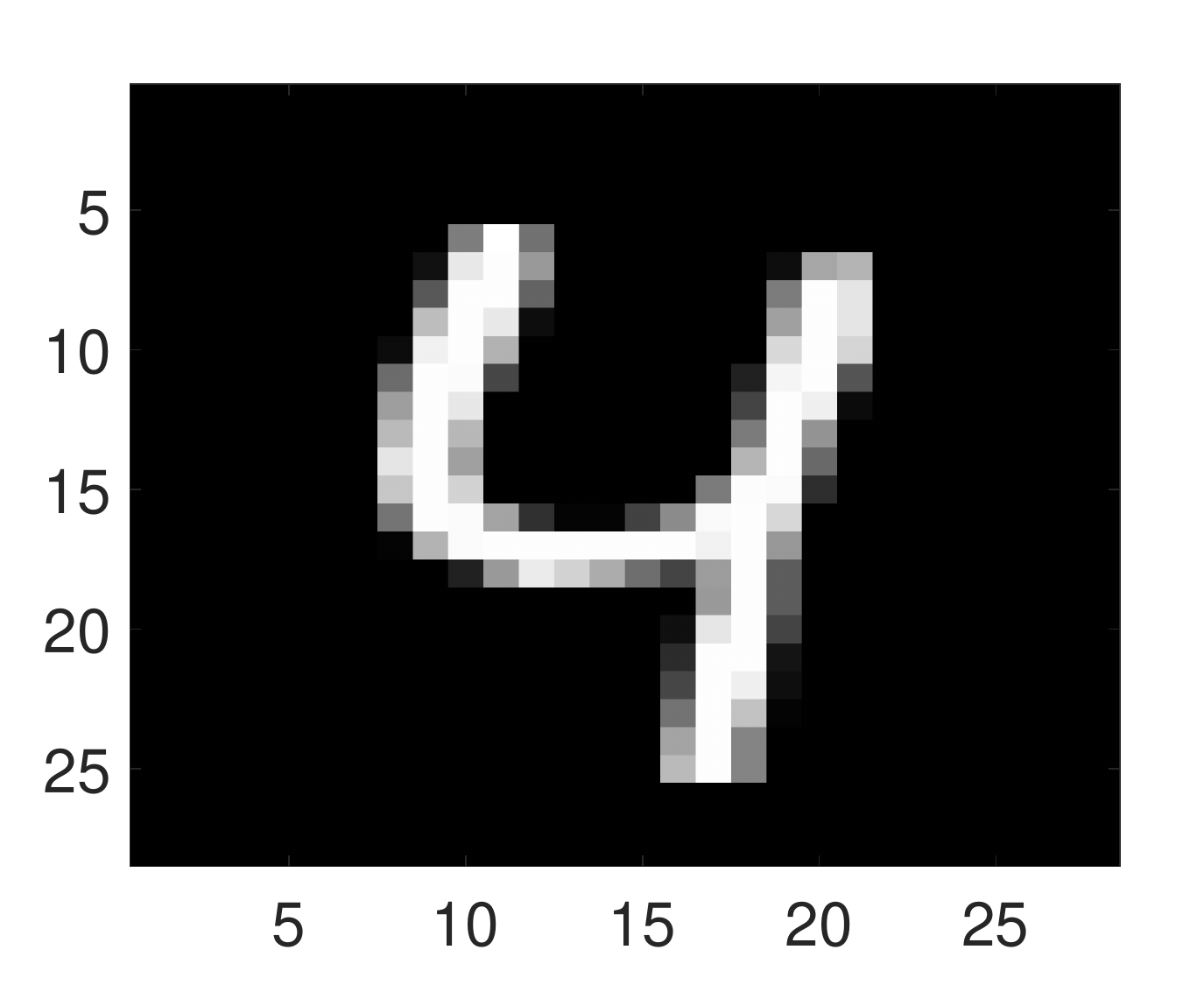}\label{fig:MNISTGroundTruth50}}
\subfloat[$t=100$]{\includegraphics[width=0.32\columnwidth]{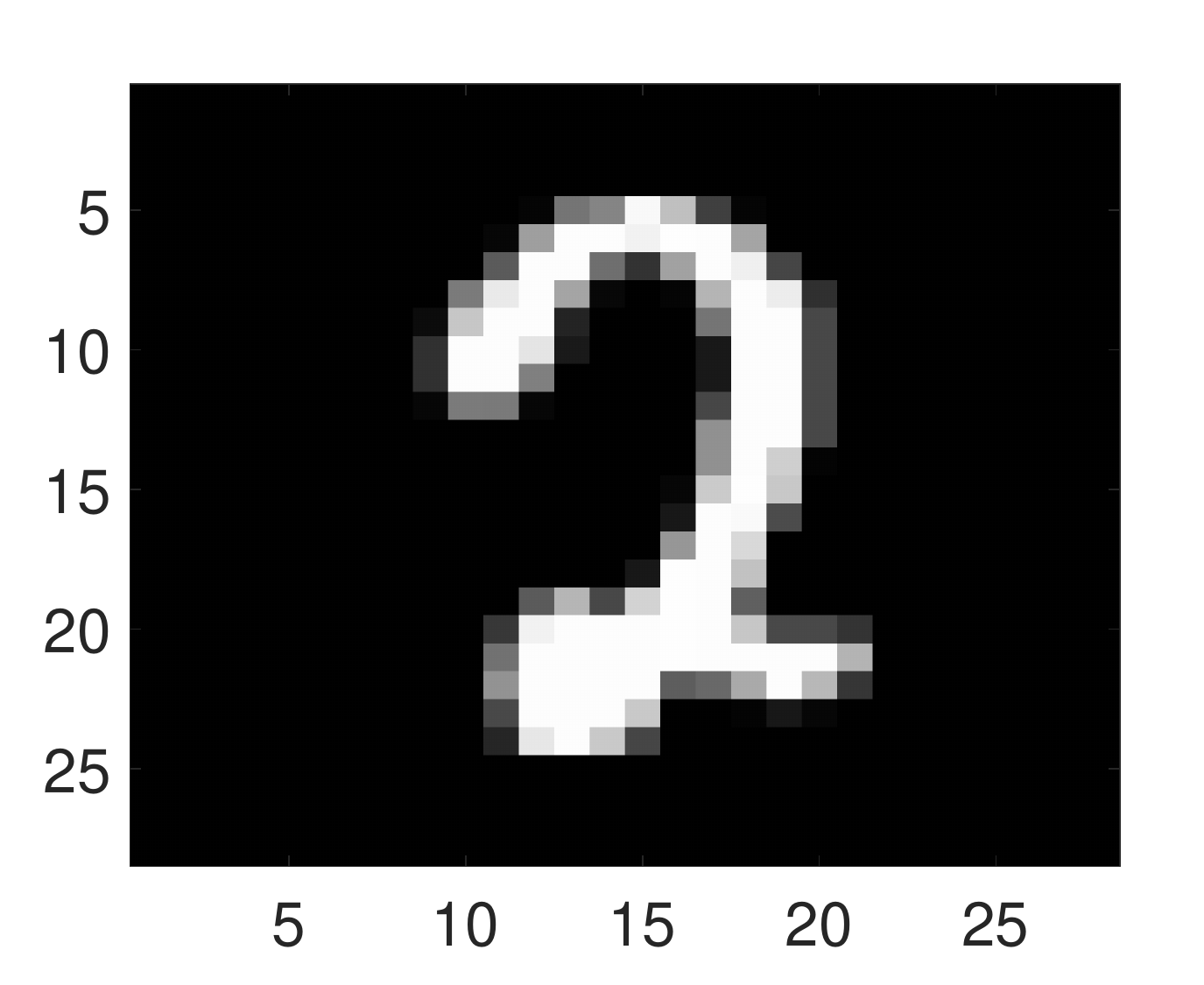}\label{fig:MNISTGroundTruth100}}
\end{subfigures}
\caption{Imitation problem: Visualization of three images (ground truth) at $t=1$, $t=50$ and $t=100$.} \label{fig:MNISTGroundTruth}
\end{figure}

\subsection{Imitation problem}
We define as \textit{``imitation problem''} a generic problem in which each agent in a network is required to ``imitate`` a desired pattern (the ground truth) over time. We demonstrate this by optimizing a network of agents to imitate a sequence of $100$ images taken from the well-known MNIST handwritten digits dataset~\citep{lecun1998gradient}. For this problem, we devise two scenarios, differing for the number of agents in the network and the number of parameters describing each agent's behavior.

\subsubsection{$28 \times 28$ scenario}
In this scenario, the network consists of $28\times28$ agents, each one corresponding to one pixel of the MNIST images. The grayscale intensity in each cell is controlled by agent $a_{i,j}$ at time $t \in \{1, 2, \hdots, 100\}$, where $t$ corresponds to the $t$-th image. Thus, the genotype of $a_{i,j}$ is $\boldsymbol{X}^{i,j} = (x^{i,j}_1, x^{i,j}_2, \hdots, x^{i,j}_{100}) \in [0,1]^{100}$, where each $t$-th element determines the grayscale intensity (the phenotype) corresponding to the agent's cell at time $t$. A visualization of some selected images (ground truth) at indices $t=1$, $t=50$ and $t=100$ is shown in Figure~\ref{fig:MNISTGroundTruth}.

In the initialization phase, the genotype of each agent is generated by randomly sampling its genes in $[0,1]$, with uniform probability. We use then Embodied Evolution to find the optimal parameters of each agent that can collectively ``imitate'' the sequence of $100$ images as close as possible. Thus, in total, there are $7.84\times10^4$ parameters to be optimized in the network ($28\times28$ agents $\times 100$ parameters). The fitness value of each agent is calculated as follows:
\begin{equation}
f_{i,j} = \frac{1}{100} \sum^{100}_{t=1} \mid I_{i,j}(t) - x^{i,j}_t \mid 
\label{eq:imitationFitnessCalc}
\end{equation}
\noindent{}where $I_{i,j}(t)$ denotes the desired ${i,j}$ pixel value of the $t$-th image selected from the MNIST dataset (pixel values are scaled in the range $[0,1]$), and $x^{i,j}_t$ represents the corresponding $t$-th element of the agent's genotype $\boldsymbol{X}^{i,j}$. The \emph{collective fitness} of the network at generation $g$, $F_g$, is then computed as the average fitness across the agents ($F_g = \sum_{i}\sum_{j} f_{i,j}$). This value is used for comparing various versions of the Embodied Evolution algorithm with different parameter settings. We run each algorithm for $10$ independent runs, each one consisting of $20000$ generations.

\subsubsection{$7 \times 7$ scenario}
One of the key aspects of Embodied Evolution approaches is to make use of the neighboring agents during the evolutionary process. As seen earlier, an agent can copy for instance the behavior parameters of the best performing agent in its neighborhood, with some mutation. However, not always these parameters can benefit the agent, since the optimal behavior of each agent can be very different from that of its neighbors. The imitation problem demonstrates this point in that there can be large differences between adjacent cells, e.g. at the edge between the background and the digit. Moreover, these differences change from one digit to another. To further stress these differences, we define an additional scenario where we reduce the number of agents to $7\times7$, by assigning to each agent the control of a tile made of of $4\times4$ pixels. As such, in this case each agent controls the intensity level of $16$ pixels (rather than just one as in the $28\times28$ scenario), again for a sequence of $100$ images. Thus, the number of parameters per agent increases from $100$ to $4\times4\times100 = 1600$. Also in this case the behavior parameters are encoded in the genotype of each agent in a real-valued vector. However, differently from the previous scenario the spatial differences of adjacent tiles are larger, such that the difference between the optimal parameters of two neighboring agents is in turn larger.

\begin{figure}[ht!]
\begin{subfigures}
\subfloat[$28\times28$ scenario]{\includegraphics[width=0.5\columnwidth]{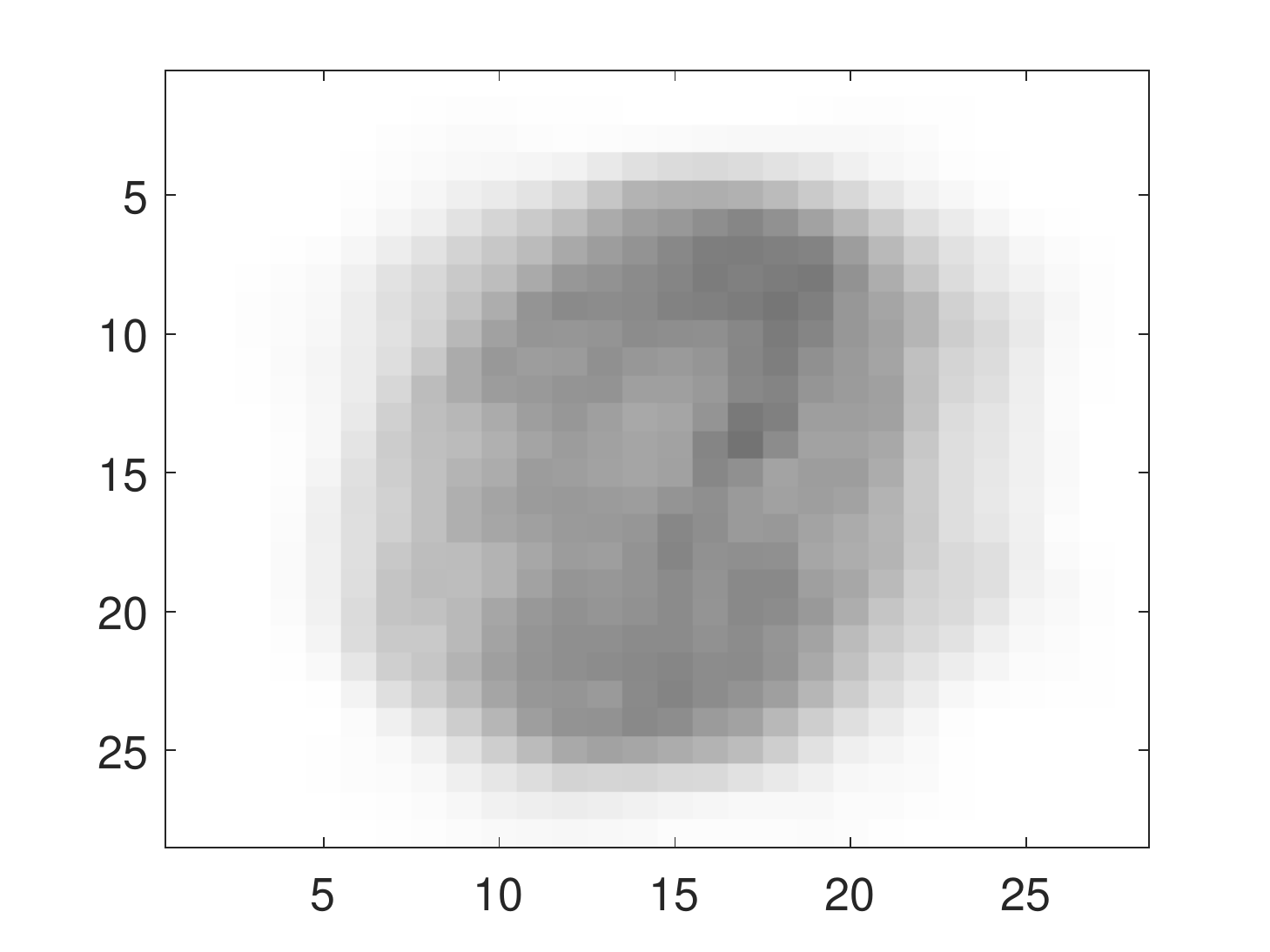}\label{fig:distMatrix28by28}}
\subfloat[$7\times 7$ scenario]{\includegraphics[width=0.5\columnwidth]{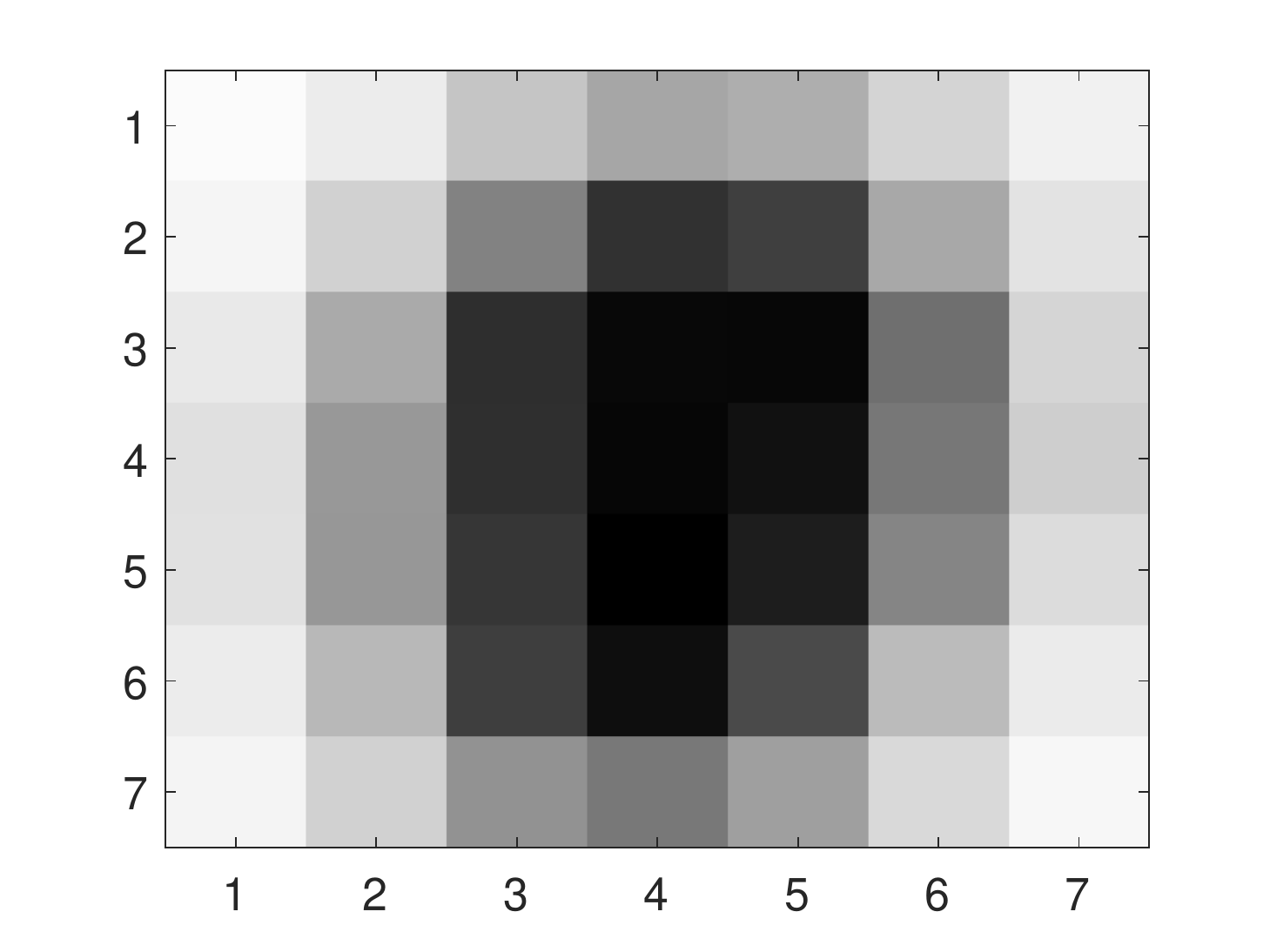}\label{fig:distMatrix7by7}}
\end{subfigures}
\caption{Imitation problem: Average differences (represented in grayscale) between the optimal parameters of each agent and those of the neighboring agents in the two scenarios.} \label{fig:compareAgentParameterDistances}
\end{figure}

To make this point clearer, in Figure~\ref{fig:compareAgentParameterDistances} we show the average (across $100$ MNIST images) differences of the optimal parameters of the agents with their neighbors for the two scenarios ($28\times28$ vs $7\times7$ agents). For each cell, we first find the average (across genes) differences with each of its neighbors, and then we take the average (across neighbors) of these differences. The two resulting matrices are finally normalized by scaling each cell value w.r.t. the maximum value among all cells in the two scenarios: in fact, the maximum value in the $7\times7$ scenario (Figure~\ref{fig:distMatrix7by7}) is about twice as big as the maximum value in the $28\times28$ scenario (Figure~\ref{fig:distMatrix28by28}), thus both matrices are scaled w.r.t. the maximum value in the $7\times7$ scenario. A higher grayscale intensity in a cell indicates a higher difference with its neighbors. It can be observed that the differences are much higher in the middle of the images (this is a consequence of the different shapes of the digits, while the cells at the border are more similar since they encode the background), and in general the differences in the $7\times7$ scenario are much higher (this is a consequence of the lower agent density, which leads to each agent controlling more pixels, such that the desired behavior can be quite different between adjacent tiles).

%TODO: actually we did 2 runs
In this scenario, we run various versions of the Embodied Evolution algorithm for $10$ independent runs, each one consisting of $100000$ generations: compared to the $28\times28$ scenario, we increase the number of generations since the number of parameters per agent is $16$ times higher. We use again the collective fitness $F_g$ for comparisons among the tested algorithms.

%---------------------------------------

\subsection{Illumination problem}
\begin{figure}[ht!]
\begin{subfigures}
\subfloat[$t=0$]{\includegraphics[width=0.32\columnwidth]{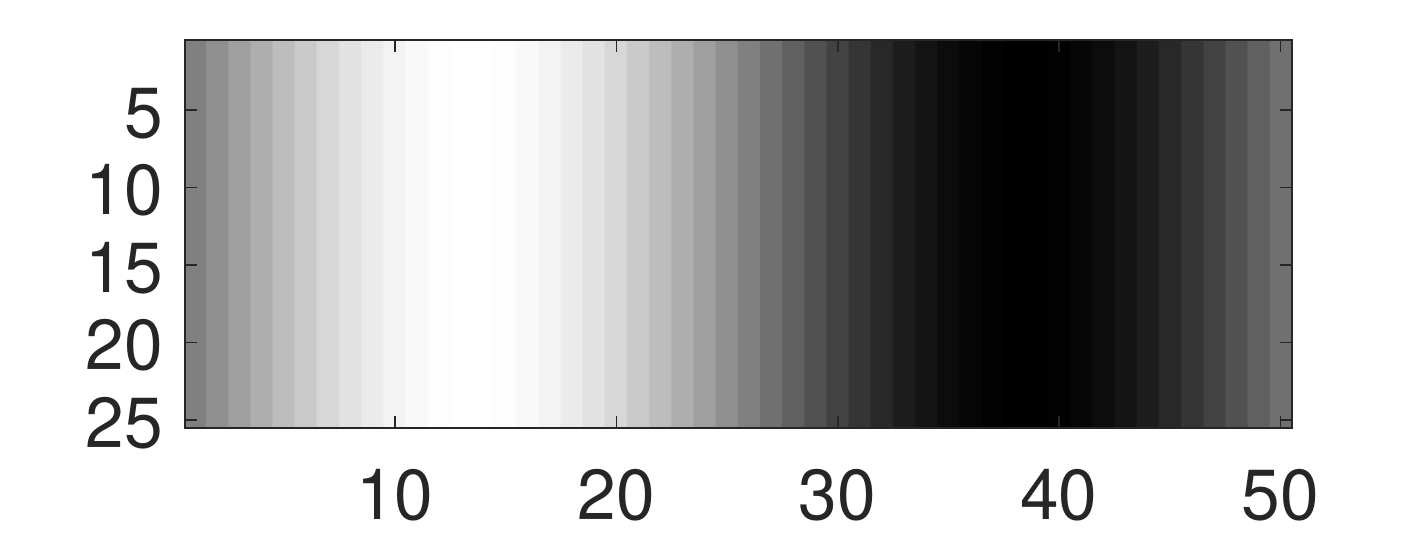}\label{fig:illuminationGroundTruthT0}}
\subfloat[$t=7$]{\includegraphics[width=0.32\columnwidth]{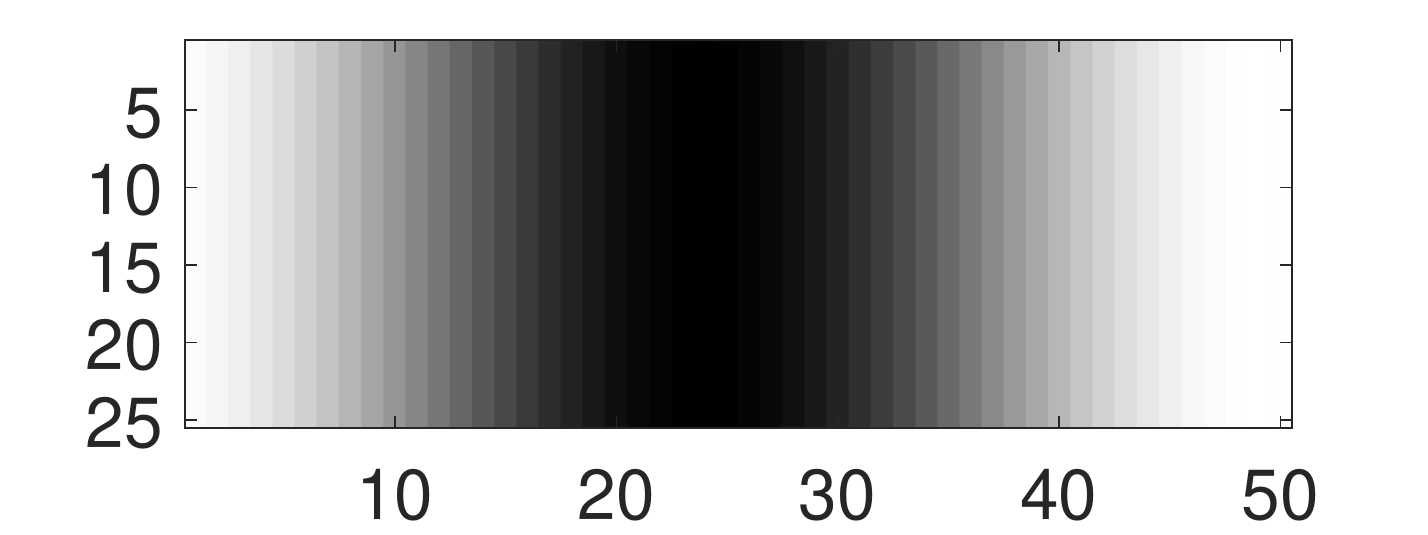}\label{fig:illuminationGroundTruthT7}}
\subfloat[$t=15$]{\includegraphics[width=0.32\columnwidth]{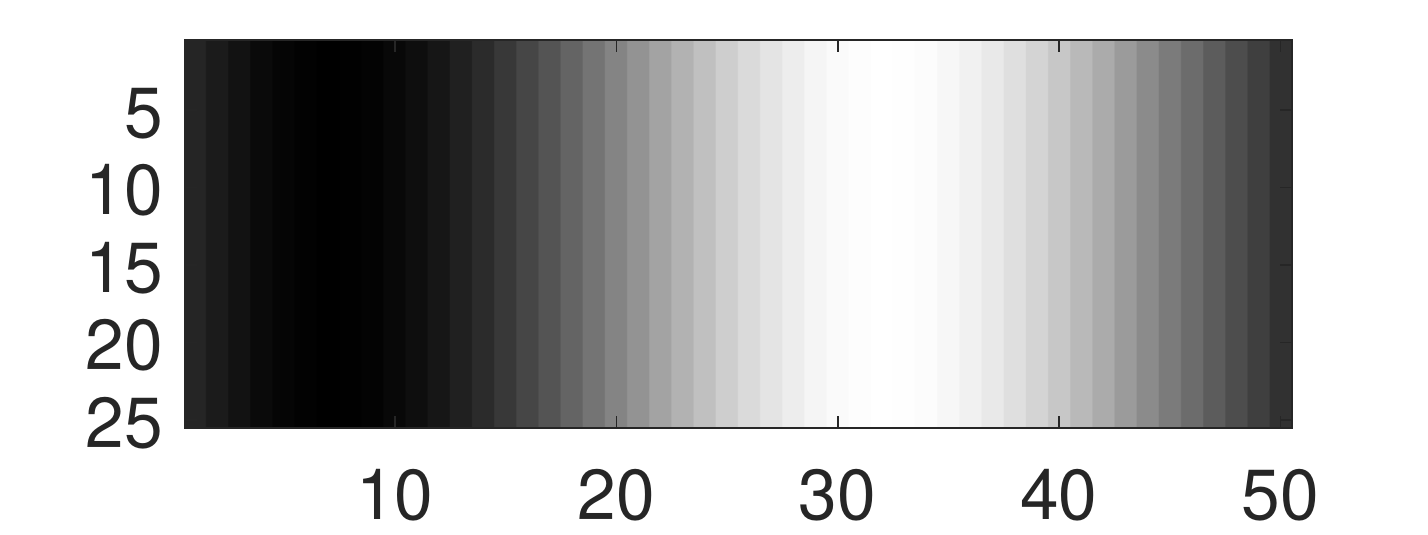}\label{fig:illuminationGroundTruthT15}}
\end{subfigures}
\caption{Illumination problem: Visualization of the optimal illumination (ground truth) at $t=0$, $t=7$ and $t=15$.} \label{fig:illuminationGroundTruth}
\end{figure}

This problem illustrates a possible application of the imitation problem. In this case, a network of agents is required to learn to optimally illuminate an environment. We consider a 2-dimensional grid of $25\times50 = 1250$ agents, where each agent controls the illumination level in its cell. We assume that different locations in the environment receive different levels of natural light during the day. Each agent $a_{i,j}$ is then required to learn the optimal local illumination, based on the hour of the day, $t \in \{0, 1, \hdots, 23\}$, and its location, ${i,j}$. We define the optimal illumination (ground truth) for each cell location as follows:
\begin{equation}
l_{i,j}(t) = sin \left( \frac{2 \pi j}{n} + \frac{2\pi t}{24} \right) 
\label{eq:illuminationEquation}
\end{equation}
$l_{i,j}(t) = sin \left( \frac{2 \pi j}{n} + \frac{2\pi t}{24} \right)$, 
\noindent{}where $n$ is the maximum value of $j$. To convert $l_{i,j}(t)$ to pixel values, we scale its values to $[0,1]$ (i.e. $l_{i,j}(t) = (l_{i,j}(t) + 1)/ 2$). A visualization of this environment at $t=0$, $t=7$ and $t=15$ is shown in Figure~\ref{fig:illuminationGroundTruth}.

In this case we further specify the problem in two scenarios, one in which each agent's behavior depends on a single real-valued parameter, and one in which the behavior is represented with a vector of real values.

\subsubsection{Single parameter scenario}
In this scenario, each agent has only one real-valued parameter, $x^{i,j} \in [0,50]$. The agent's behavior is then obtained by replacing in Equation~\eqref{eq:illuminationEquation} the desired location parameter, $j$, with the agent parameter $x^{i,j}$, and setting $n=50$, as follows:
\begin{equation}
a_{i,j}(t) = sin \left( \frac{2 \pi  x^{i,j}}{50} + \frac{2 \pi  t}{24} \right)
\label{eq:illuminationEquationAgent}
\end{equation}
At the beginning of the algorithm, $x^{i,j}$ is randomly initialized in $[0,50]$. The fitness value of each agent $f_{i,j}$ is calculated as the average the difference from the optimal illumination during the day, namely:
\begin{equation}
f_{i,j} = \frac{1}{24} \sum^{23}_{t=0} \mid l_{i,j}(t) - a_{i,j}(t) \mid 
\label{eq:illuminationFitnessCalcSingle}
\end{equation}
We use the Embodied Evolution approach to find the optimal parameters of each agent that can collectively produce the desired illumination pattern. The whole network has in this case $25\times50$ parameters in total. We use again the collective fitness $F_g$ (defined as for the imitation problem) to perform comparisons among various versions of the Embodied Evolution algorithm. In this case, we perform $10$ independent runs for each version of the Embodied Evolution algorithm for $5000$ generations.

\subsubsection{Vector representation scenario}
In this scenario, we encode the agents' behavior using a $24$-dimensional real-valued vector (similarly to the imitation problem), $\boldsymbol{X}^{i,j} = (x^{i,j}_1,x^{i,j}_2,\hdots, x^{i,j}_{24}) \in [0,1]^{24}$, to control the agent-local level of illumination at each time step $t = \{0, 1, \hdots, 23\}$. Here, the values in $[0,1]$ encode the amount of light, from no light ($0$) to the maximum amount of light ($1$). The main difference w.r.t. the single parameter scenario is then that in this case we do not enforce a sinusoidal pattern in each agent, but this should be discovered by the Embodied Evolution process. At the beginning of the algorithm, the vector $\boldsymbol{X}^{i,j}$ of each agent is randomly initialized in $[0,1]^{24}$. Similarly to the single parameter scenario, the fitness value of each agent $f_{i,j}$ is the average difference from the optimal illumination during the day, namely:
\begin{equation}
f_{i,j} = \frac{1}{24} \sum^{23}_{t=0} \mid l_{i,j}(t) - x^{i,j}_t \mid
\label{eq:illuminationFitnessCalc}
\end{equation}
In this scenario, there are in total $3\times10^4$ parameters ($25\times50$ agents $\times 24$ parameters). We perform $10$ independent runs for each version of the algorithm for $20000$ generations (the number of generations is increased w.r.t. the single parameter scenario due to the larger number of parameters). We use again the collective fitness $F_g$ to perform comparisons among the tested algorithms.

%---------------------------------------------

\subsection{Distributed model of indoor human presence and activity}

IoT is a growing application area of Machine Learning and distributed algorithms. Typical applications in this field involve, for instance, distributed anomaly detection~\cite{bosman2013anomaly,bosman2013online,bosman2014online,bosman2015ensembles,bosman2017spatial}, transmission power control~\cite{pace2019intelligence} or other forms of distributed intelligence~\cite{dartmann2019big,kumar2019machine}. Here, we consider an IoT-based distributed model in which a number of sensors nodes are deployed in an indoor environment in order to detect human presence and activity, an urgent matter in terms of security and privacy. One important difference w.r.t. the two previous problems is that in this case the local behavior is not intended as a node-local actuation affecting the environment (like setting the local intensity or illumination), but rather the output of a node-local data-driven model that does not affect the environment (in this case, the detected presence/activity). This somehow reflects the flexibility and general-purposeness of the Embodied Evolution scheme.

In our experiments, we use a real-world dataset taken from~\cite{morgner2017privacy}, which consists of temperature and humidity sensor readings of three to five sensor nodes distributed in three different rooms. The sensor readings are associated with ground truth labels on the human presence (if the room is occupied or not by a human) and activity (one of four human activities, i.e.: reading, standing, walking, and working on a PC).

For training and testing the model, we use a similar experimental setting as the original paper~\cite{morgner2017privacy}, i.e. we split the sensor readings of each node in windows of $150$ samples each, where each window is obtained by shifting the previous one by $30$ samples. Overall, we obtain about $800$, $1400$ and $450$ windows for the human presence task, and $300$, $450$ and $150$ windows for the human activity task in room A, B and C respectively. We use $80\%$ of these windows for computing the fitness of the nodes' model (training accuracy) and the rest for testing. We perform testing at each generation during the evolutionary process, in order to collect the test accuracy values used for comparing the different versions of the EE algorithm.

As for the model, we employ on each node a feed-forward neural network (FFNN) architecture. We concatenate the raw measurements of the temperature and humidity sensors within a window and feed these input data to the FFNN. Therefore, the input layer of the networks has $300$ neurons ($150$ each for temperature and humidity). The temperature and humidity measurements are scaled (separately) in the range $[0,1]$, by subtracting the minimum value and dividing by the maximum value. We use one hidden layer with $100$ hidden neurons. 

We consider the human presence detection and activity detection as two separate tasks, i.e. we train and test a separate FFNN on each node for each of the two tasks. In the case of presence detection, the output layer consists of two neurons for deciding if there is a human in the room or not. In the case of activity detection, we use four output neurons, one for each activity. In both cases the output neuron with the highest activation level is selected as the final output of the network.

\begin{figure}[ht!]
\begin{subfigures}
% \subfloat[room A]{\includegraphics[width=0.33\columnwidth,valign=t]{figs/locationA.pdf}\label{fig:locationA}}
% \subfloat[room B]{\includegraphics[width=0.33\columnwidth,valign=t]{figs/locationB.pdf}\label{fig:locationB}} 
% \subfloat[room C]{\includegraphics[width=0.33\columnwidth,valign=t]{figs/locationC.pdf}\label{fig:locationC}}
\captionsetup[subfigure]{labelformat=empty}
\subfloat[]{\includegraphics[width=0.3\columnwidth,valign=c]{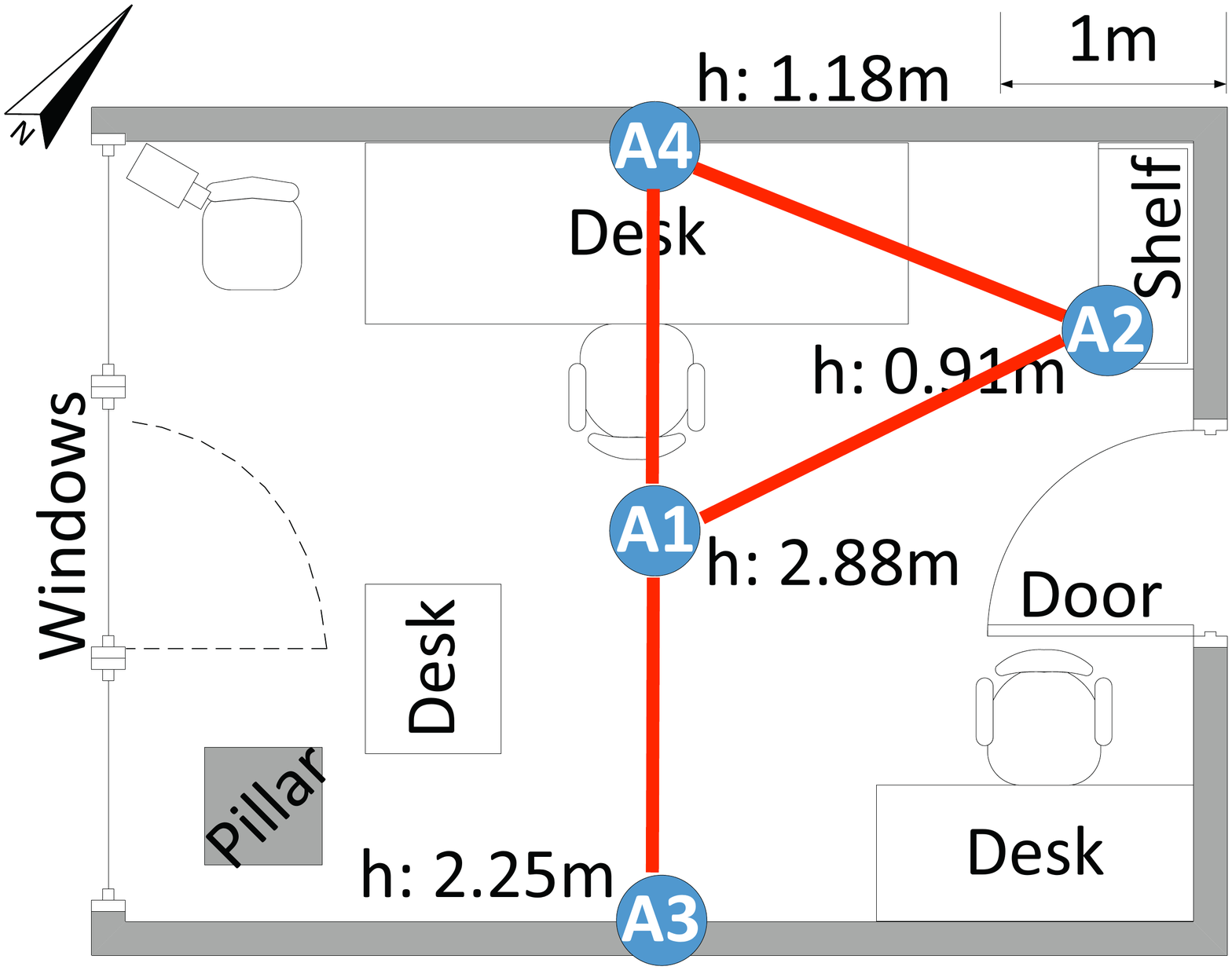}\label{fig:locationA}}
\subfloat[]{\includegraphics[width=0.39\columnwidth,valign=c]{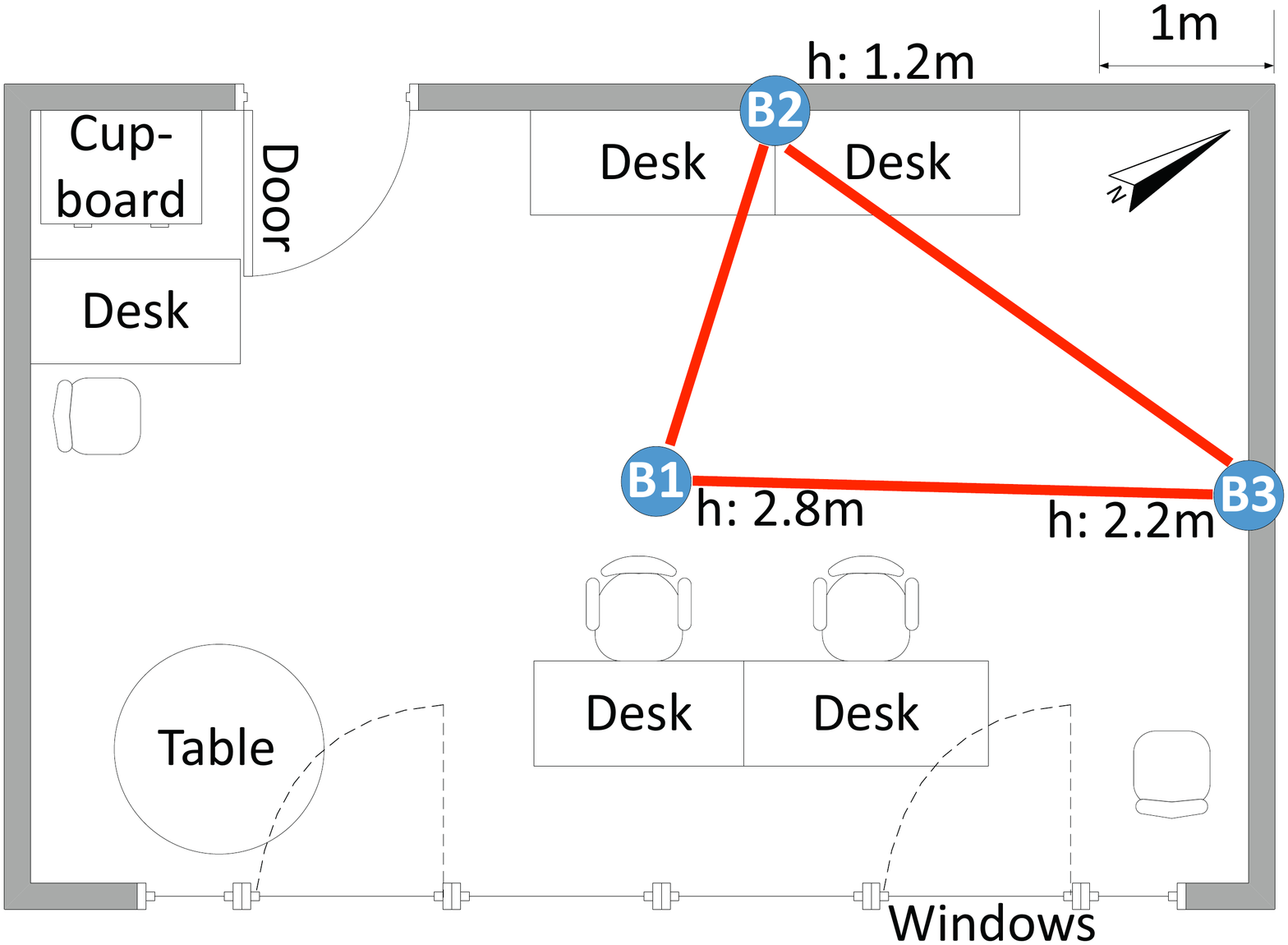}\label{fig:locationB}} 
\subfloat[]{\includegraphics[width=0.3\columnwidth,valign=c]{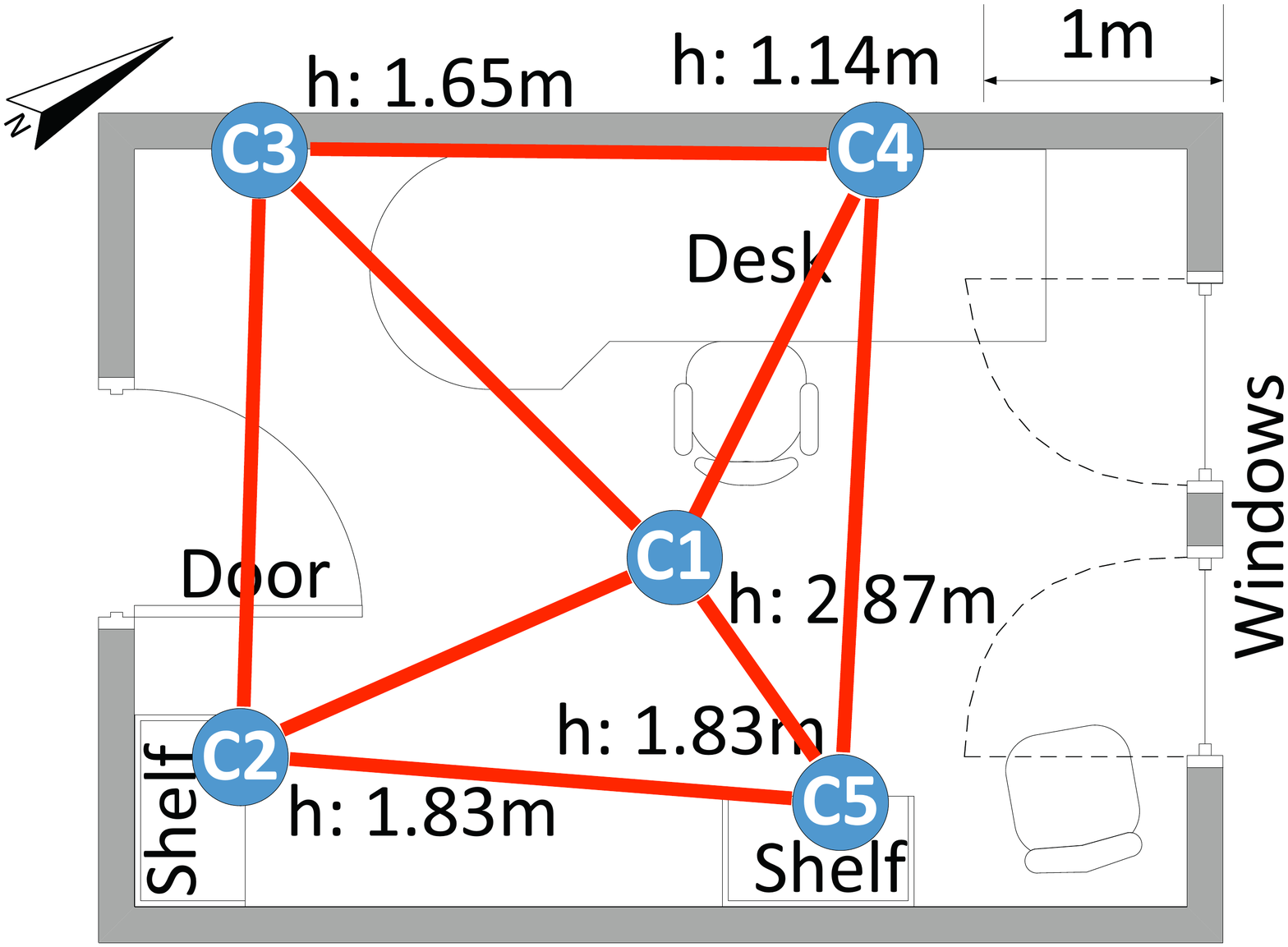}\label{fig:locationC}}
\end{subfigures}
\caption{Network topology for the three different rooms described in~\cite{morgner2017privacy}: room A, B and C (from left to right).} \label{fig:roomClimate}
\end{figure}

It is important to remark that in~\cite{morgner2017privacy} no communication occurs between nodes. Rather, the authors extract the relevant features from the sensor readings and use these features to train a model --separately for each node-- in order to detect the human presence and activity. Differently from this approach, here we assume instead that there is communication between nodes. However, we use communication only to allow parameter exchange for the copy and crossover operators, i.e., we do not use communication to transfer sensor data across the nodes. The network topologies (one per room) we assumed in our experimentation are shown in Figure~\ref{fig:roomClimate}.

In this problem, the behavior parameters of each node are the weights (in $[-1,1]$) of its FFNN. Based on the size of the networks used, there are in total of $(300 + 1) \times 100 + (101 + 1)\times 2 = 30302$ and $(300 + 1) \times 100 + (101 + 1)\times 2 = 30504$ parameters in the human presence and human activity detection tasks respectively ($+1$ indicates bias neurons to hidden and output layers). These parameters are optimized using the distributed Embodied Evolution approach, by performing local mutations and/or crossover by means of each node's parameters exchange with its neighbors. As a note, since the nodes are positioned at different locations in each room, their local readings are usually different, although there is a certain spatial continuity. Therefore, the optimum parameter settings of their FFNNs may be different and as such our goal is to test if parameter exchange can be beneficial also in this application scenario.

Similarly to the imitation and illumination problems, also in this case we perform $10$ independent runs for each version of the EE algorithm for $10000$ generations, for each room and task. As said earlier, we use the test accuracy to perform comparisons among the different tested algorithms.

%---------------------------------------------

\subsection{Versions of the Embodied Evolution algorithm}
We tested in total five versions of the Embodied Evolution algorithm presented in the previous section, configured as follows:
\begin{itemize}[leftmargin=*]
    \item[-] \textbf{XoverRand:} this version of the algorithm is the one given in Algorithm~\ref{alg:embodiedNetworkEvolution}.
    
    \item[-] \textbf{XoverBest:} this version is similar to the previous one, the only difference being that the function $rand(\boldsymbol{X}^{i,j})$ in Line~\ref{ln:xover} of Algorithm~\ref{alg:embodiedNetworkEvolution} is replaced with $best(\boldsymbol{X}^{i,j})$. In other words, this algorithm performs crossover with the best agent in each agent's neighborhood.
    
    \item[-] \textbf{HillClimbing:} in this version, we do not use any neighborhood function, therefore the agents do not share any information with their neighbors. We modify Algorithm~\ref{alg:embodiedNetworkEvolution} by deleting Line~\ref{ln:xover} (where the crossover operator is used), and changing Line~\ref{ln:mutation} with: $\boldsymbol{X}^{\prime} \gets \boldsymbol{X}^{i,j} + \mathcal{N}(0,\sigma)$.
    % Line~\ref{ln:hc} given in Algorithm~\ref{alg:HillClimbing}.
    % %
    % \begin{algorithm}[ht]
    %     \begin{algorithmic}[1]
    %         \State $\boldsymbol{X}^{\prime} \gets \boldsymbol{X}^{i,j} + \mathcal{N}(0,\sigma)$\label{ln:hc}
    %     \end{algorithmic}
    % \caption{HillClimbing}\label{alg:HillClimbing}
    % \end{algorithm}
    
    \item[-] \textbf{CopyBest:} in this version, each agent copies the genotype of the best agent in its neighborhood and then applies the mutation operator. Similar to HillClimbing, crossover is not used. Thus,
    we modify Algorithm~\ref{alg:embodiedNetworkEvolution} by deleting Line~\ref{ln:xover} and changing Line~\ref{ln:mutation} with: 
    $\boldsymbol{X}^{\prime} \gets best(\boldsymbol{X}^{i,j}) + \mathcal{N}(0,\sigma)$.
    % Line~\ref{ln:CopyBestMutation} given in Algorithm~\ref{alg:CopyBest}.
    % %
    % \begin{algorithm}[ht]
    %     \begin{algorithmic}[1]
    %         \State $\boldsymbol{X}^{\prime} \gets best(\boldsymbol{X}^{i,j}) + \mathcal{N}(0,\sigma)$\label{ln:CopyBestMutation}
    %     \end{algorithmic}
    % \caption{CopyBest}\label{alg:CopyBest}
    % \end{algorithm}
    
    \item[-] \textbf{CopyRand:} in this version, each agent copies the genotype of a randomly selected agent in its neighborhood and then applies the mutation operator. Similarly to CopyBest, also in this case crossover is not used. Thus,
    we modify Algorithm~\ref{alg:embodiedNetworkEvolution} by deleting Line~\ref{ln:xover} and changing Line~\ref{ln:mutation} with: $\boldsymbol{X}^{\prime} \gets rand(\boldsymbol{X}^{i,j}) + \mathcal{N}(0,\sigma)$.
    % Line~\ref{ln:CopyRandMutation} given in Algorithm~\ref{alg:CopyRand}.
    % %
    % \begin{algorithm}[ht]
    %     \begin{algorithmic}[1]
    %         \State $\boldsymbol{X}^{\prime} \gets rand(\boldsymbol{X}^{i,j}) + \mathcal{N}(0,\sigma)$\label{ln:CopyRandMutation}
    %     \end{algorithmic}
    % \caption{CopyRand}\label{alg:CopyRand}
    % \end{algorithm}
    
\end{itemize}

%% file: 4_experimentalResults.tex
\section{Experimental Results}
\label{sec:results}
On the three problems and related scenarios introduced above, we tested the five versions of the EE algorithm with different combinations of mutation rates ($mr$) and crossover probability ($cp$)\footnote{We report a detailed comparative analysis of the different algorithms and parameter settings in the Appendix.}. We use uniform crossover ($cr = 0.5$) for all experiments. In the case of the imitation and illumination problems with vector representation (both with domain $[0,1]$), we use $mr \in \{0.0001, 0.001, 0.01\}$, while for the illumination problem with single parameter (with domain $[0,50]$) we use $mr \in \{0.005, 0.05, 0.5\}$. In the case of the distributed model of indoor human presence and activity (with domain $[-1,1]$), we use $mr \in \{0.0001, 0.001, 0.01\}$.%, namely $1\%$, $0.1\%$ and $0.01\%$ of the search domains

It should be noted that the imitation and illumination problems are formulated as \emph{minimization} problems, while the distributed model problem is a \emph{maximization} problem.

%---------------------------------------------

\subsection{Imitation problem}
In the following, we analyze separately the numerical results on the imitation problem in the two scenarios described above: $28\times28$ agents and $7\times7$ agents.

\subsubsection{$28\times28$ scenario}

\begin{figure}[ht!]
\begin{subfigures}
\subfloat[Imitation problem ($28\times28$ scenario)]{\includegraphics[width=0.48\columnwidth]{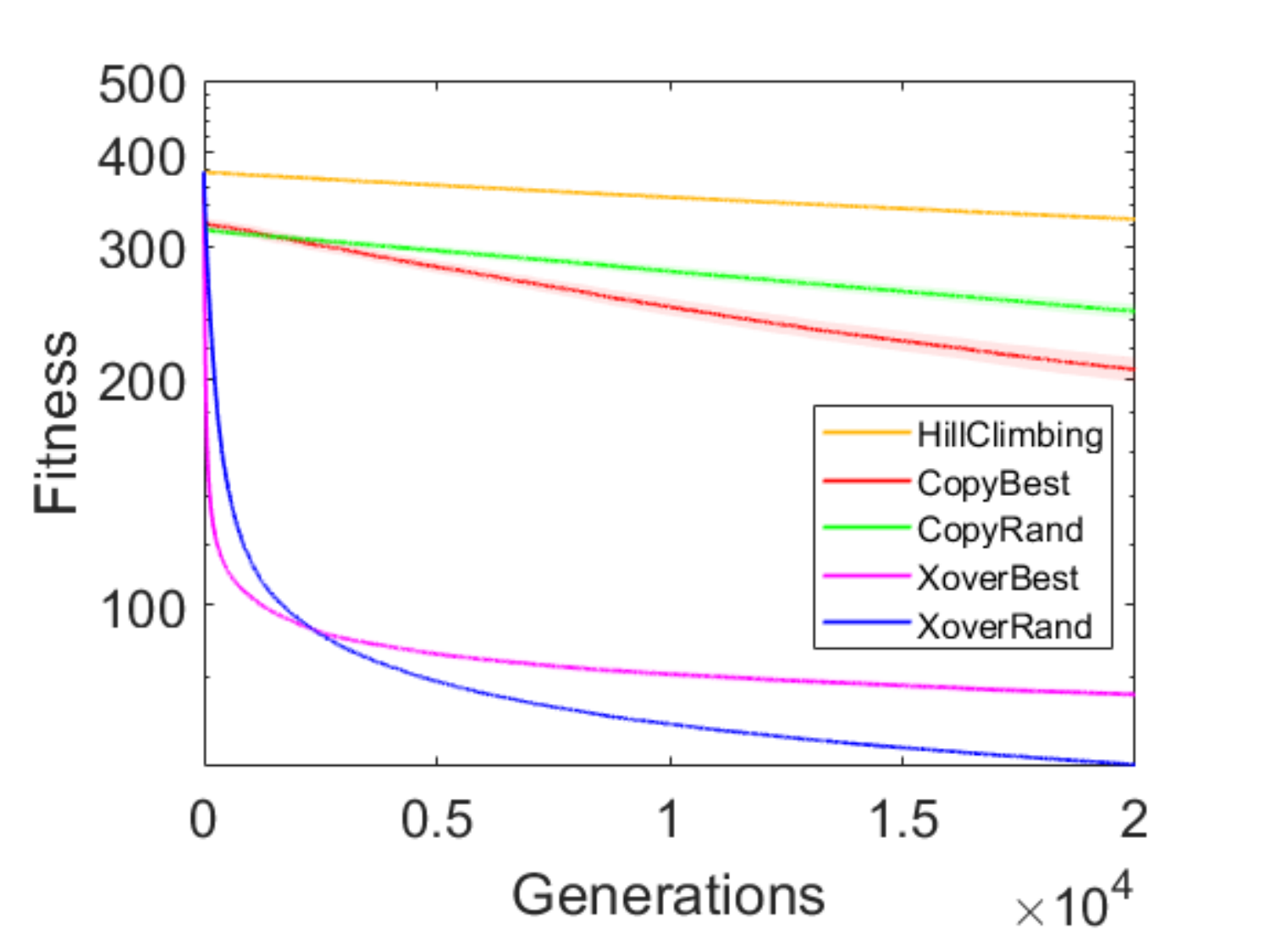}\label{fig:fitnessTrendMNIST28x28}}
\subfloat[Imitation problem ($7\times7$ scenario)]{\includegraphics[width=0.48\columnwidth]{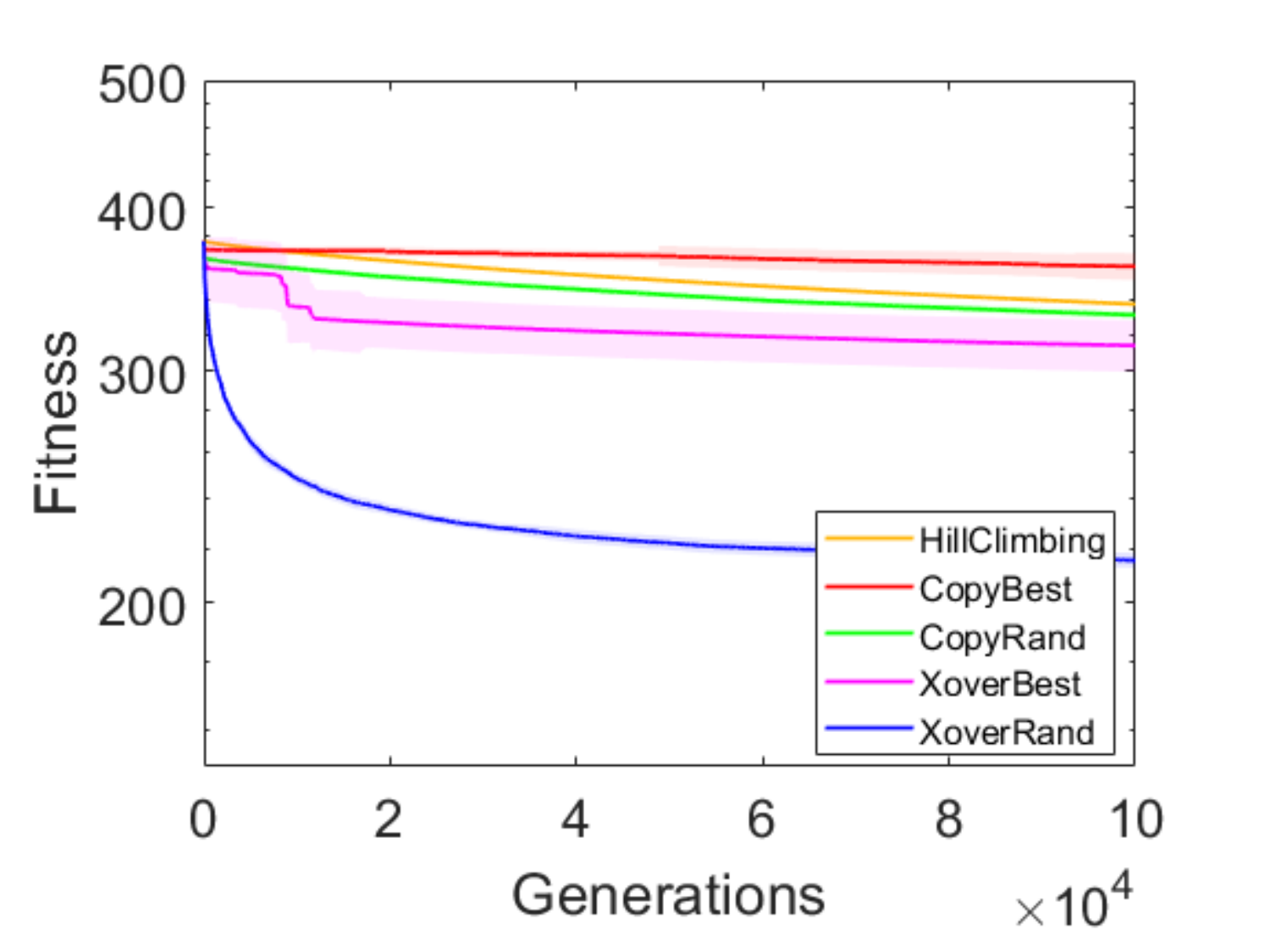}\label{fig:fitnessTrendMNIST7x7}}

\subfloat[Illumination problem (Single parameter scenario)]{\includegraphics[width=0.48\columnwidth]{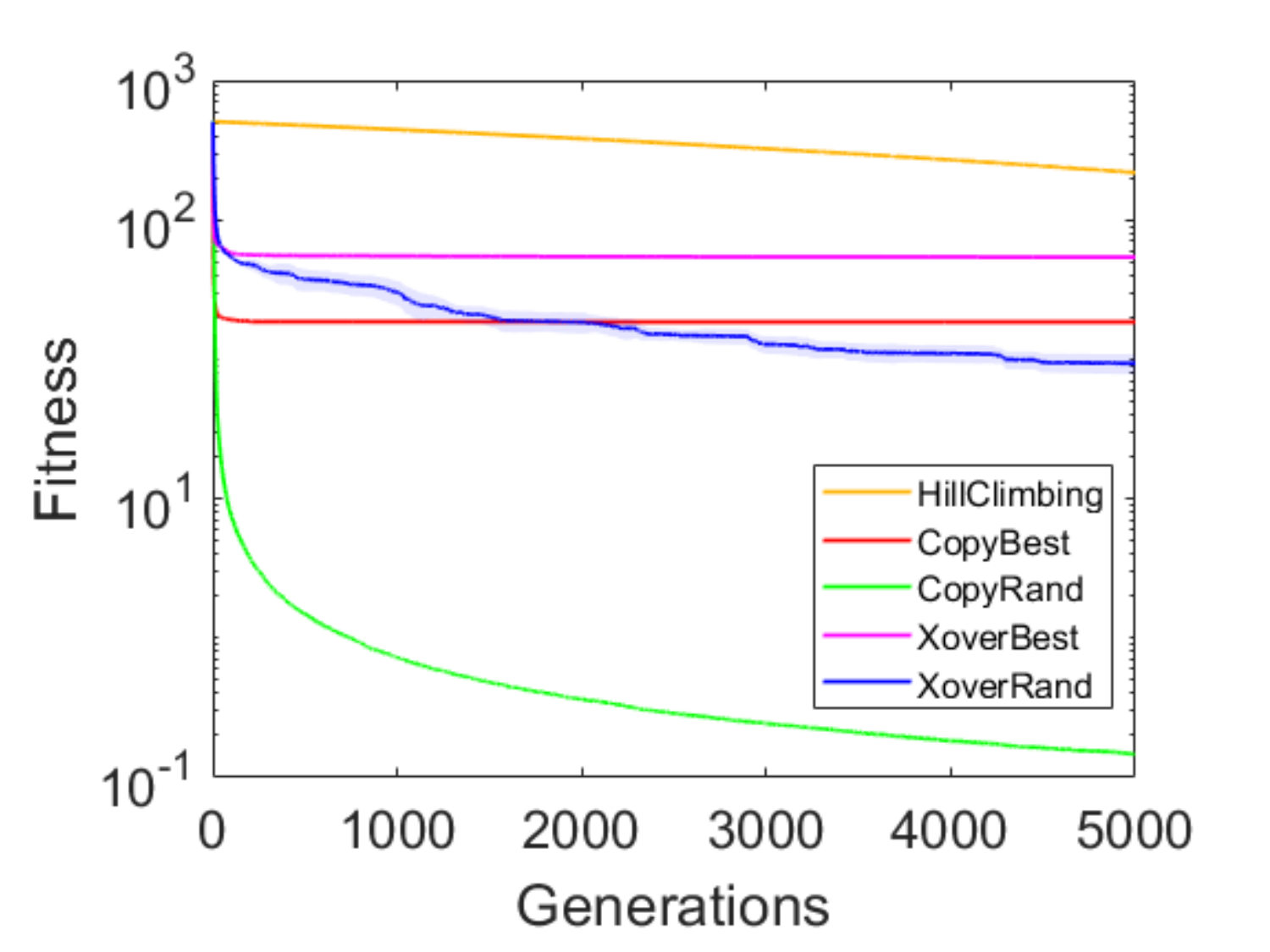}\label{fig:fitnessTrendsIlluminationSingle}}
\subfloat[Illumination problem (Vector representation scenario)]{\includegraphics[width=0.48\columnwidth]{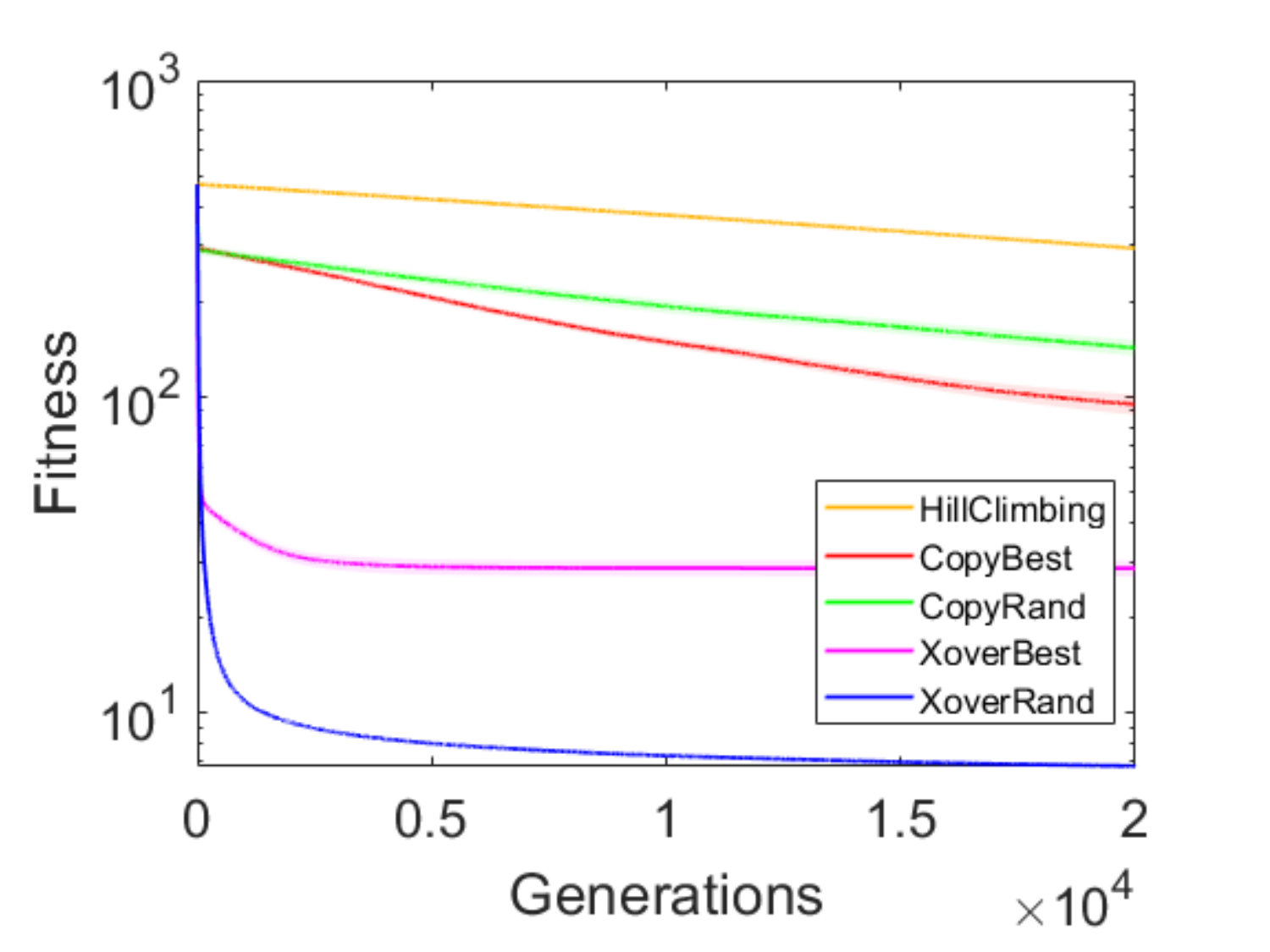}\label{fig:fitnessTrendsIlluminationVector}}

\end{subfigures}
\caption{Imitation and illumination problems: Collective fitness (mean $\pm$ std. dev. across $10$ runs) obtained with different EE algorithms during the evolutionary process. For CopyBest, CopyRand, XoverBest and XoverRand we show the fitness trend obtained with the best parameter setting.} \label{fig:comparisonOfAlgorithms}
\end{figure}

We tested the five Embodied Evolution algorithms, with $cp \in \{0.2, 0.5, 1.0\}$, $cr=0.5$, and $mr \in \{0.0001, 0.001, 0.01\}$. Figure~\ref{fig:fitnessTrendMNIST28x28} shows the average (across $10$ runs per algorithm) collective fitness ($F_g$) trends obtained with the tested algorithms. The shaded areas represent the range $\pm$ std. dev. In the case of CopyBest, CopyRand, XoverBest and XoverRand we show the fitness trend obtained with the best parameter setting according to the Nemenyi test \cite{nemenyi1962distribution}, see~\ref{app:stats}.

We observe that performing crossover with a random neighbor performs better than performing crossover with the best neighbor, which in turn performs better than not doing any crossover. This indicates that exchanging partial components of the genotype helps the optimization process, even though the optimal parameters of the neighbors are different. The fact that crossing over with a random neighbor is more efficient than doing that with the best neighbor is likely due to a higher chance of selecting a neighbor whose optimal behavior is closer to that of the focal agent, which might not be the case of the best neighbor (e.g. if the focal agent is on the digit while its best neighbor is on the background).

We further observe that the versions that copy the neighbors perform better than HillClimbing. Also, when crossover is not used, lower mutation rates appear to slow the convergence: being the algorithm population-less, the only way to converge faster is to perform larger mutations. However, we observe the opposite effect when crossover is used: since mutation is performed \emph{after} crossover, in this case too high mutation rates might obliterate any advantage obtained from the neighbor's exchanged genes. Overall, XoverRandCP05 (this notation, used also in the following, indicates $cp=0.5$) appears to perform the best. However, as we show in the Appendix, different mutation rates perform differently ($mr=0.01$ performs the worst).

\begin{figure}[ht!]
\begin{subfigures}
\subfloat[]{\includegraphics[width=0.25\columnwidth]{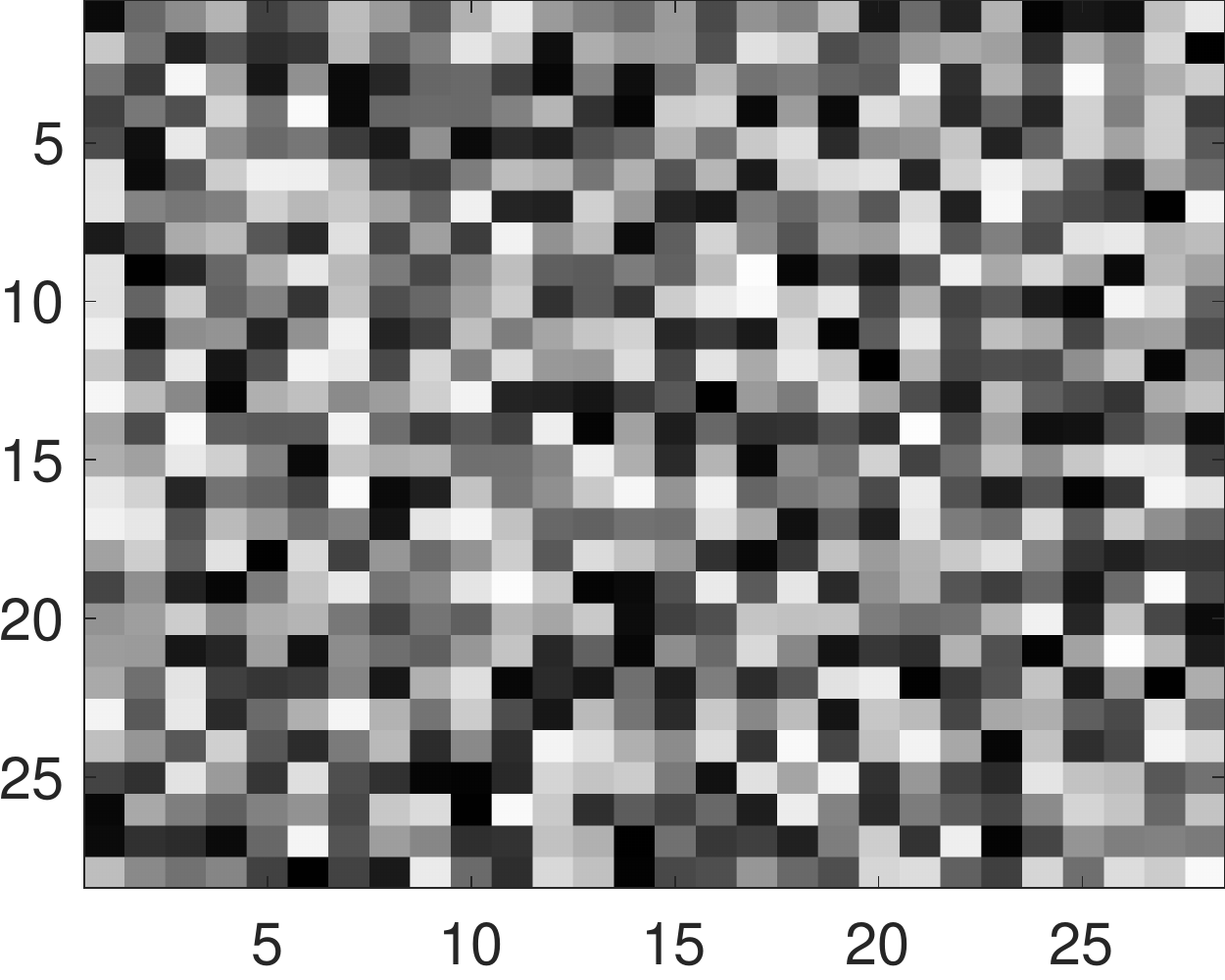}\label{fig:intervalsHillClimbing101}}
\subfloat[]{\includegraphics[width=0.25\columnwidth]{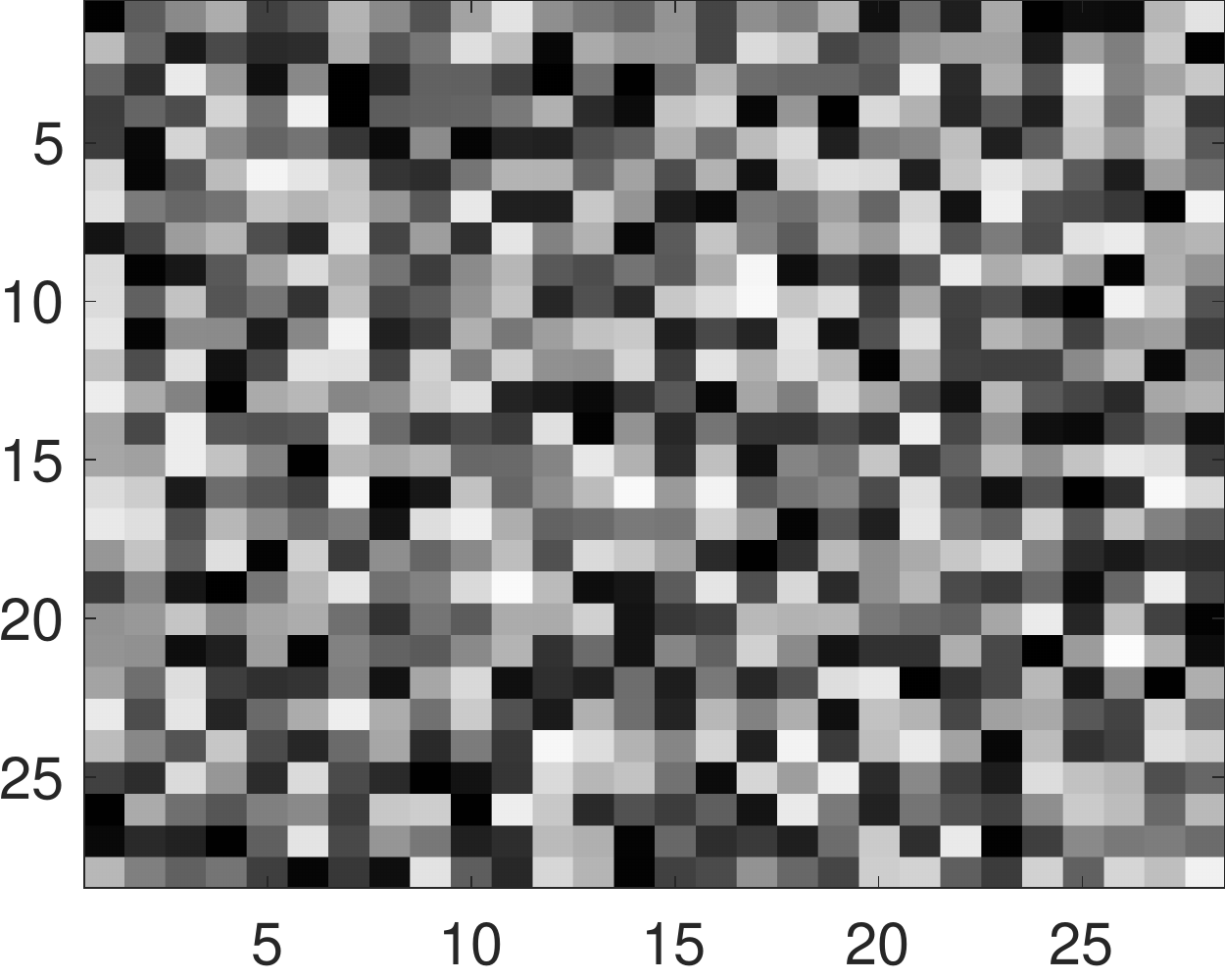}\label{fig:intervalsHillClimbing1000}}
\subfloat[]{\includegraphics[width=0.25\columnwidth]{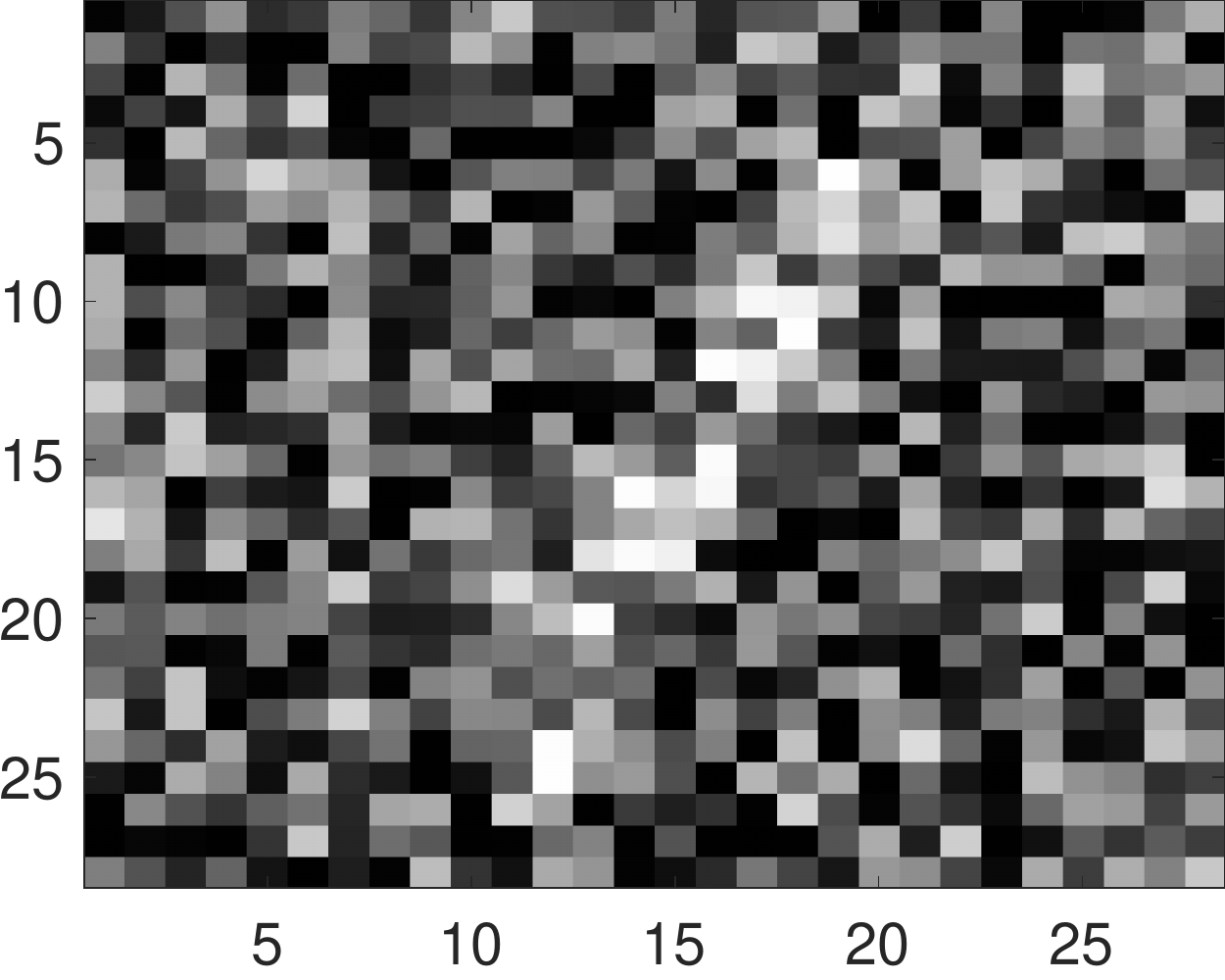}\label{fig:intervalsHillClimbing10000}}
\subfloat[]{\includegraphics[width=0.25\columnwidth]{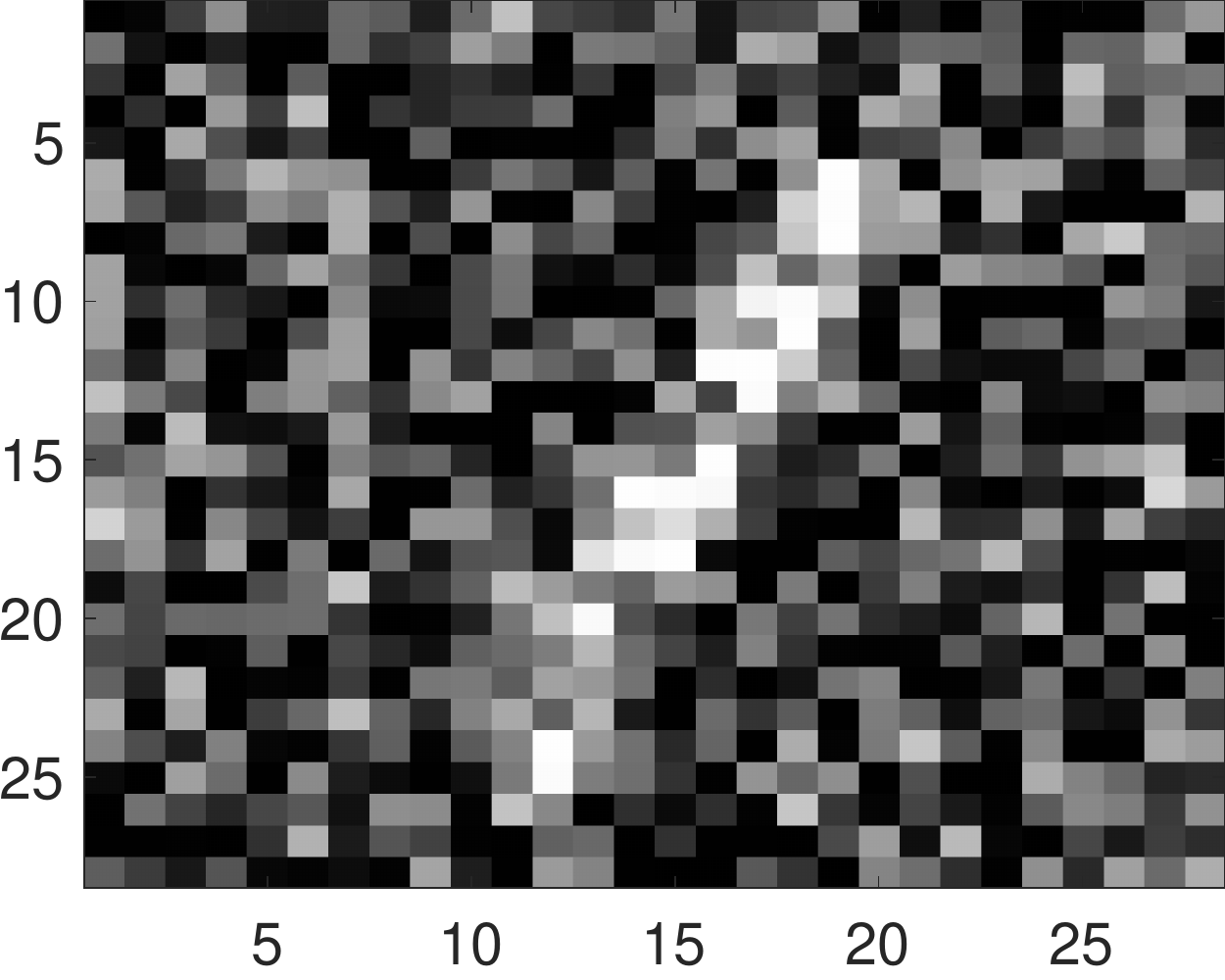}\label{fig:intervalsHillClimbing20000}}

\subfloat[]{\includegraphics[width=0.25\columnwidth]{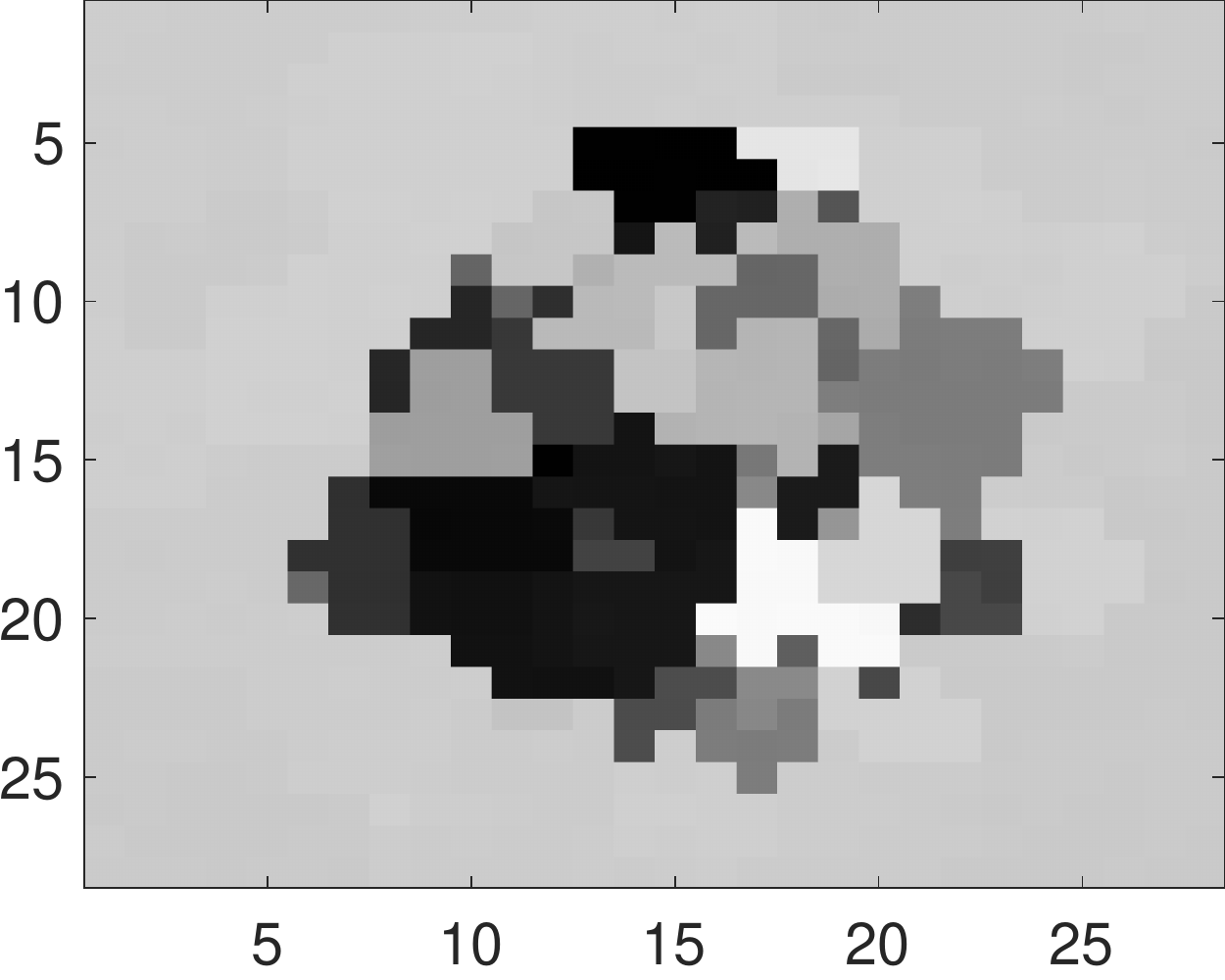}\label{fig:intervalsCopyBest101}}
\subfloat[]{\includegraphics[width=0.25\columnwidth]{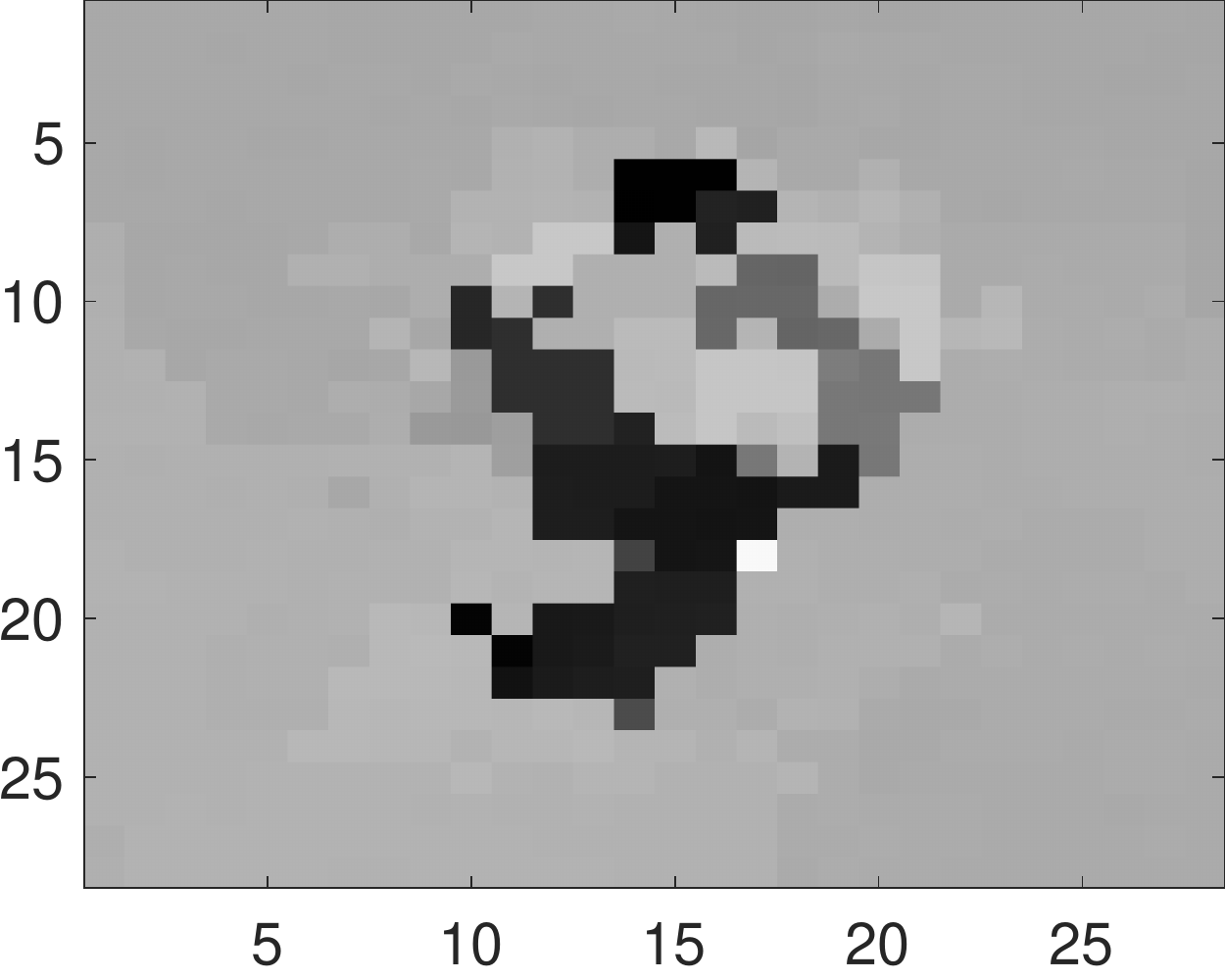}\label{fig:intervalsCopyBest1000}}
\subfloat[]{\includegraphics[width=0.25\columnwidth]{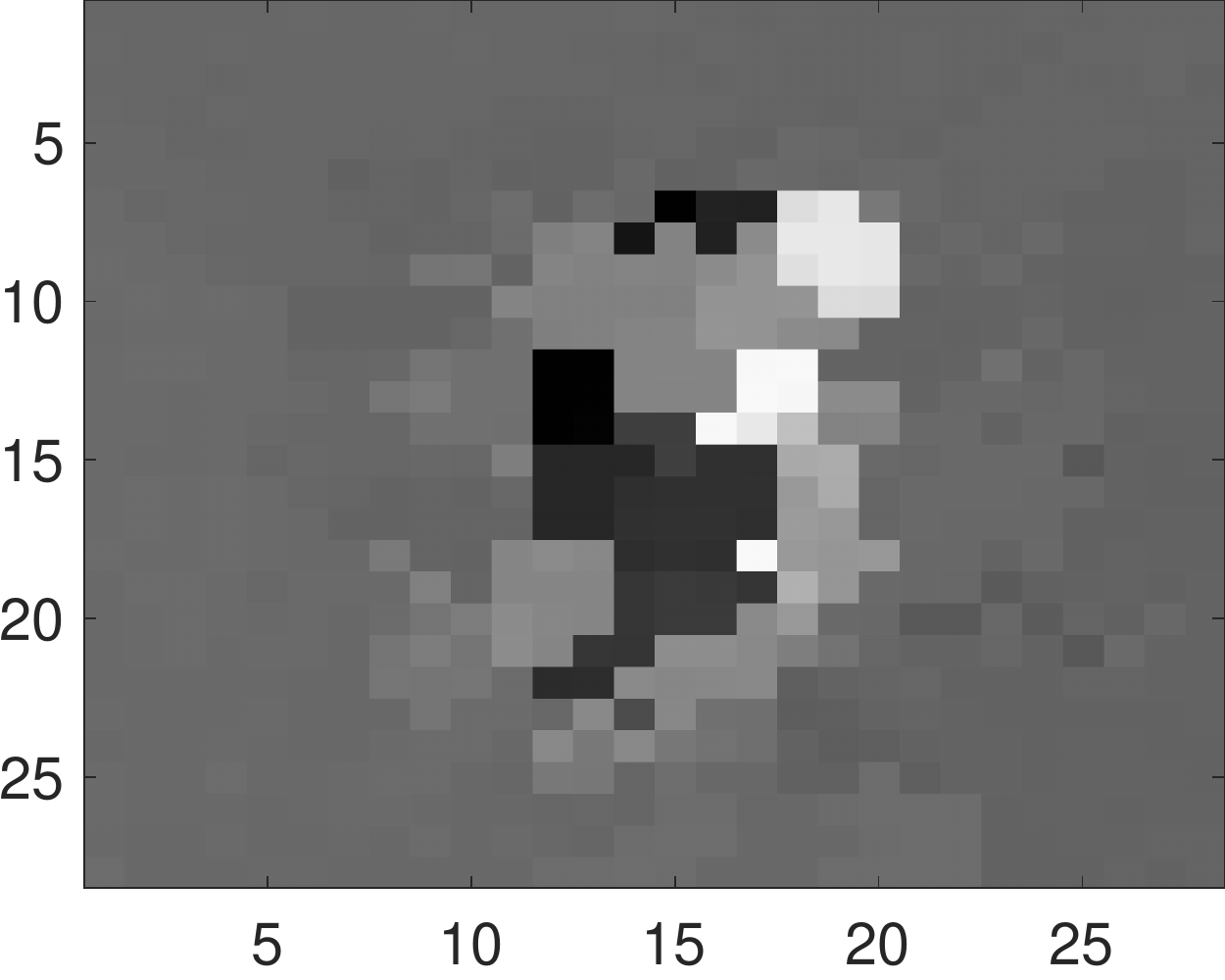}\label{fig:intervalsCopyBest10000}}
\subfloat[]{\includegraphics[width=0.25\columnwidth]{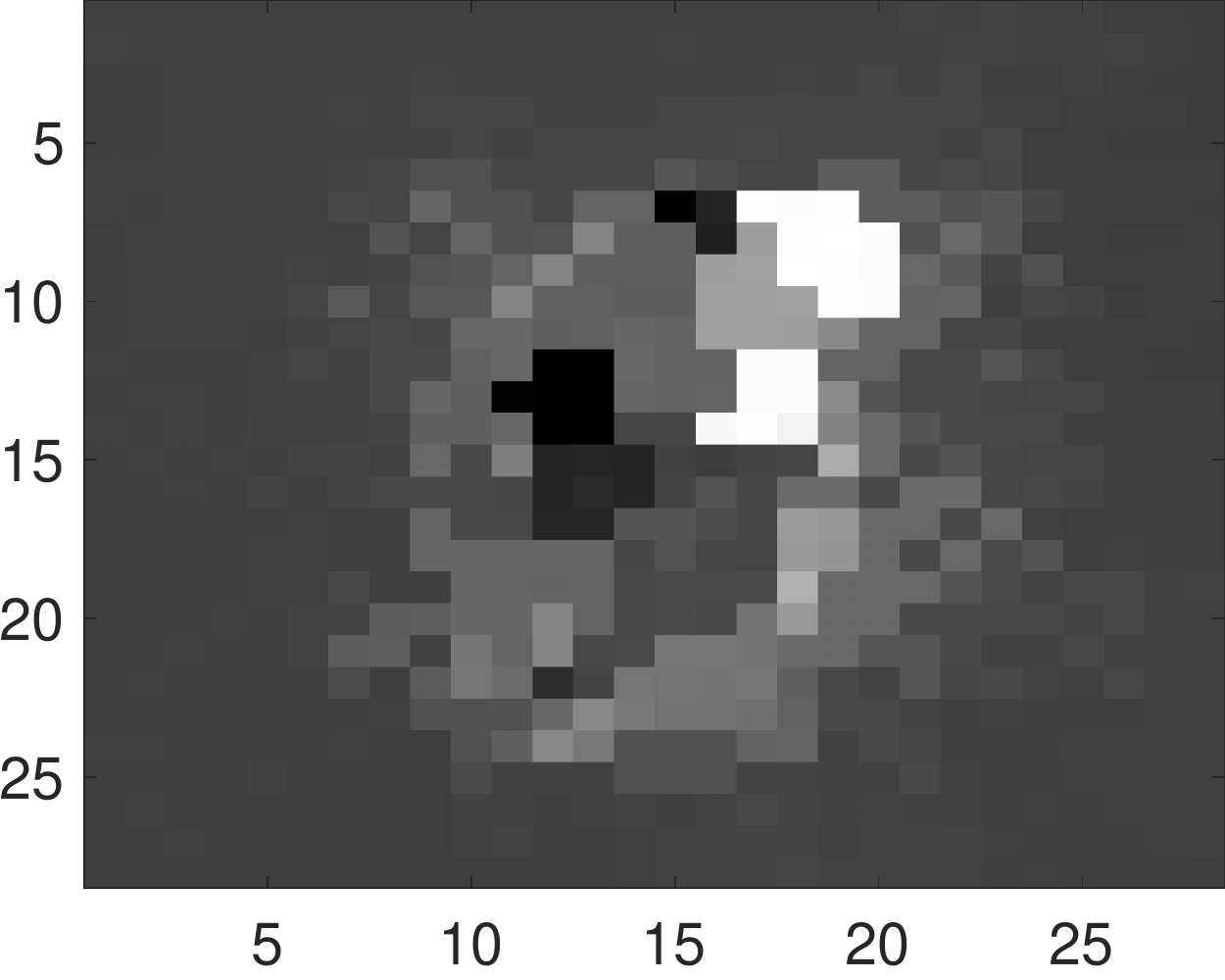}\label{fig:intervalsCopyBest20000}}

\subfloat[]{\includegraphics[width=0.25\columnwidth]{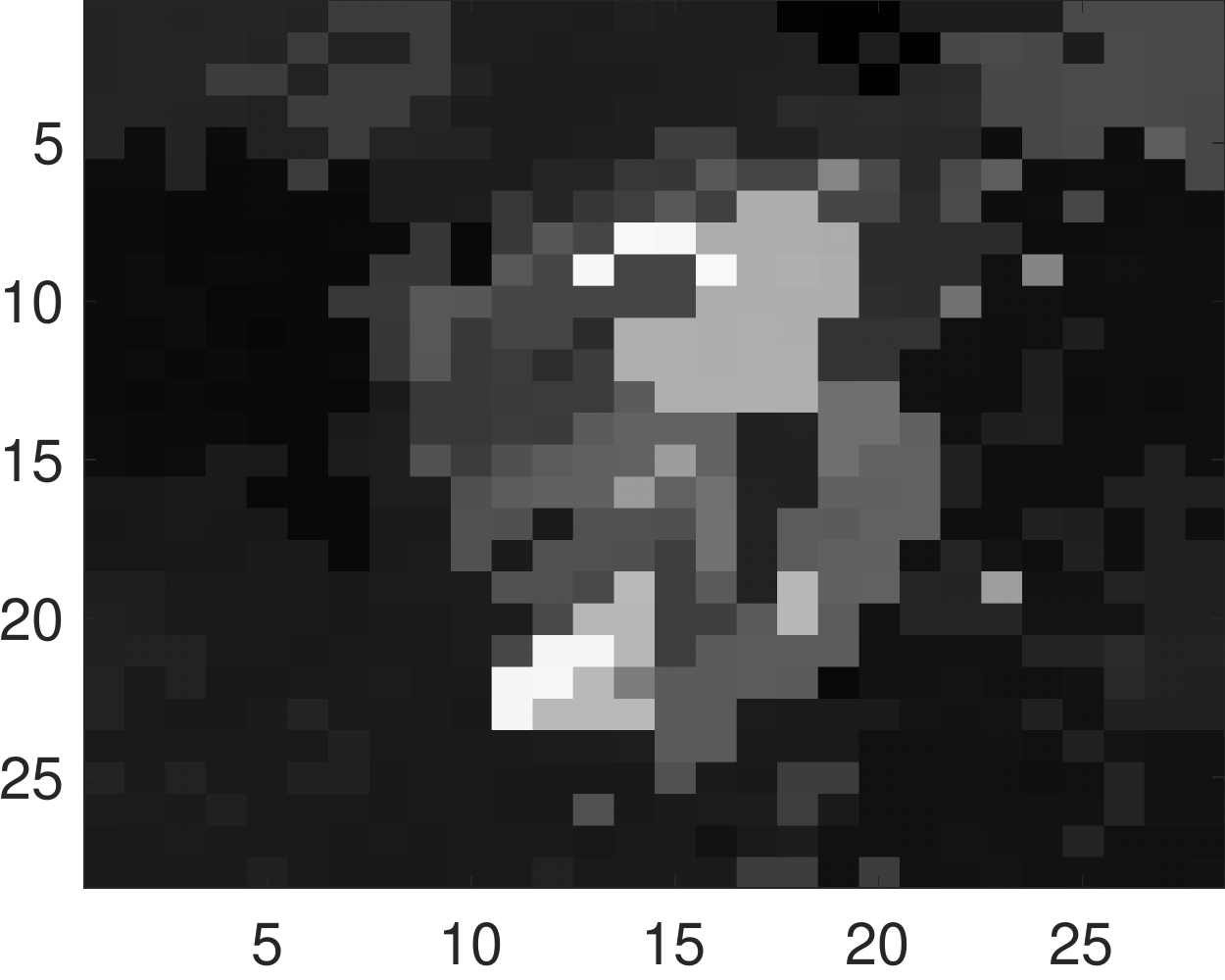}\label{fig:intervalsXoverRand101}}
\subfloat[]{\includegraphics[width=0.25\columnwidth]{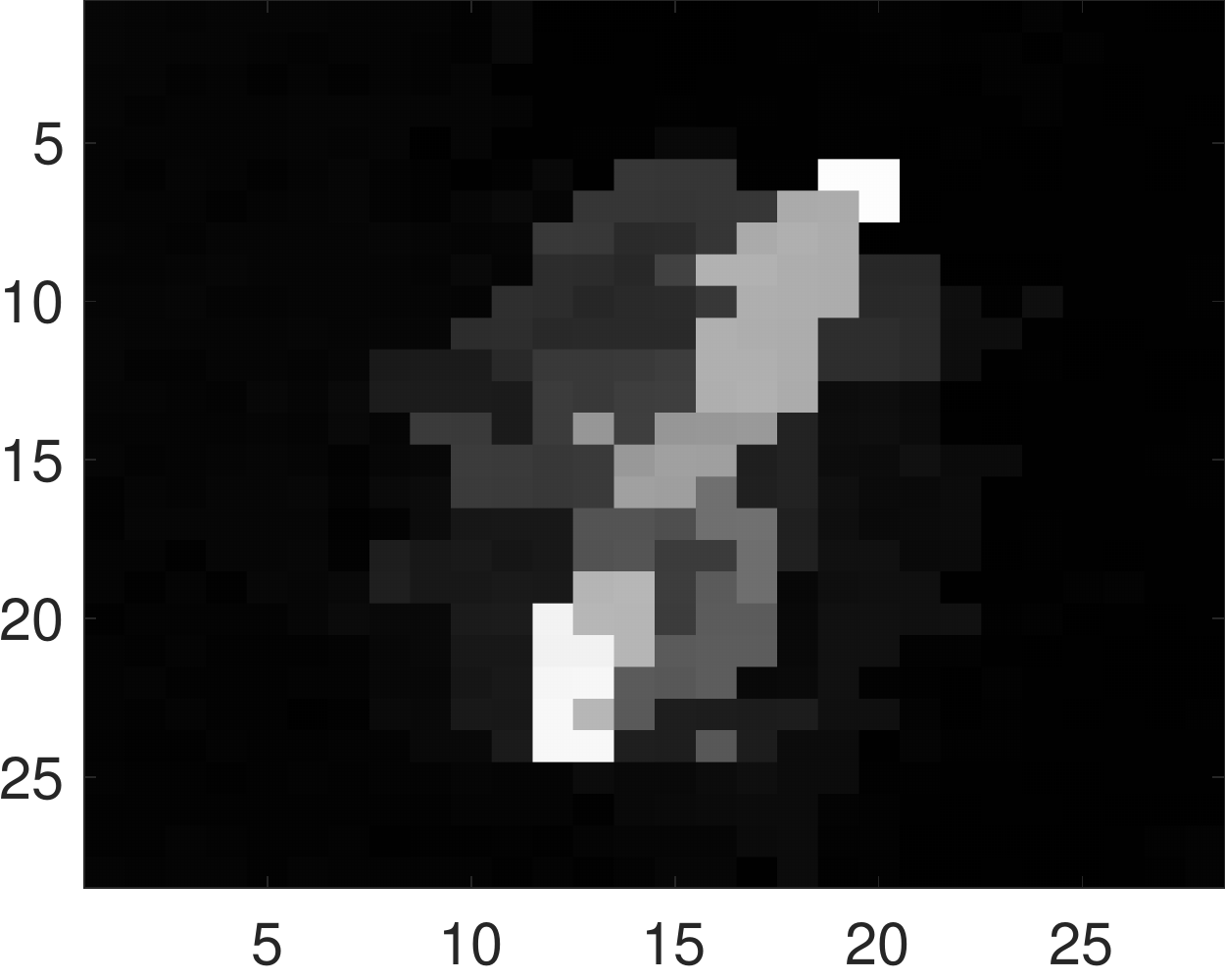}\label{fig:intervalsXoverRand1000}}
\subfloat[]{\includegraphics[width=0.25\columnwidth]{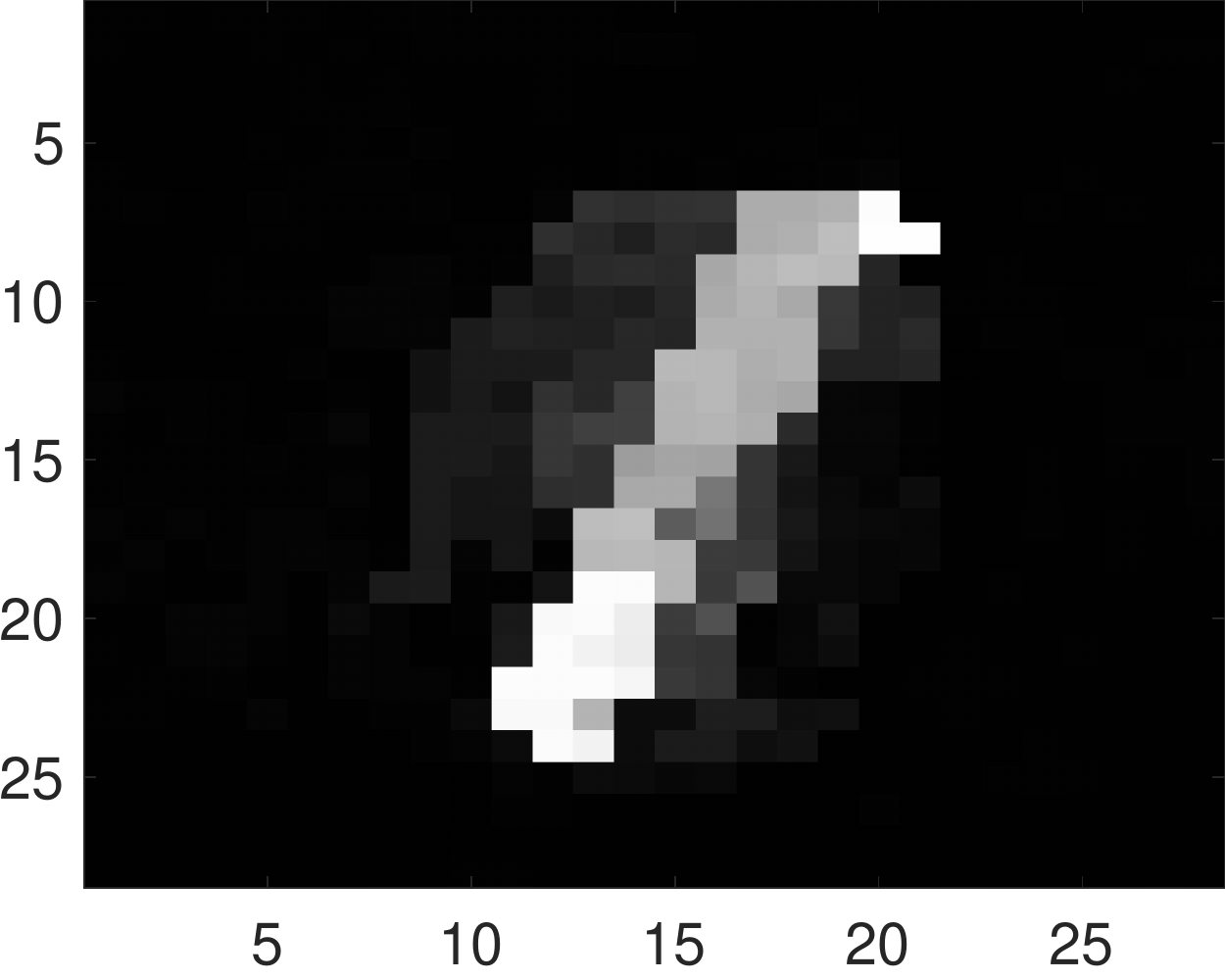}\label{fig:intervalsXoverRand10000}}
\subfloat[]{\includegraphics[width=0.25\columnwidth]{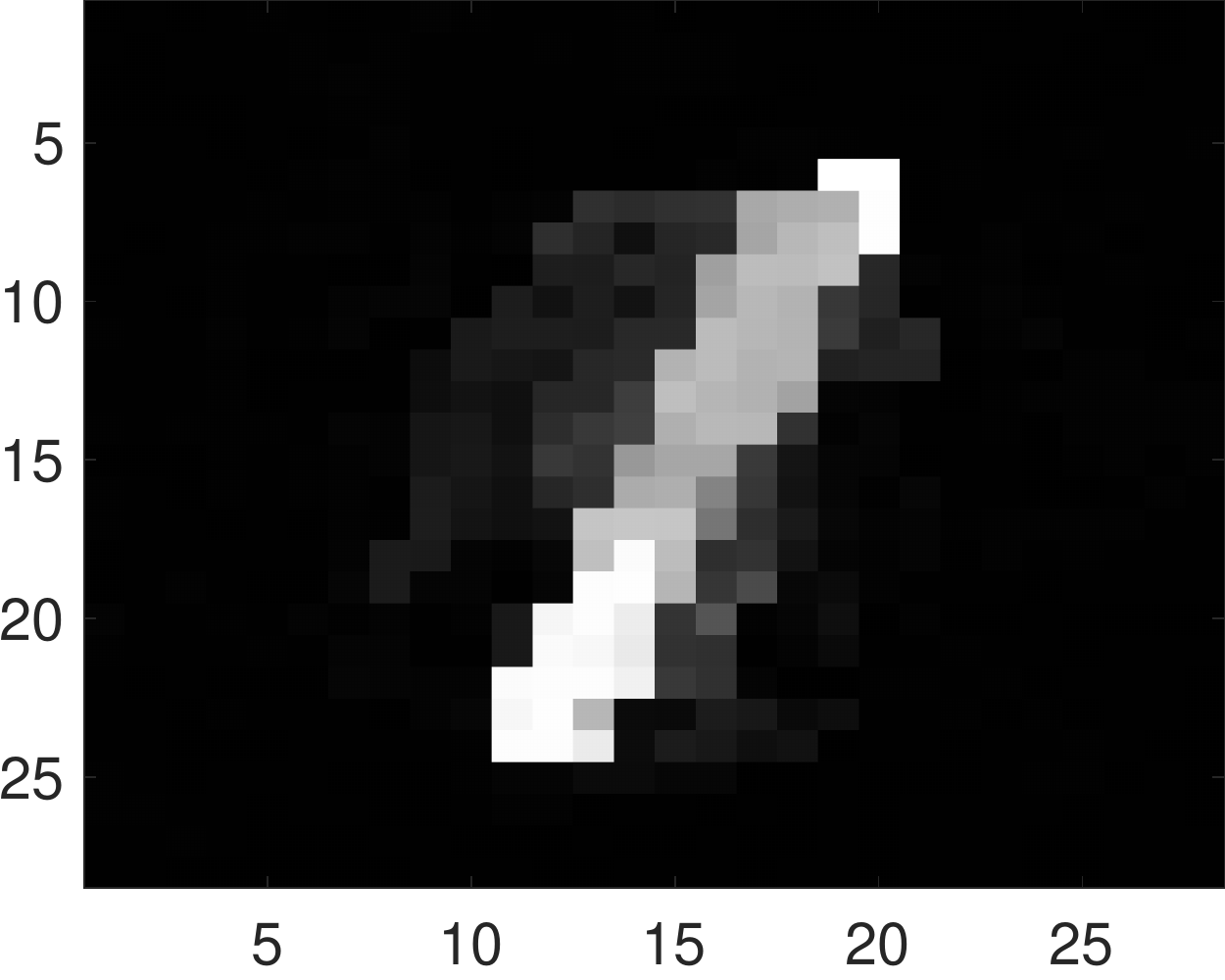}\label{fig:intervalsXoverRand20000}}
\end{subfigures}
\caption{Imitation problem ($28\times28$ scenario): Visualization of the evolutionary process at generations $g \in \{100, 1000, 10000, 20000\}$ for each row from left to right (ground truth shown in Figure~\ref{fig:MNISTGroundTruth1}). The rows present the results of HillClimbing, CopyBest and XoverRand ($mr=0.001$) respectively.} \label{fig:imitationVisualizationProcess}
\end{figure}

Figure~\ref{fig:imitationVisualizationProcess} shows a visualization of the results obtained by HillClimbing, CopyBest and XoverRandCP05 (with $mr=0.001$) during the evolutionary process\footnote{Video of the evolutionary process available at: \url{https://youtu.be/DfjSvKA6KNI}.}. We observe that different versions of the algorithm show different behaviors. For instance, with HillClimbing the reconstructed image appears very noisy: whereas some agents appear to imitate well the ground truth, there are many isolated agents that are not optimized. This is expected since each agent is trying to optimize its own local fitness function without any parameter exchange. On the other hand, in the case of CopyBest and XoverRand we observe that the images are much smoother and less noisy. Also, CopyBest seems unable to optimize the agents in the middle of the image: this is due to the fact that those agents have much larger differences (in terms of their optimal parameters) w.r.t. their neighbors. Therefore, copying the neighbors provides little or no benefit.

% We have also investigated the effect of $cp$ and $cr$, which affect how frequently crossover occurs, and how many parameters are exchanged (see~\ref{app:imitation}). Again, we observe that in general XoverRand performs better than XoverBest. Also, the results indicate a better performance when crossover is performed frequently (higher $cp$) but exchanging a small number of components (lower $cr$). In particular, the best result is achieved by XoverRandCP1CR005, which performs crossover with a randomly selected neighbor exchanging a very small number of components (the expected number of exchanged components is $0.005\times100 = 5$).

%---------------------------------------------

\subsubsection{$7\times7$ scenario}

We tested the five Embodied Evolution algorithms with the same parameter settings tested for the $28\times28$ scenario. Figure~\ref{fig:fitnessTrendMNIST7x7} shows the average (across $10$ runs per algorithm) collective fitness ($F_g$) trends obtained with the tested algorithms and the corresponding std. dev. In the case of CopyBest, CopyRand, XoverBest and XoverRand, again we show the fitness trend obtained with the best parameter setting according to the Nemenyi test, see \ref{app:stats}.

Similarly to the $28\times28$ scenario, the versions of the algorithm that do not employ crossover perform worse than those that use it. However in this case the difference between XoverRand and XoverBest is much larger, with XoverBest obtaining similar results to those of the versions without crossover. This confirms our previous hypothesis, since in this case it is even more unlikely that the best agent in the neighborhood of one agent (which has more parameters) might provide a benefit during crossover. In any case, these results show that even if the optimal parameters of the neighbor agents are quite different (as we have shown in Figure~\ref{fig:compareAgentParameterDistances}), sharing behavior parameters with randomly selected neighbors helps the optimization process.

\begin{figure}[ht!]
\begin{subfigures}
\subfloat[]{\includegraphics[width=0.25\columnwidth]{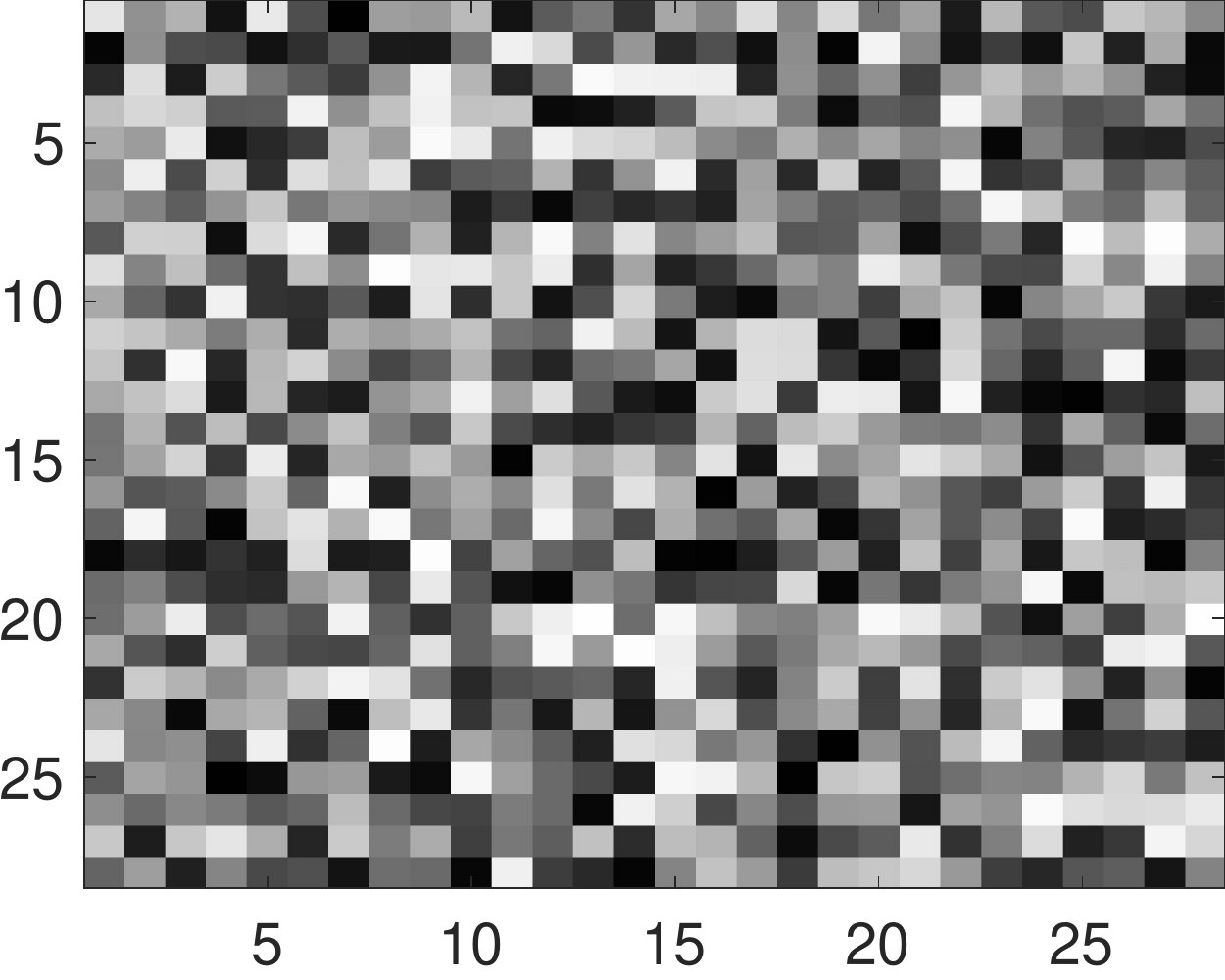}\label{fig:intervals7x7HillClimbing100}}
\subfloat[]{\includegraphics[width=0.25\columnwidth]{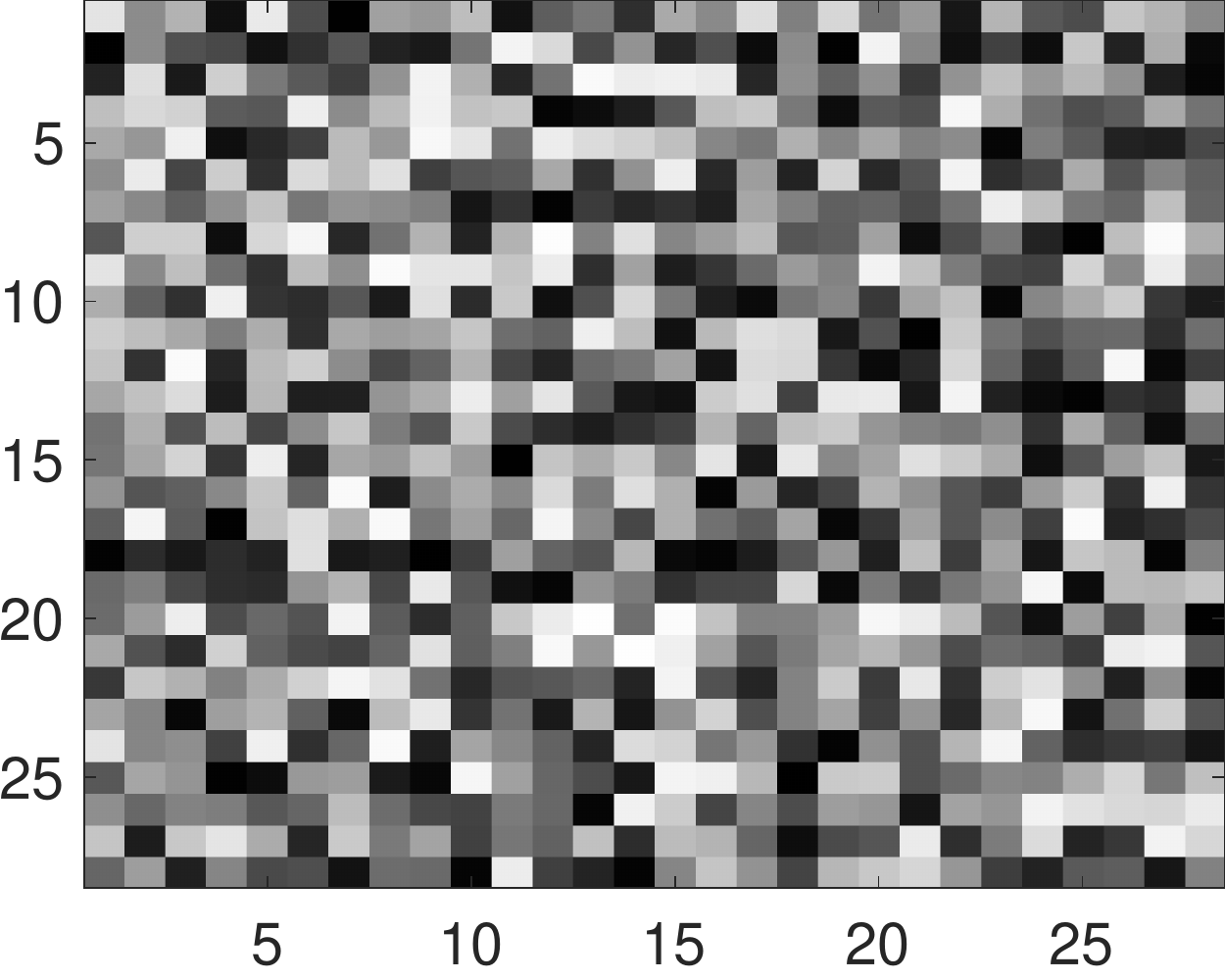}\label{fig:intervals7x7HillClimbing1000}}
\subfloat[]{\includegraphics[width=0.25\columnwidth]{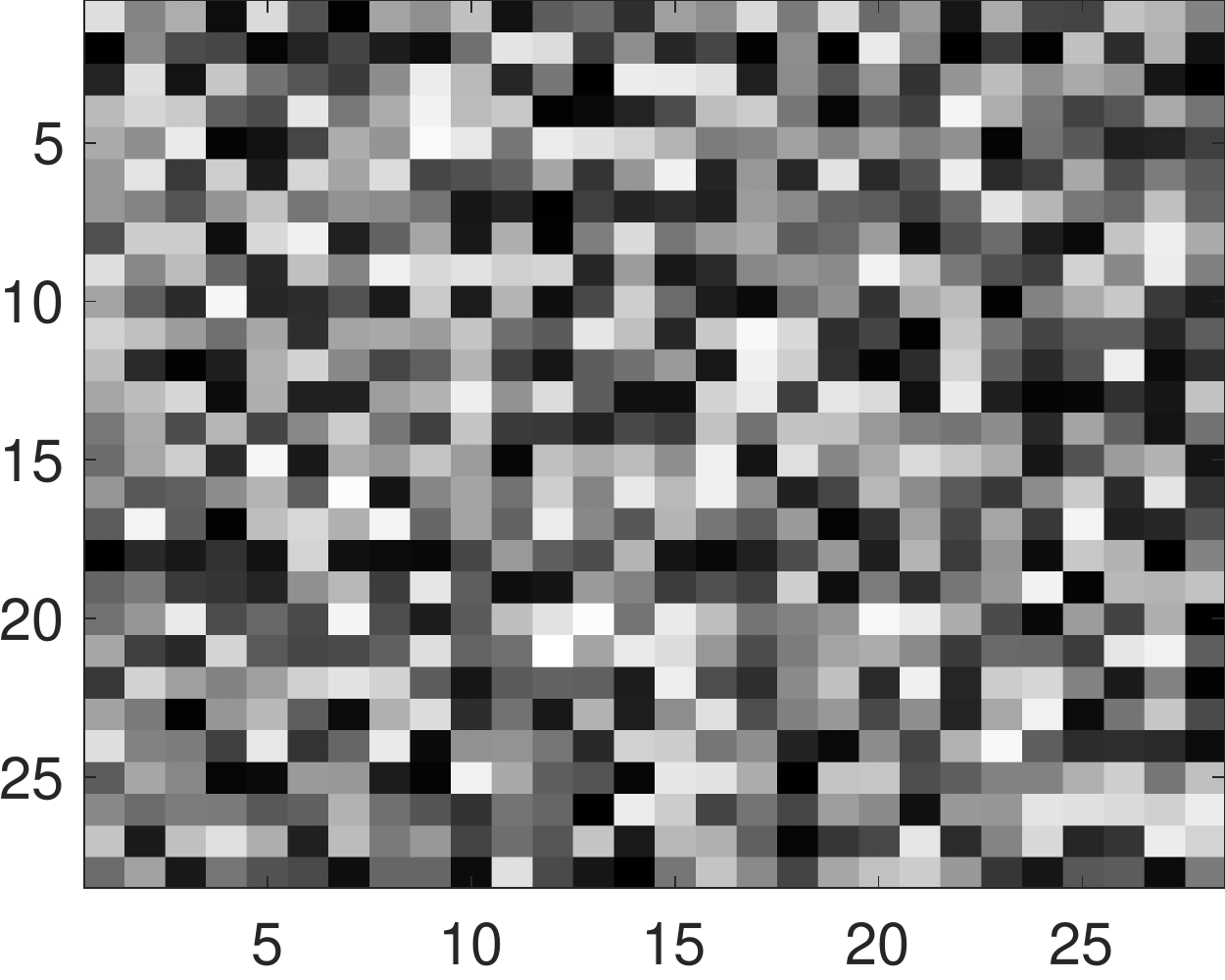}\label{fig:intervals7x7HillClimbing10000}}
\subfloat[]{\includegraphics[width=0.25\columnwidth]{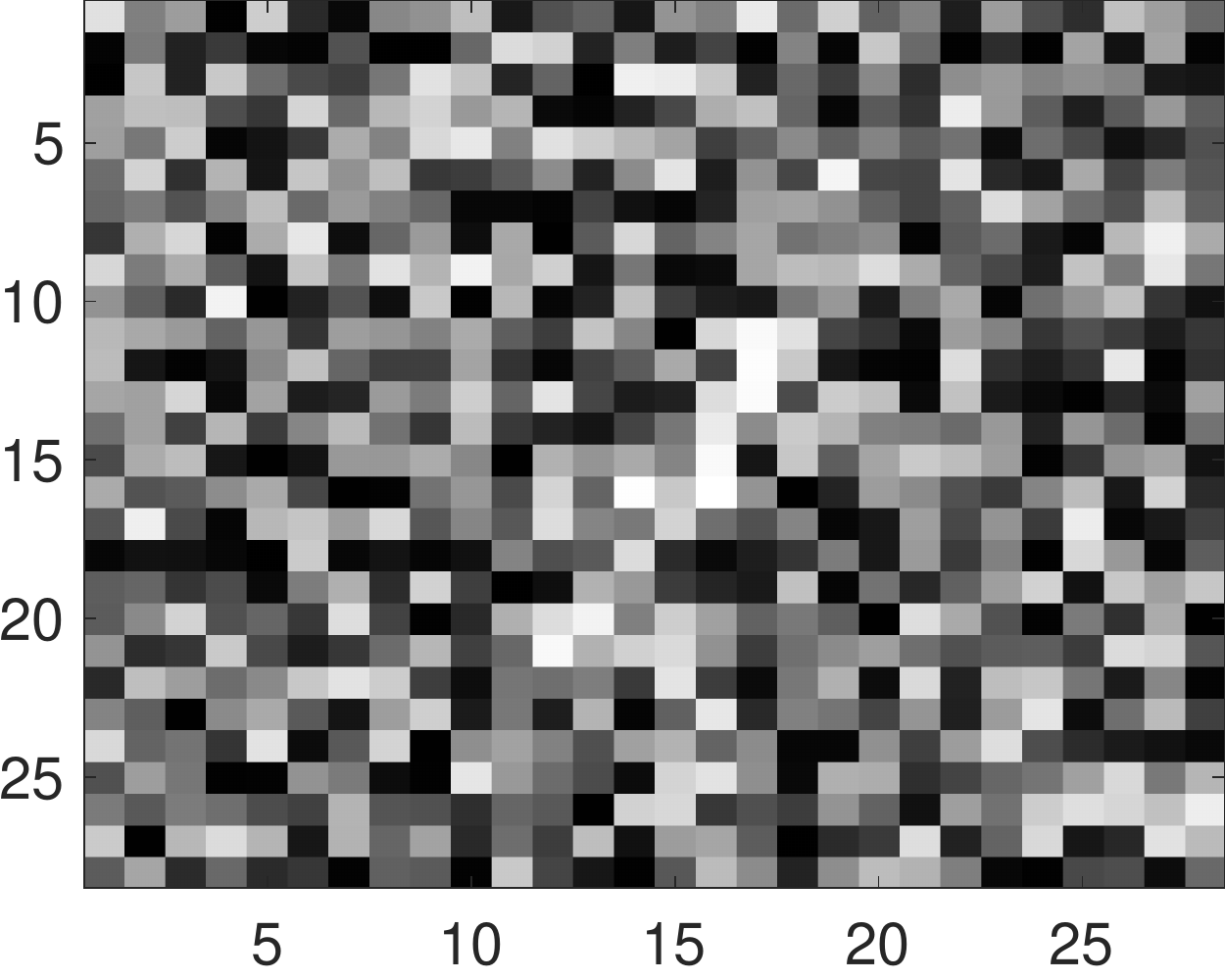}\label{fig:intervals7x7HillClimbing100000}}

\subfloat[]{\includegraphics[width=0.25\columnwidth]{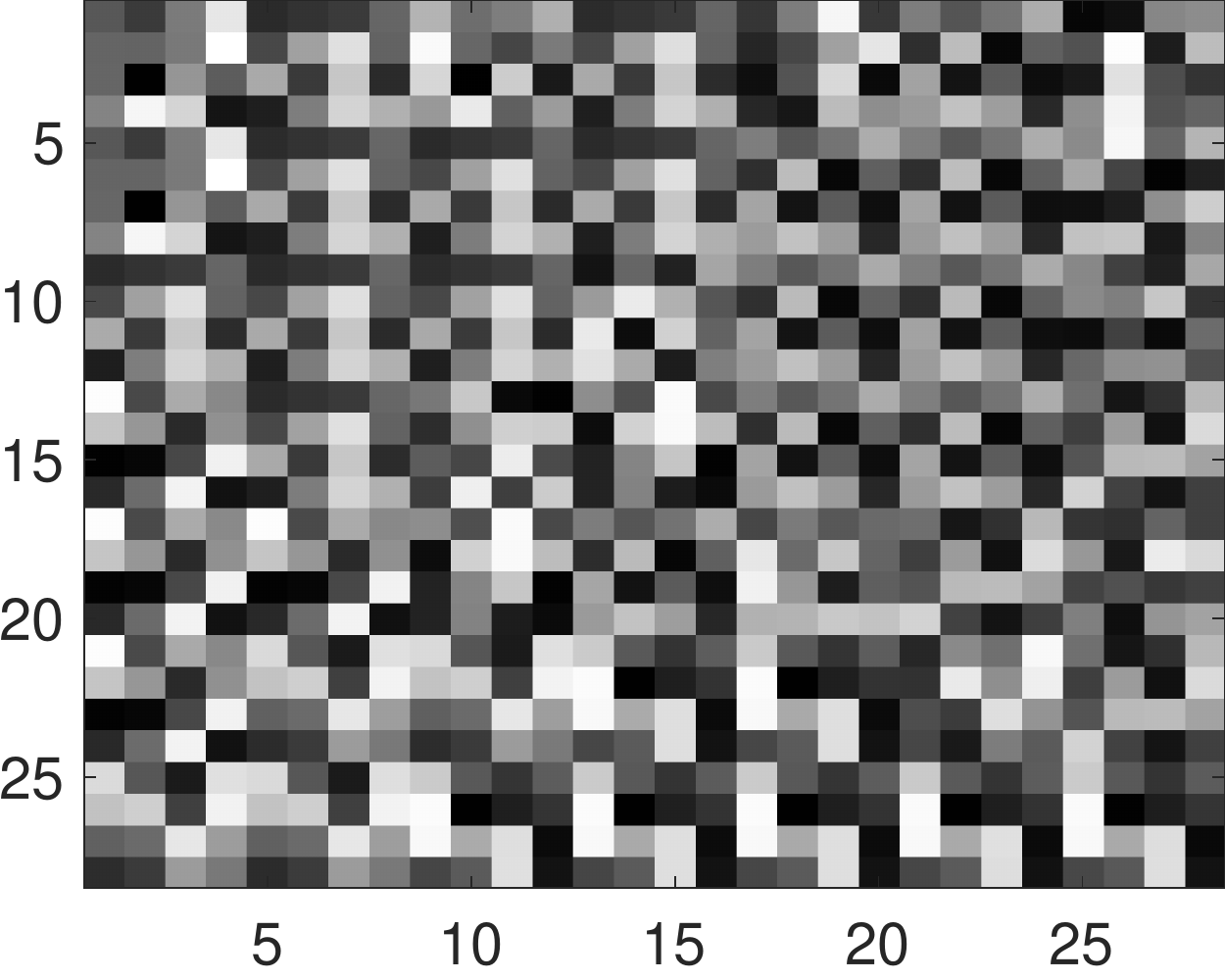}\label{fig:intervals7x7CopyBest100}}
\subfloat[]{\includegraphics[width=0.25\columnwidth]{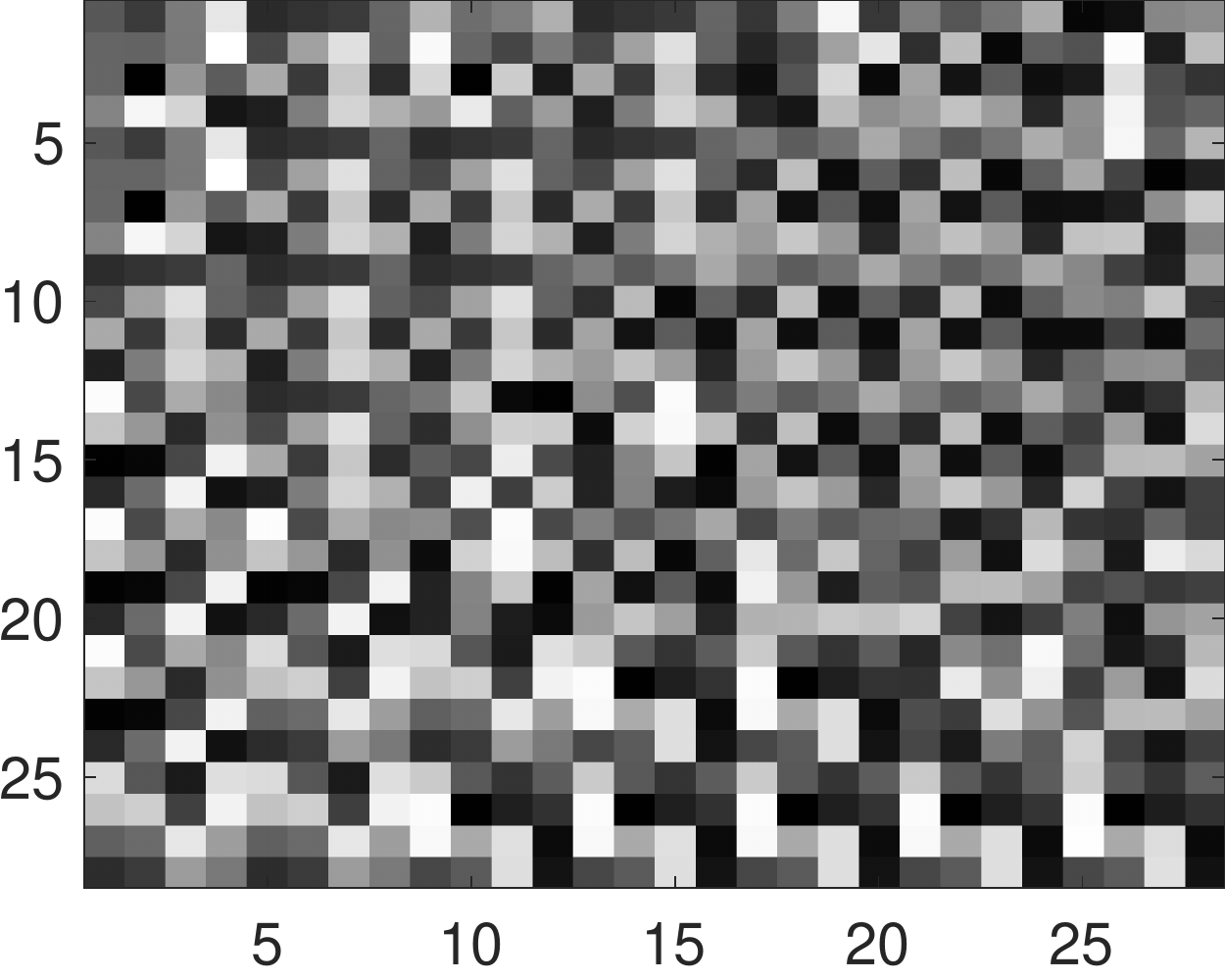}\label{fig:intervals7x7CopyBest1000}}
\subfloat[]{\includegraphics[width=0.25\columnwidth]{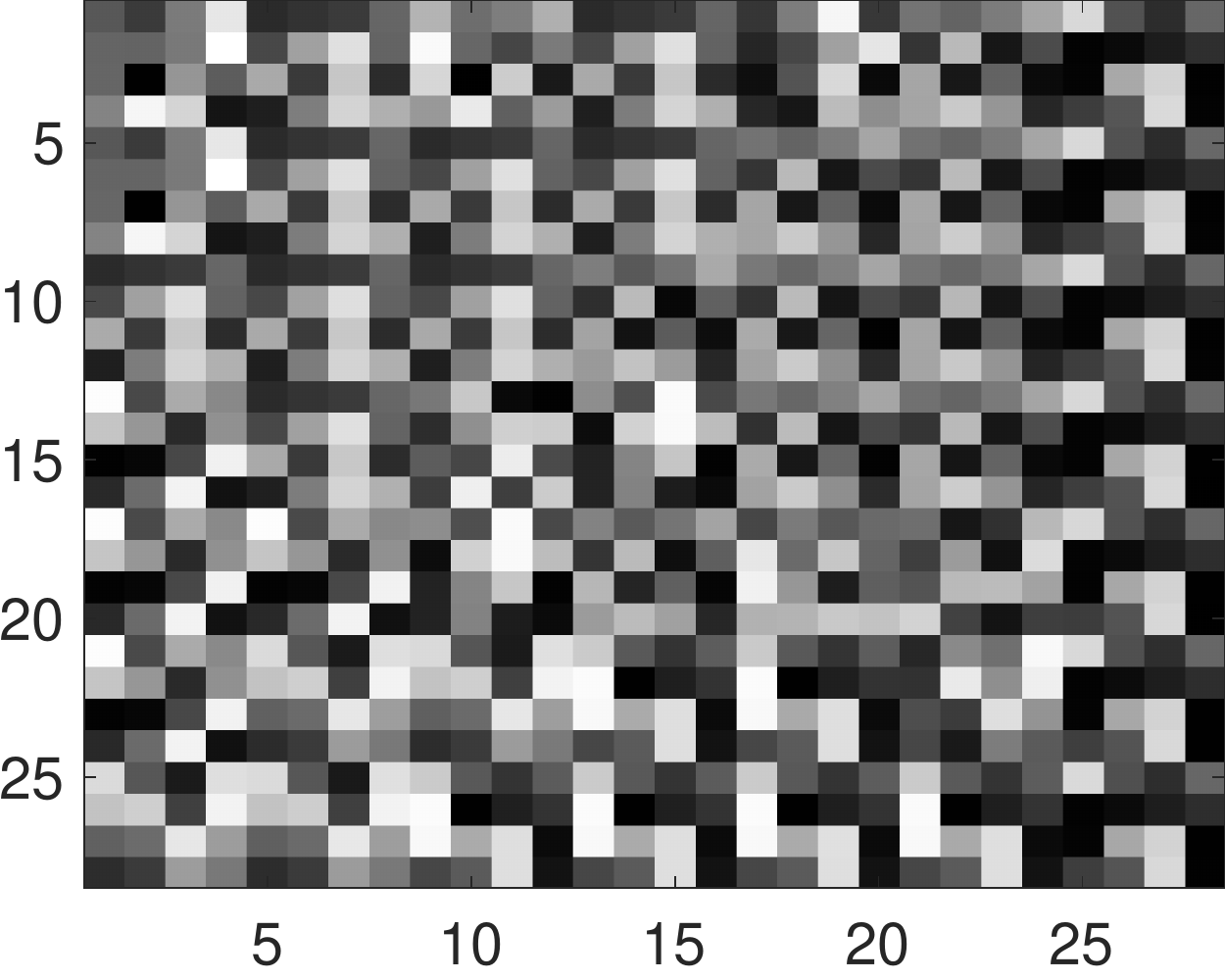}\label{fig:intervals7x7CopyBest10000}}
\subfloat[]{\includegraphics[width=0.25\columnwidth]{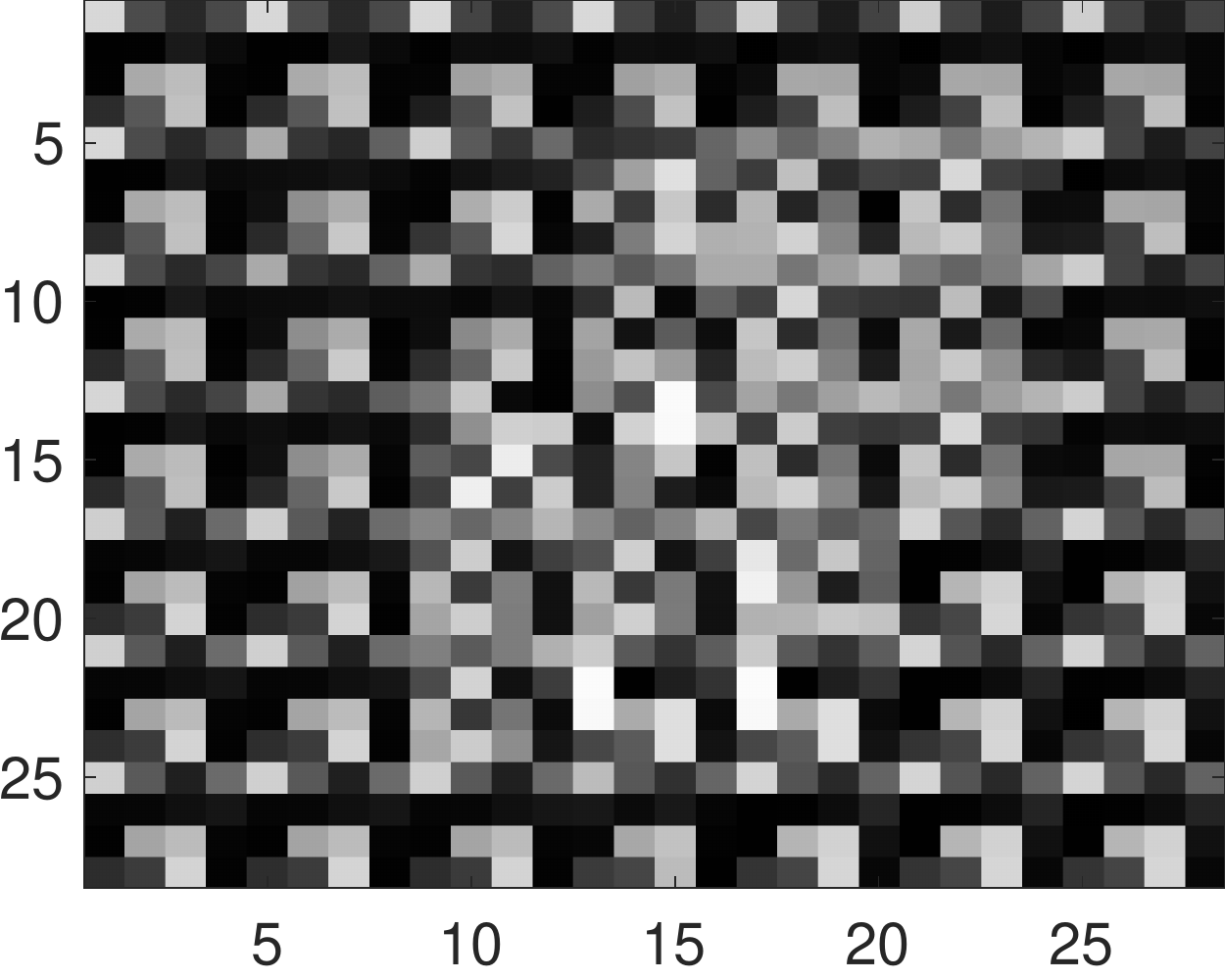}\label{fig:intervals7x7CopyBest100000}}

\subfloat[]{\includegraphics[width=0.25\columnwidth]{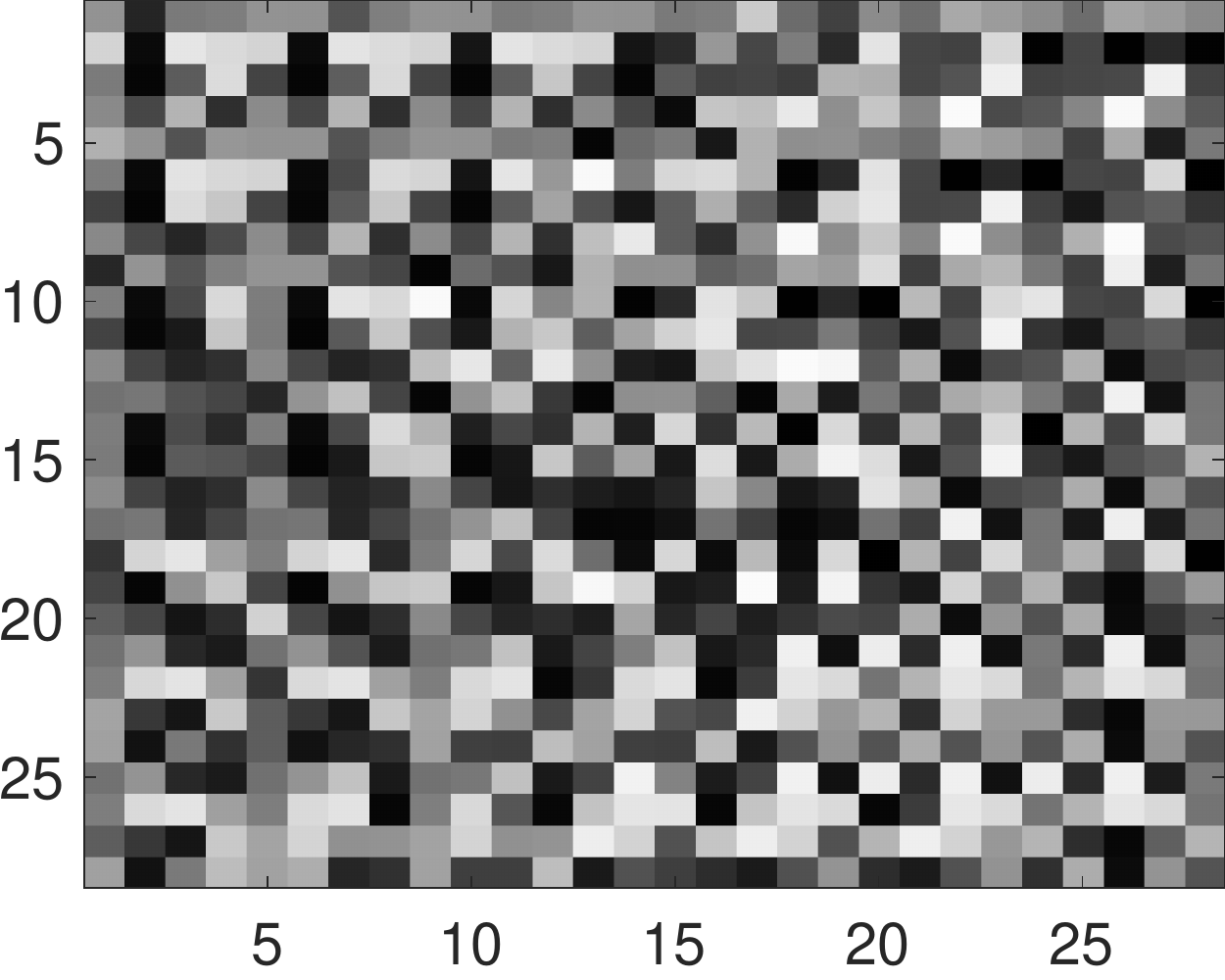}\label{fig:intervals7x7XoverRand100}}
\subfloat[]{\includegraphics[width=0.25\columnwidth]{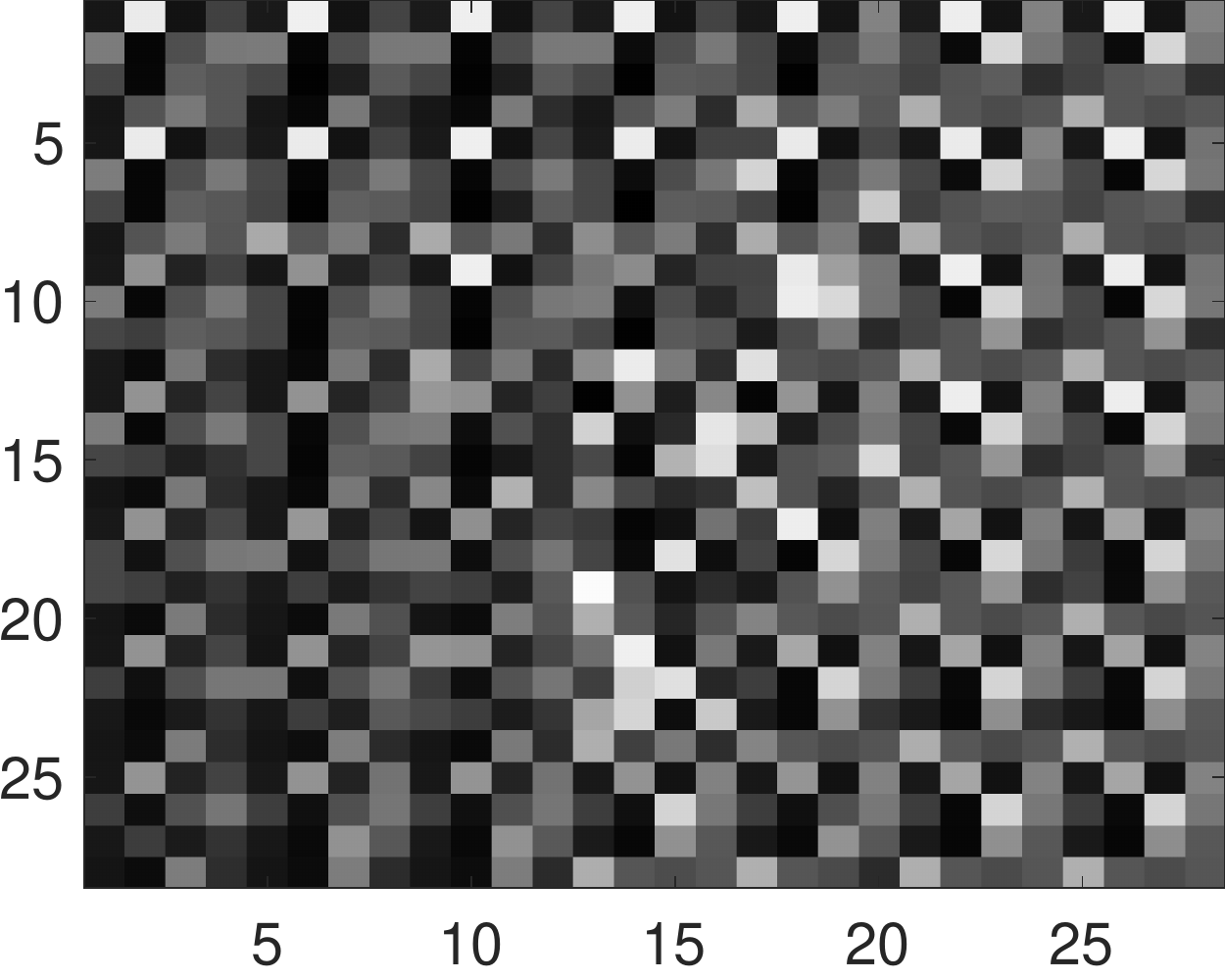}\label{fig:intervals7x7XoverRand1000}}
\subfloat[]{\includegraphics[width=0.25\columnwidth]{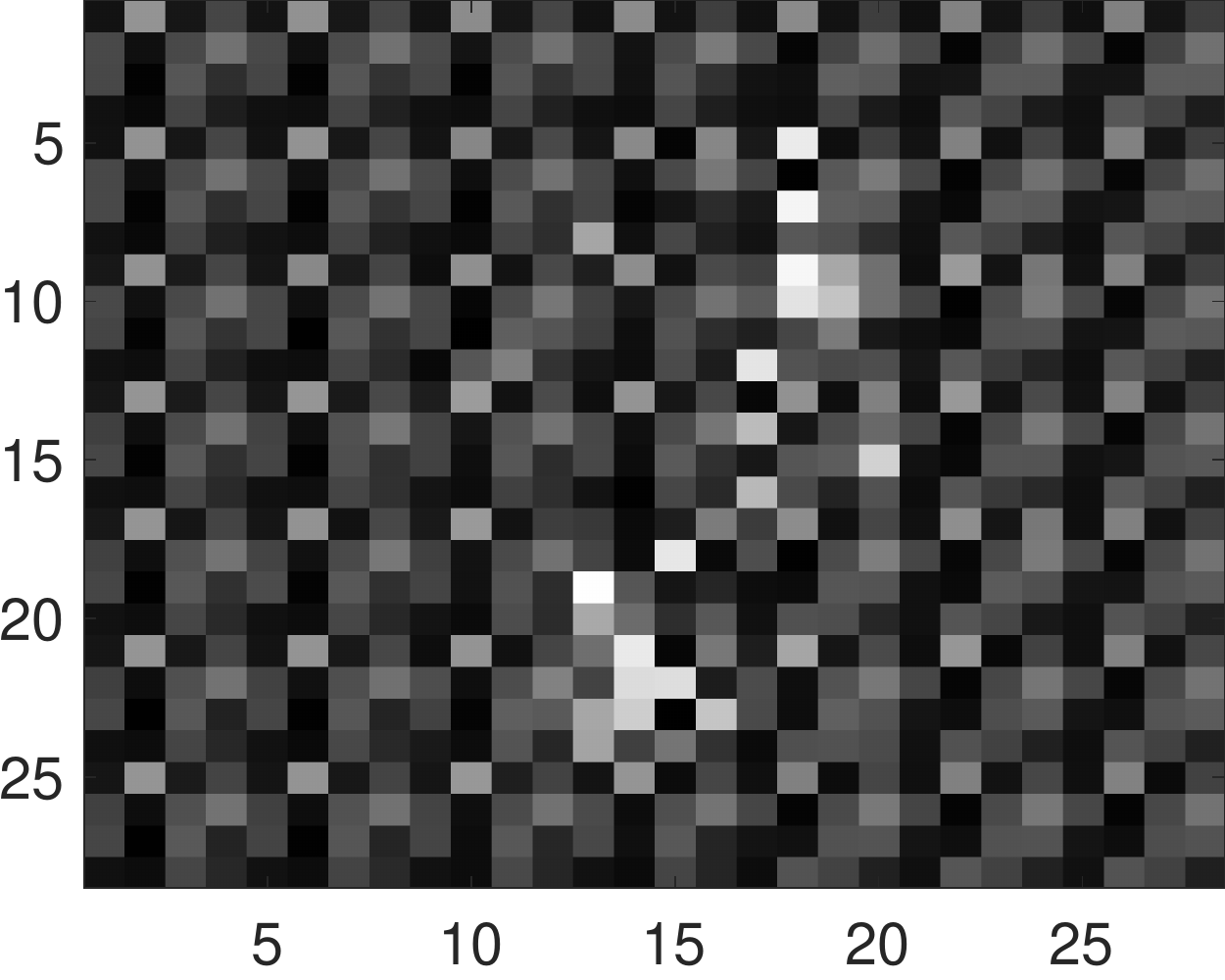}\label{fig:intervals7x7XoverRand10000}}
\subfloat[]{\includegraphics[width=0.25\columnwidth]{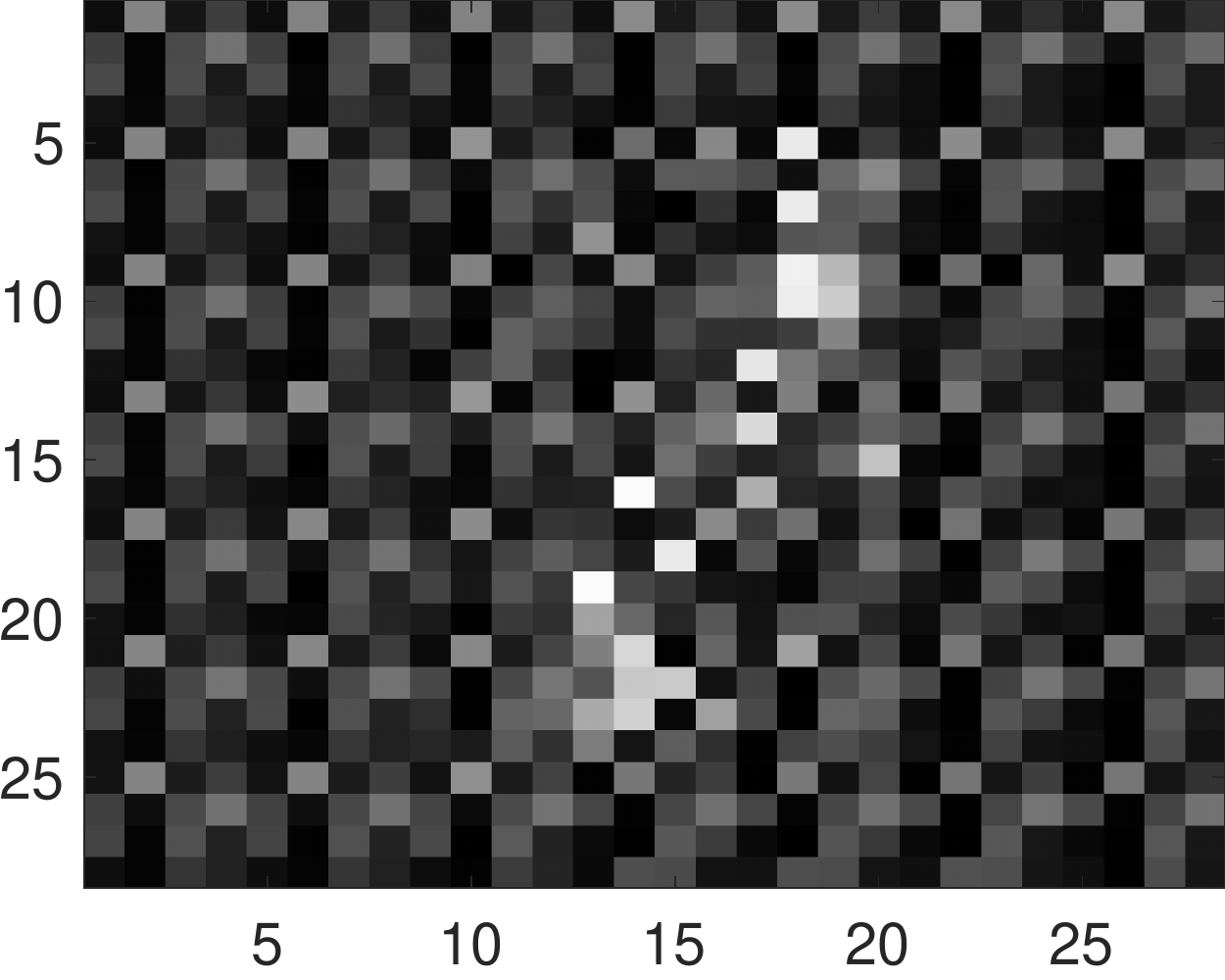}\label{fig:intervals7x7XoverRand100000}}
\end{subfigures}
\caption{Imitation problem ($7\times 7$ scenario): Visualization of the evolutionary process at generations $g \in \{100, 1000, 10000, 100000\}$ for each row from left to right (ground truth shown in Figure~\ref{fig:MNISTGroundTruth1}). The rows present the results of HillClimbing, CopyBest and XoverRand ($mr=0.001$) respectively.} \label{fig:imitationVisualization7x7}
\end{figure}

In Figure~\ref{fig:imitationVisualization7x7}, we show the visualization of the results obtained by HillClimbing, CopyBest and XoverRandCP02 (with $mr=0.001$) at generations $g=\{100, 1000, 10000, 100000\}$: also here we observe that HillClimbing produces a much noisier image, while XoverRandCP02 performs the best. 

%---------------------------------------------

\subsection{Illumination problem}
In the following, we analyze separately the numerical results on the illumination problem in the two scenarios described above: agents controlled by a single parameter, and agents controlled by a vector of parameters.

\subsubsection{Single parameter scenario} 
As seen earlier, in this scenario each agent aims to optimize a parameter corresponding to the phase shift of the sine function used for the ground truth, see Equations~\eqref{eq:illuminationEquation} and~\eqref{eq:illuminationEquationAgent}.

We tested again the five Embodied Evolution algorithms, with $cp \in \{0.2, 0.5, 1.0\}$ and $mr \in \{0.005, 0.05, 0.5\}$ (in this case we use arithmetic crossover, so $cr$ is not needed). Figure~\ref{fig:fitnessTrendsIlluminationSingle} shows the average (across $10$ runs per algorithm) collective fitness ($F_g$) trends obtained with the tested algorithms, and the corresponding std. dev. Also in this case, for CopyBest, CopyRand, XoverBest and XoverRand we show the fitness trend obtained with the best parameter setting according to the Nemenyi test, see \ref{app:stats}.

%XoverBestCP02 performs best with $mr=0.05$ and $mr=0.5$, while for the lowest value ($mr = 0.005$)
The main observation is that in this case CopyRand obtains the best results, better than the algorithms with crossover. This might be due to the fact that, with one parameter, mutations cancel out the effect of the arithmetic crossover (which is essentially another kind of mutation), thus producing a slower convergence (or even stagnation) if crossover is applied. So, it is just more efficient to copy a random neighbor (rather than doing crossover) and apply a small mutation.

%NOTE:
%the arithmetic crossover is basically a sort of "large mutation"
%we can rewrite it as (x+y)/2 = x + (y-x)/2 
%and this is the same in all conditions of mr
%now, when mr is 0.5 or 0.05, on top of the mutation due to xover, we have another mutation that is somehow large, and this probably speeds up the convergence
%while in the mr=0.005 case this mutation is to slow (that's why we see the xover curves converging very slowly)
%it seems as if the (y-x)/2 and N(0,sigma) mutations cancel each other
%which doesn't happen in the copyrand case, where we take the parameters of a random neighbor completely + some mutation N(0,sigma)

% In~\ref{app:illumination}, we show a visualization of the results obtained by three selected versions of the algorithm, and we observe again that with no parameter sharing the results are very noisy, while with crossover results are clearly better.

\begin{figure}[!ht]
\begin{subfigures}
\subfloat[]{\includegraphics[width=0.25\columnwidth, height=0.15\columnwidth]{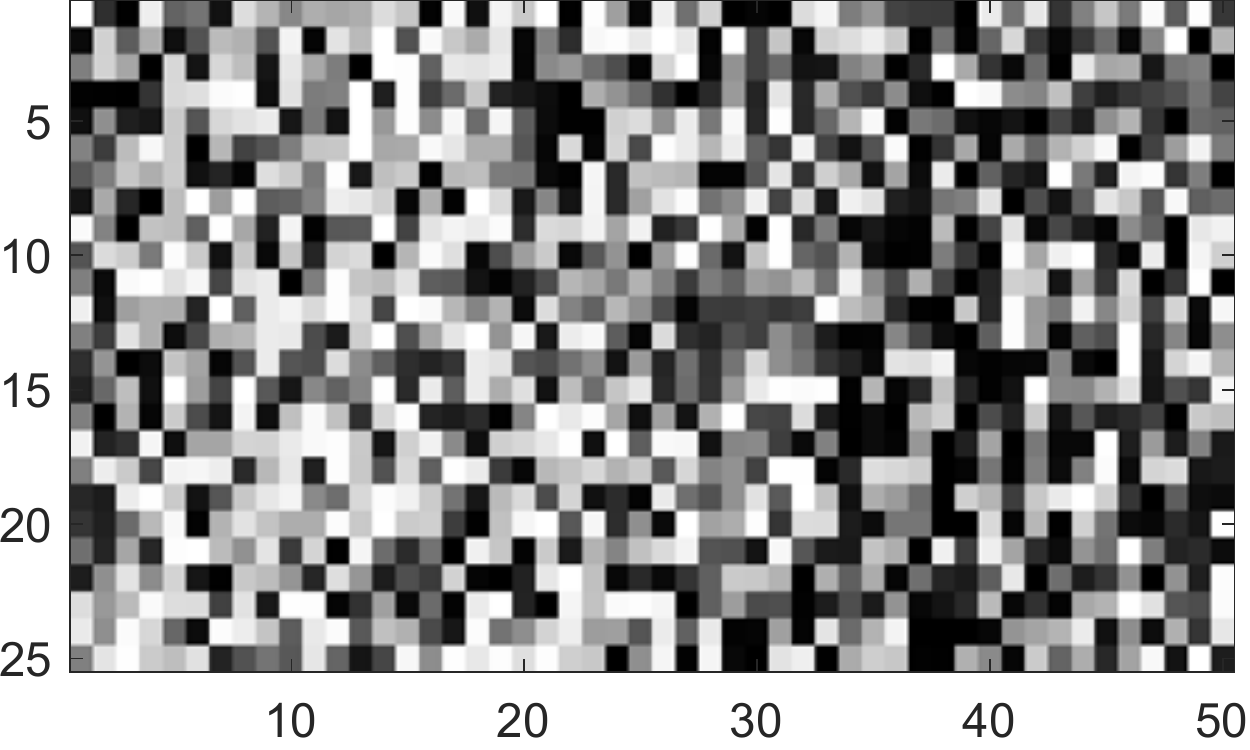}\label{fig:illuminationSingleHillClimbing101}}
\subfloat[]{\includegraphics[width=0.25\columnwidth, height=0.15\columnwidth]{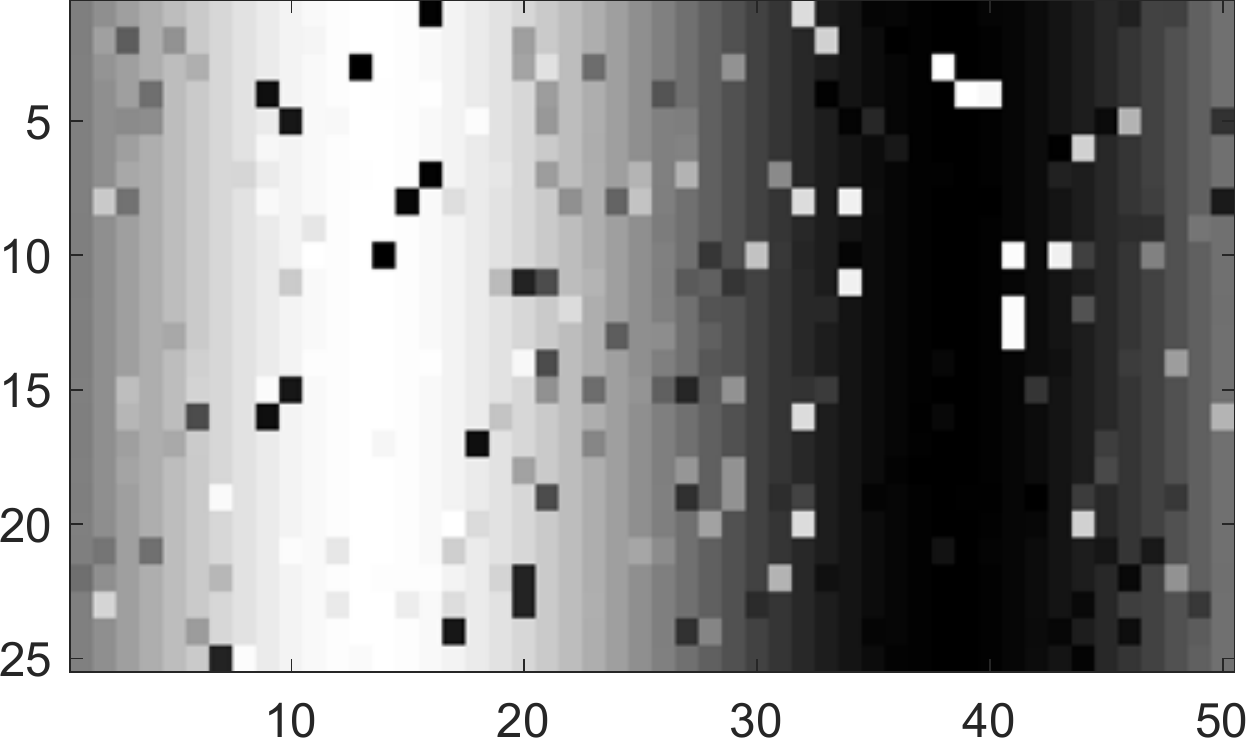}\label{fig:illuminationSingleHillClimbing1000}}
\subfloat[]{\includegraphics[width=0.25\columnwidth, height=0.15\columnwidth]{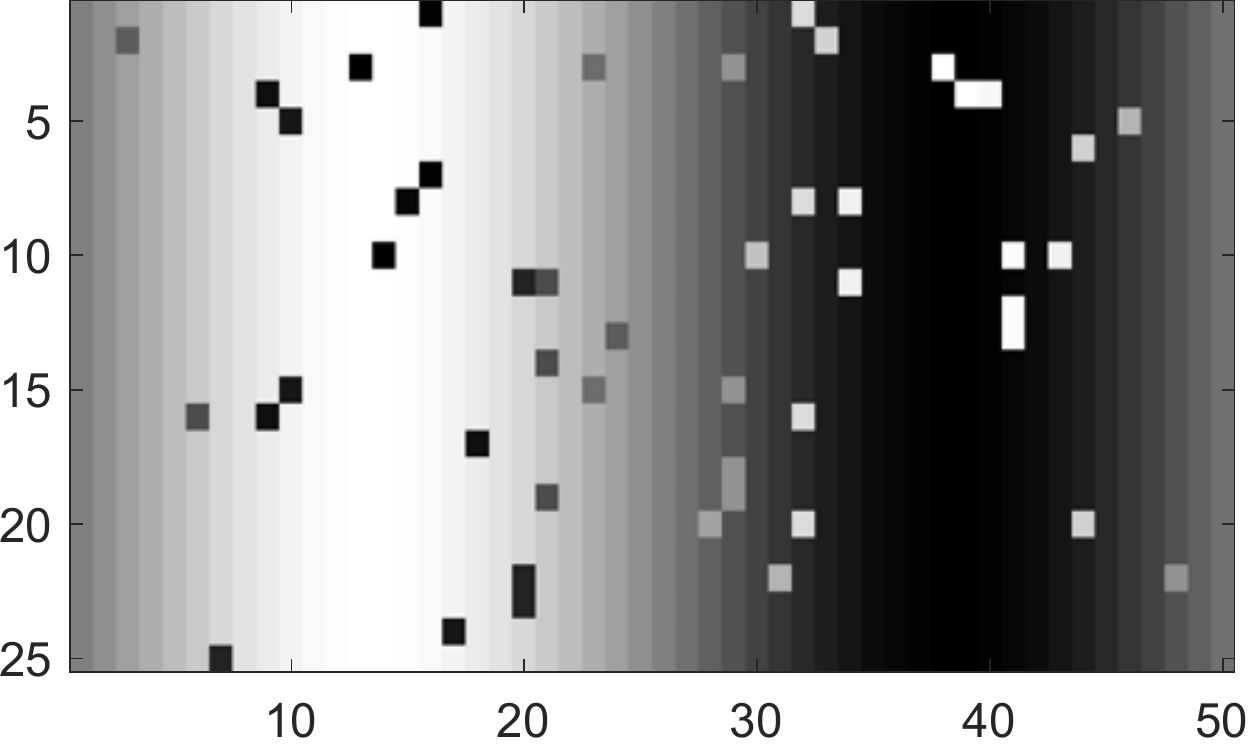}\label{fig:illuminationSingleHillClimbing3000}}
\subfloat[]{\includegraphics[width=0.25\columnwidth, height=0.15\columnwidth]{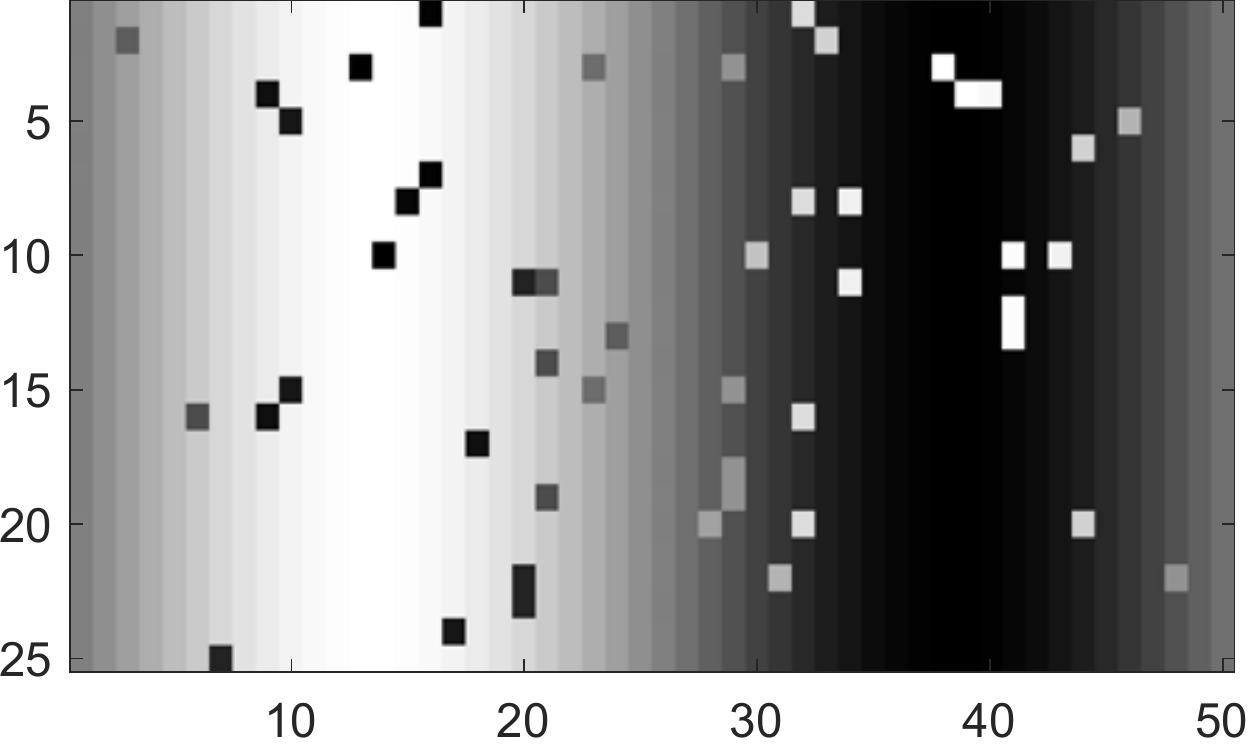}\label{fig:illuminationSingleHillClimbing5000}}

\subfloat[]{\includegraphics[width=0.25\columnwidth, height=0.15\columnwidth]{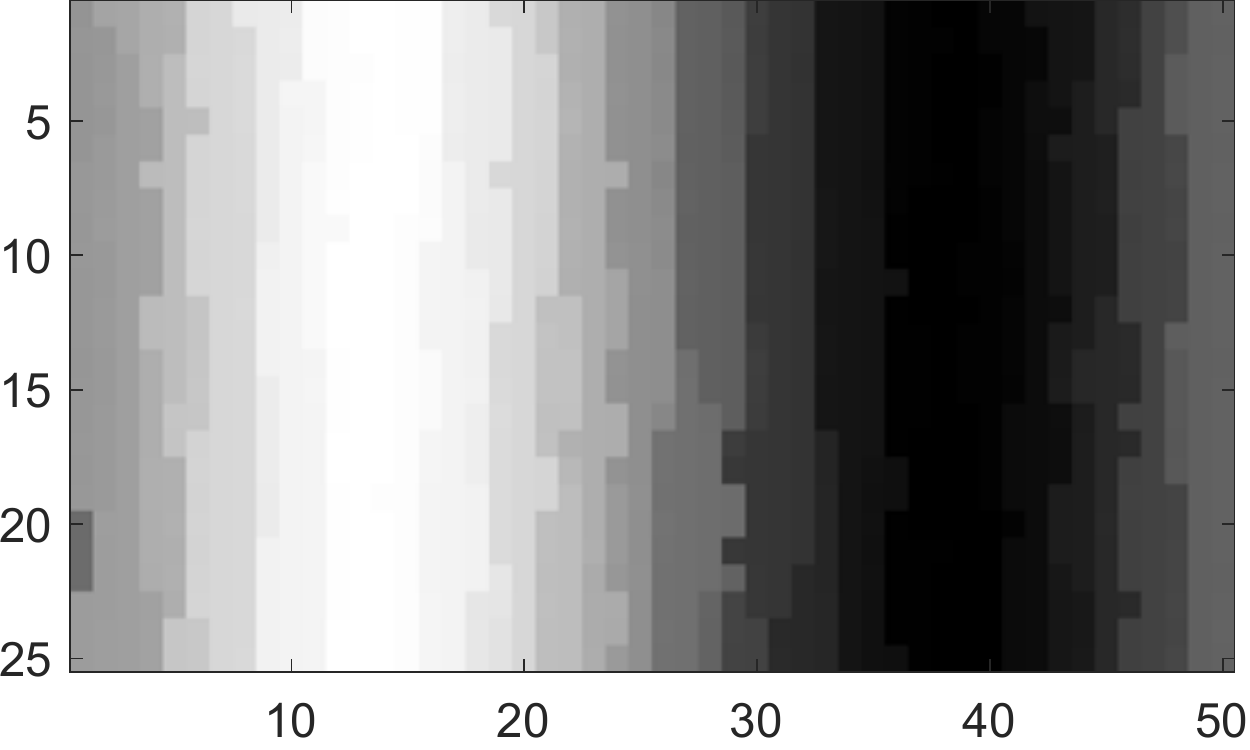}\label{fig:illuminationSingleCopyBest101}}
\subfloat[]{\includegraphics[width=0.25\columnwidth, height=0.15\columnwidth]{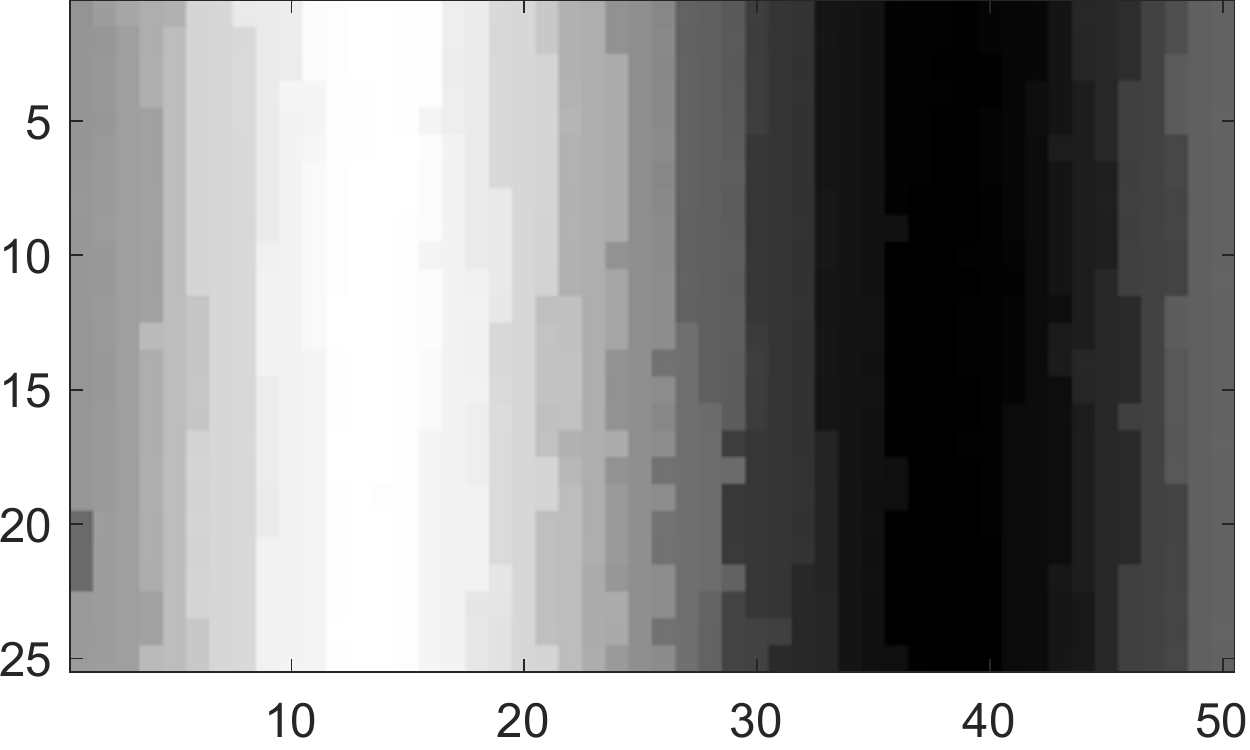}\label{fig:illuminationSingleCopyBest1000}}
\subfloat[]{\includegraphics[width=0.25\columnwidth, height=0.15\columnwidth]{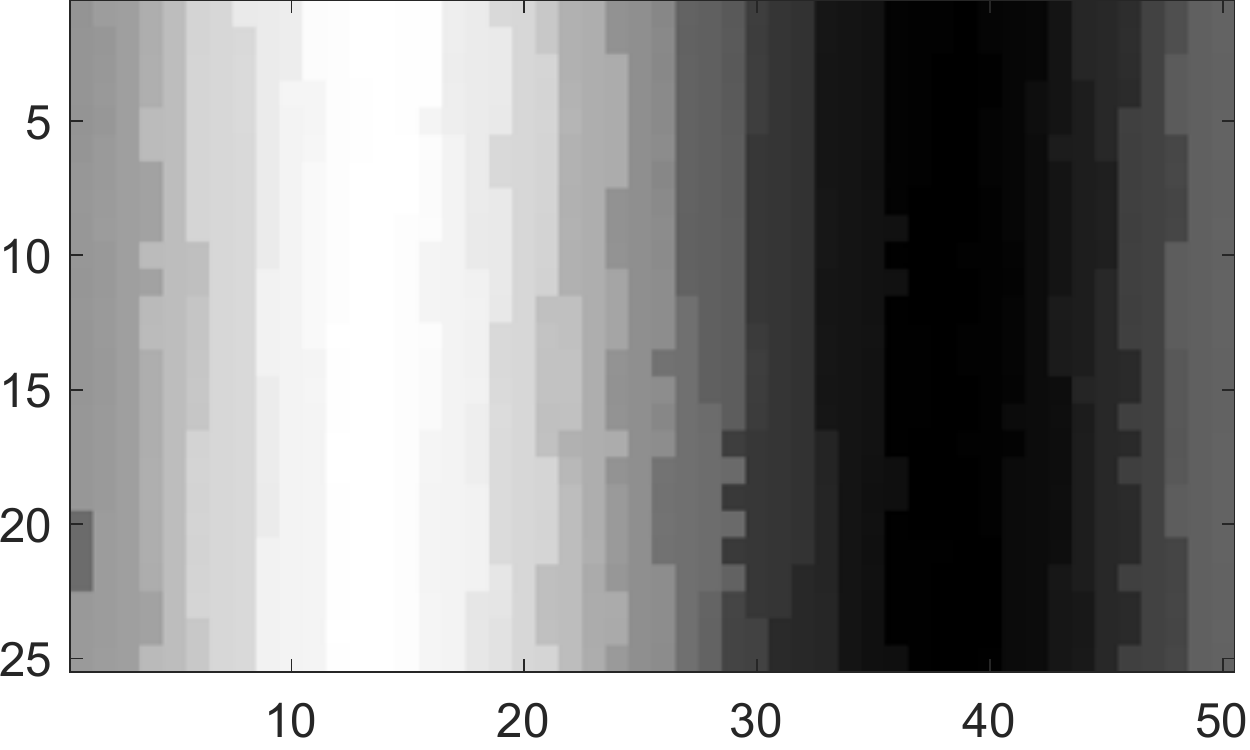}\label{fig:illuminationSingleCopyBest3000}}
\subfloat[]{\includegraphics[width=0.25\columnwidth, height=0.15\columnwidth]{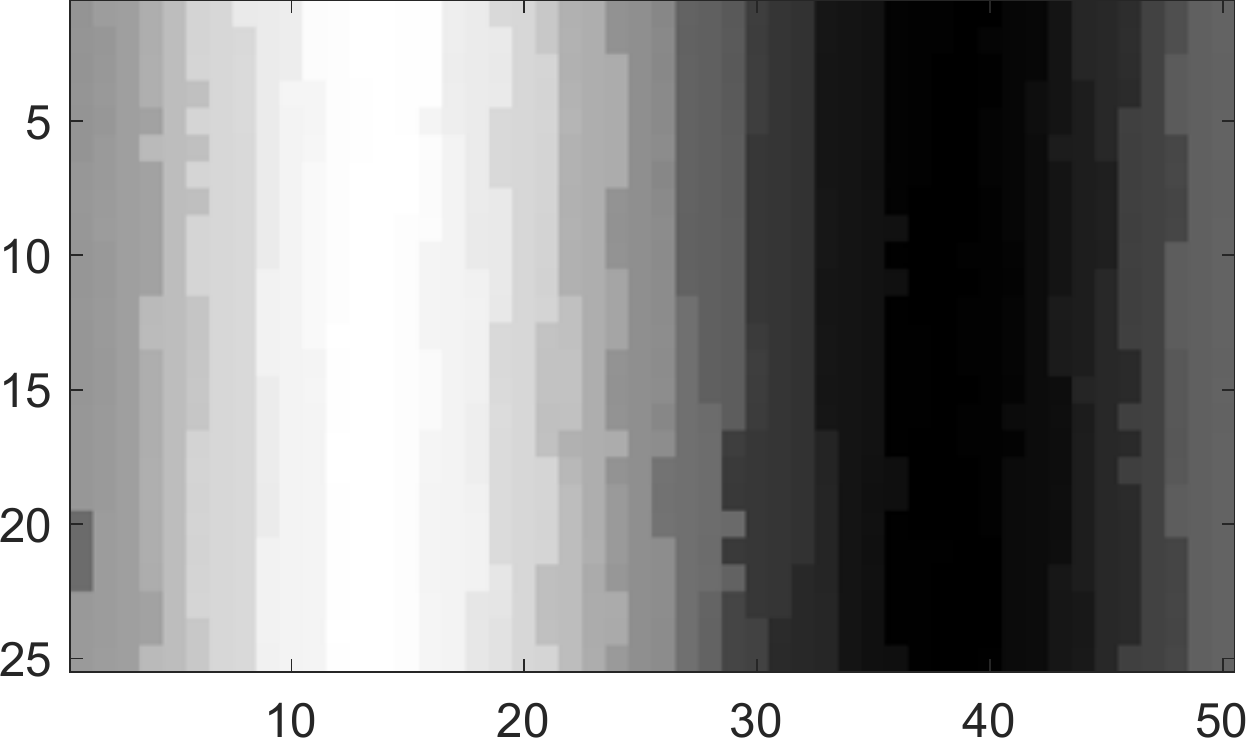}\label{fig:illuminationSingleCopyBest5000}}

\subfloat[]{\includegraphics[width=0.25\columnwidth, height=0.15\columnwidth]{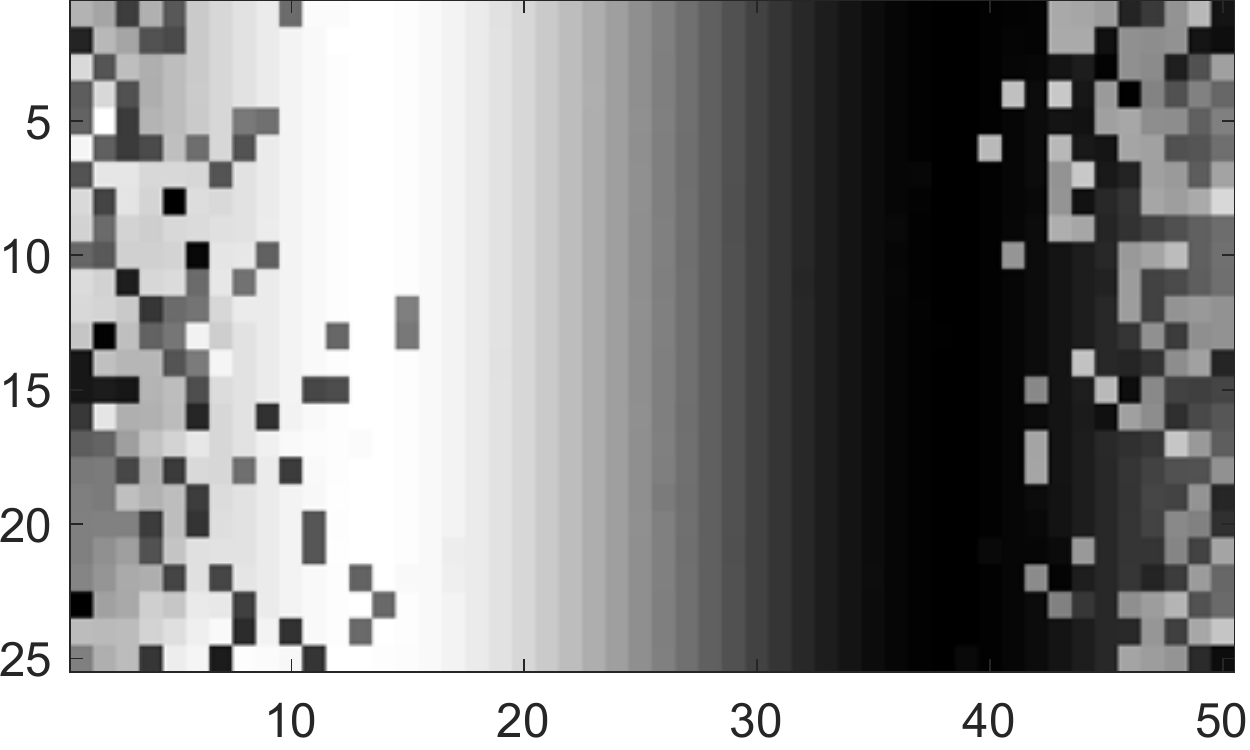}\label{fig:illuminationSingleXoverRand02101}}
\subfloat[]{\includegraphics[width=0.25\columnwidth, height=0.15\columnwidth]{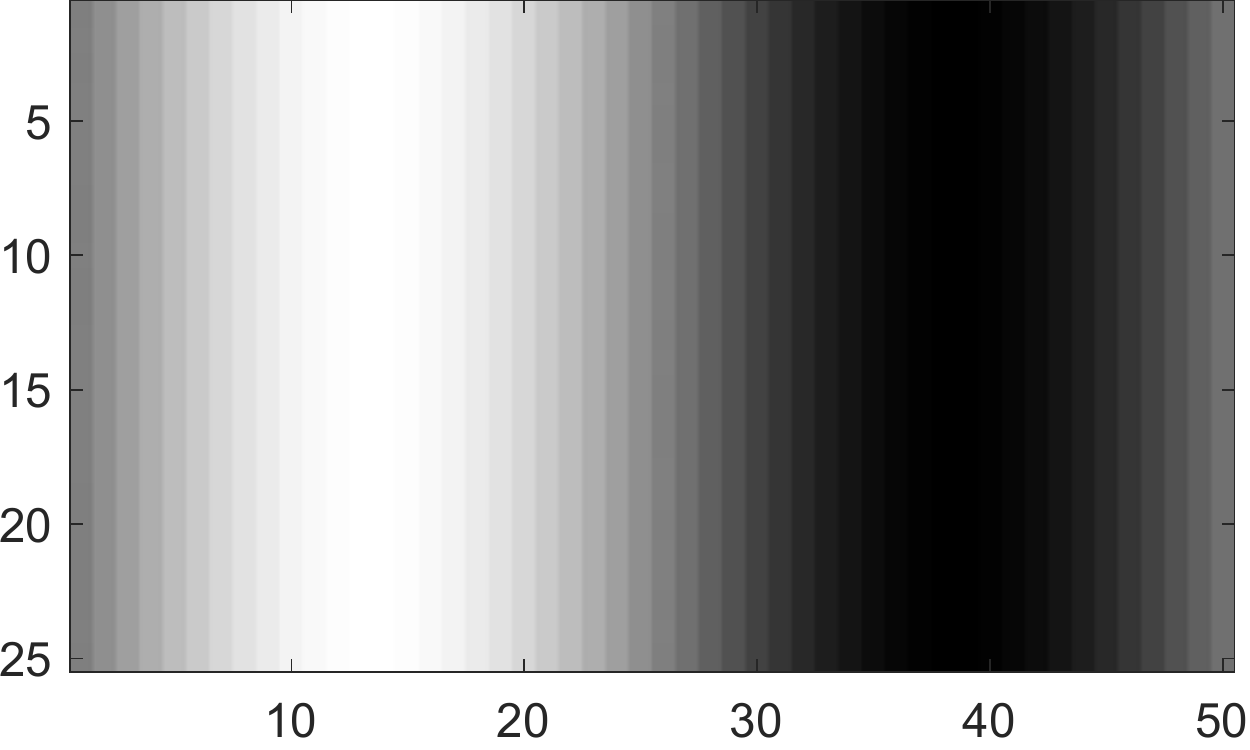}\label{fig:illuminationSingleXoverRand021000}}
\subfloat[]{\includegraphics[width=0.25\columnwidth, height=0.15\columnwidth]{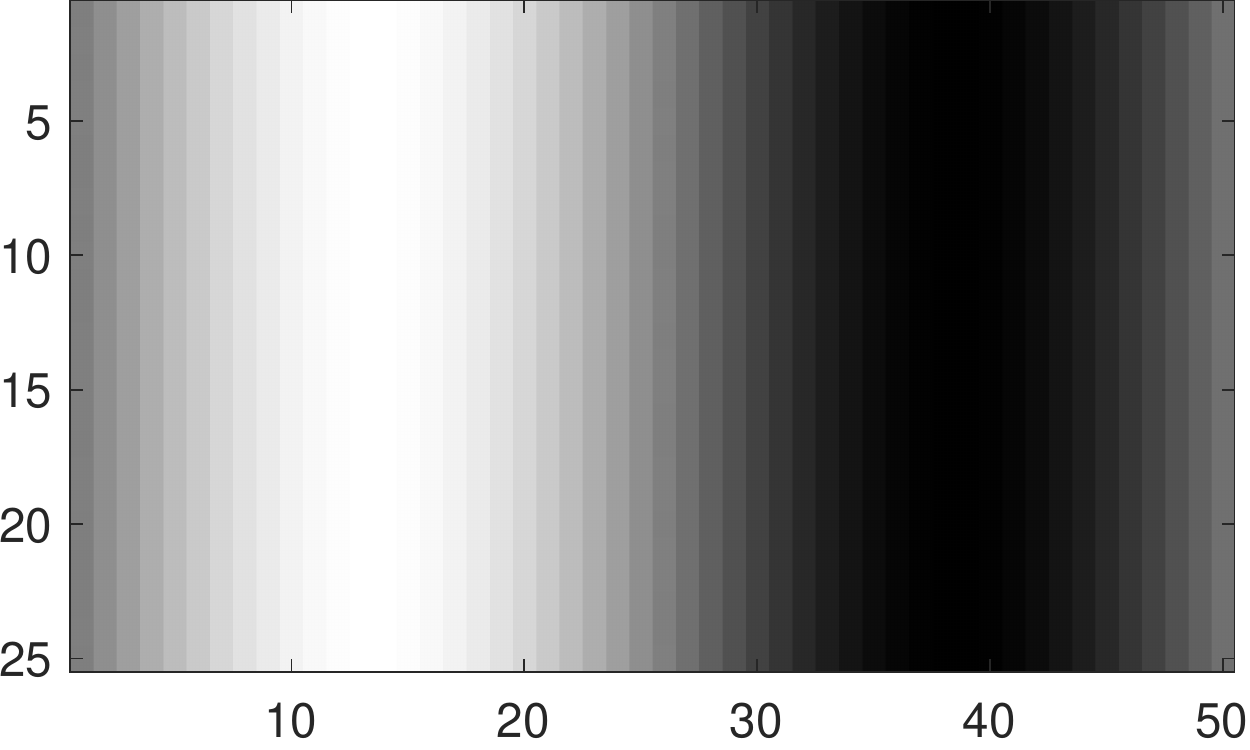}\label{fig:illuminationSingleXoverRand023000}}
\subfloat[]{\includegraphics[width=0.25\columnwidth, height=0.15\columnwidth]{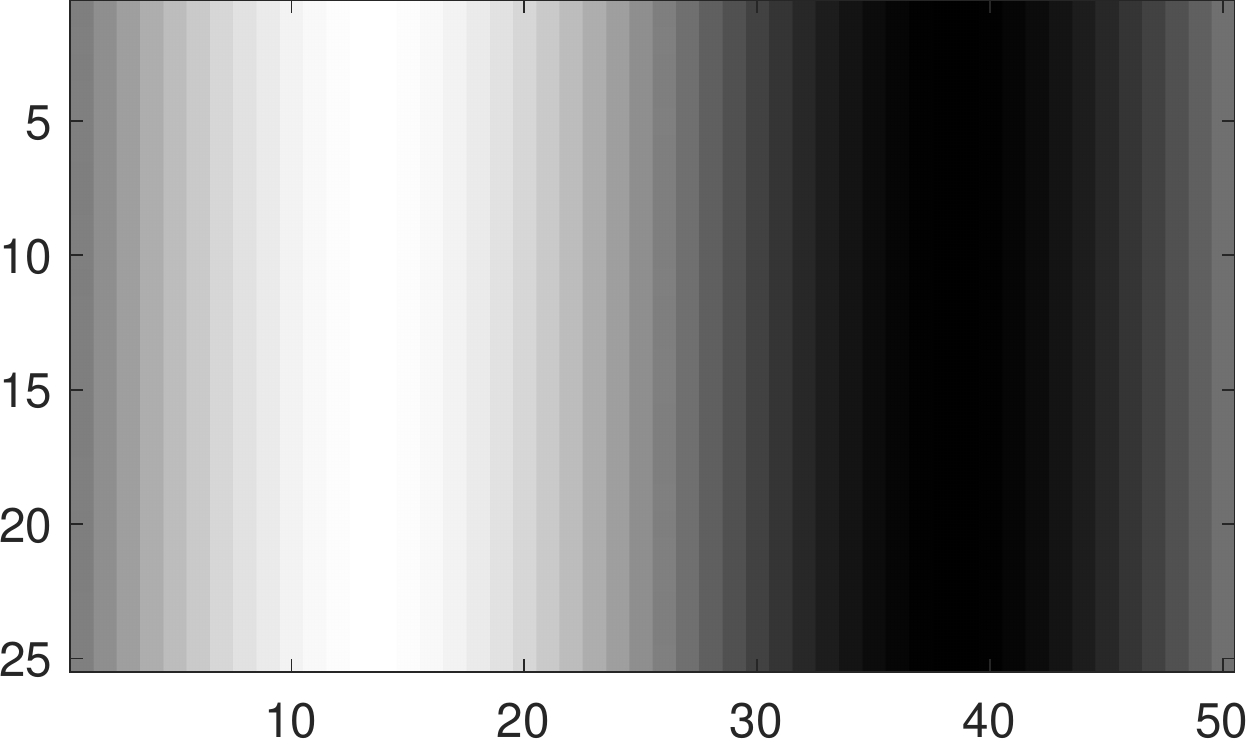}\label{fig:illuminationSingleXoverRand025000}}
\end{subfigures}
\caption{Illumination problem (single parameter scenario): Visualization of the evolutionary process at generations $g \in \{100, 1000, 3000, 5000\}$ in each row from left to right (ground truth shown in Figure~\ref{fig:illuminationGroundTruthT0}). The rows present the results of HillClimbing, CopyBest and XoverRandCP02 ($mr=0.05$) respectively.} \label{fig:illuminationSingleRepresentationVisualizationProcess}
\end{figure}

Figure~\ref{fig:illuminationSingleRepresentationVisualizationProcess} shows a visualization of the results obtained by HillClimbing, CopyBest and XoverRandCP02 ($mr=0.05$) during the evolutionary process. We observe that with no parameter sharing (HillClimbing), the results are very noisy, with several isolated agents that are non-optimal (Figures~\ref{fig:illuminationSingleHillClimbing101}-\ref{fig:illuminationSingleHillClimbing5000}). With crossover, results are clearly better, see Figures~\ref{fig:illuminationSingleXoverRand02101}-\ref{fig:illuminationSingleXoverRand025000}. However, we observe in the results of XoverRand that at $g=100$ there are many non-optimal agents at the right and left edges of the image, see Figure~\ref{fig:illuminationSingleXoverRand02101}. This is due to the problem formulation and the arithmetic crossover operator: in fact, there are two optimal values for the left and right edge of the image ($0$ and $50$), due to the periodicity of the sine function. Thus, agents are optimized towards one of these values, which may be different from that of their neighbors. Moreover, arithmetic crossover in this case does not help because the fitness of the average of these two values is worse. %We observe this also in Figure~\ref{fig:compareIlluminationSingle} where some versions of the algorithm with crossover operator do not appear to perform well.

%---------------------------------------------

\subsubsection{Vector representation scenario}
We tested the five Embodied Evolution algorithms with the same parameter settings tested for the imitation problem. Figure~\ref{fig:fitnessTrendsIlluminationVector} shows the average (across $10$ runs per algorithm) collective fitness ($F_g$) trends obtained with the tested algorithms, and the corresponding std. dev. Also in this case, for CopyBest, CopyRand, XoverBest and XoverRand we show the fitness trend obtained with the best parameter setting according to the Nemenyi test, see \ref{app:stats}.

Similarly to imitation problem, we observe that the versions that use the crossover operator tend to perform better. Moreover, XoverRand performs the best for different mutation rates, as it does in the imitation problem. Also, we observe that smaller mutation rates lead to slower convergence in all cases. 

\begin{figure}[!ht]
\begin{subfigures}
\subfloat[]{\includegraphics[width=0.25\columnwidth, height=0.15\columnwidth]{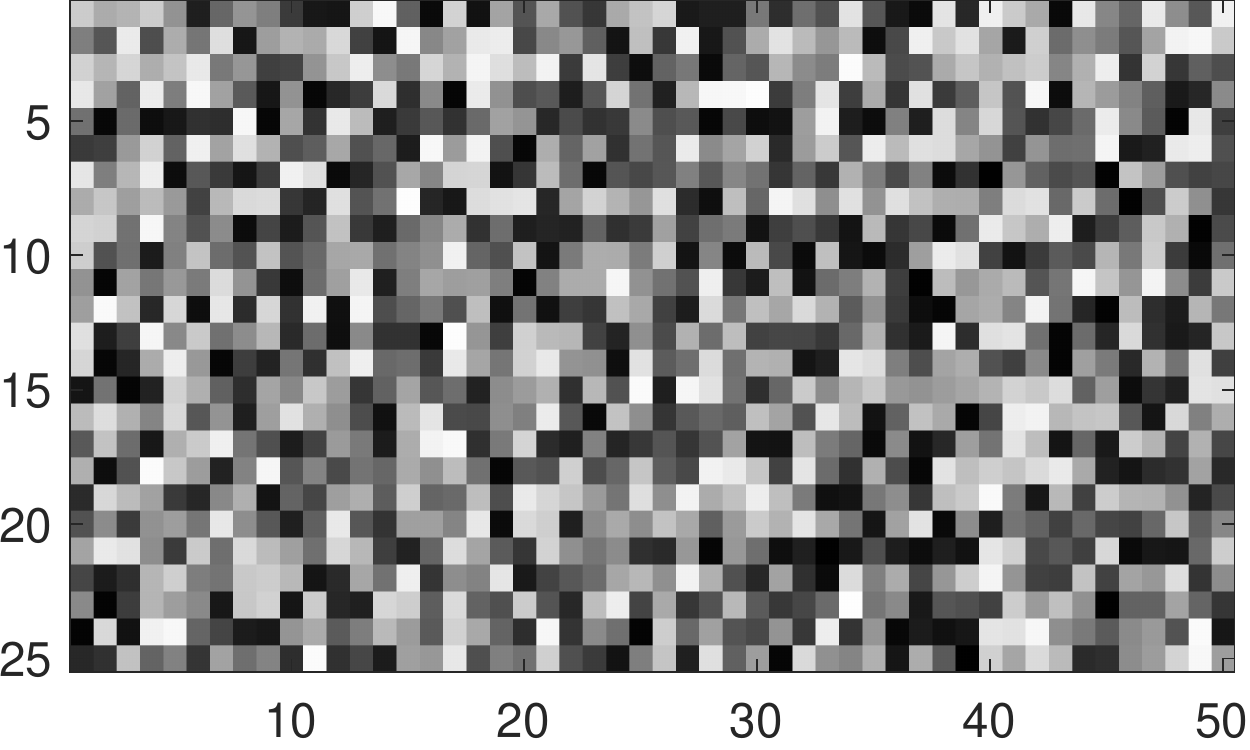}\label{fig:intervalsHillClimbingIllumination101}}
\subfloat[]{\includegraphics[width=0.25\columnwidth, height=0.15\columnwidth]{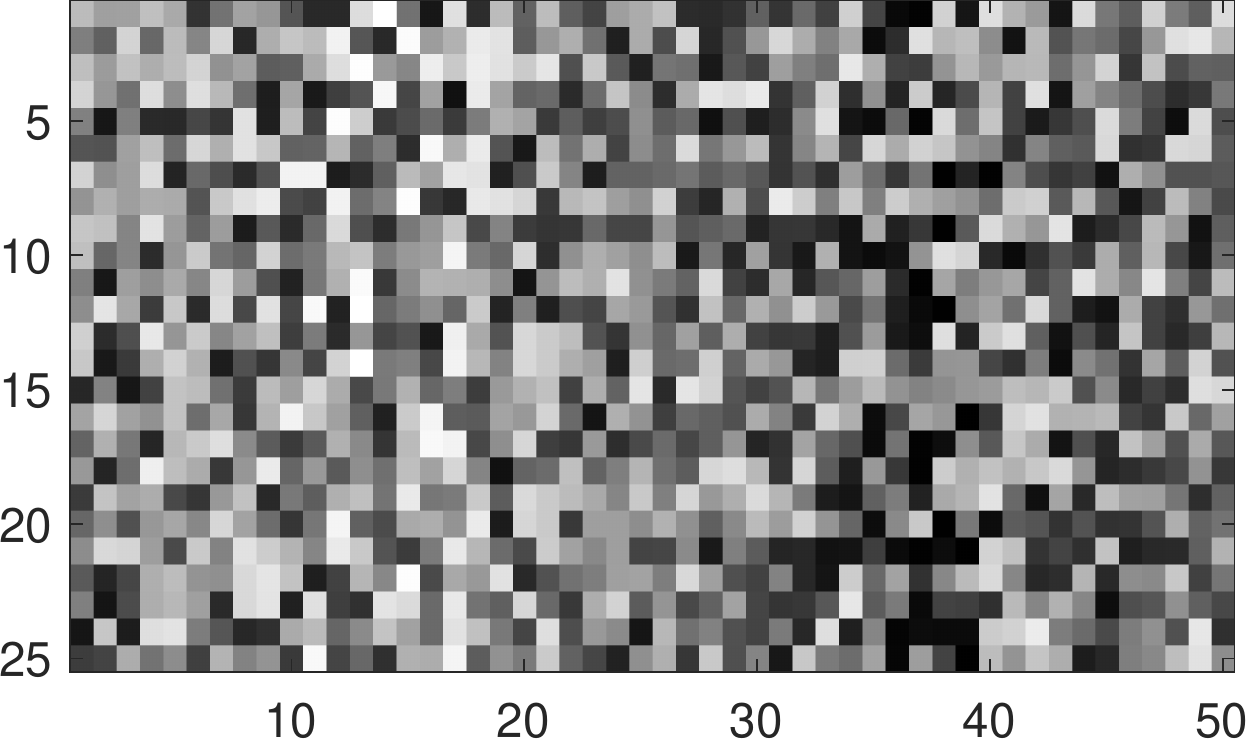}\label{fig:intervalsHillClimbingIllumination1000}}
\subfloat[]{\includegraphics[width=0.25\columnwidth, height=0.15\columnwidth]{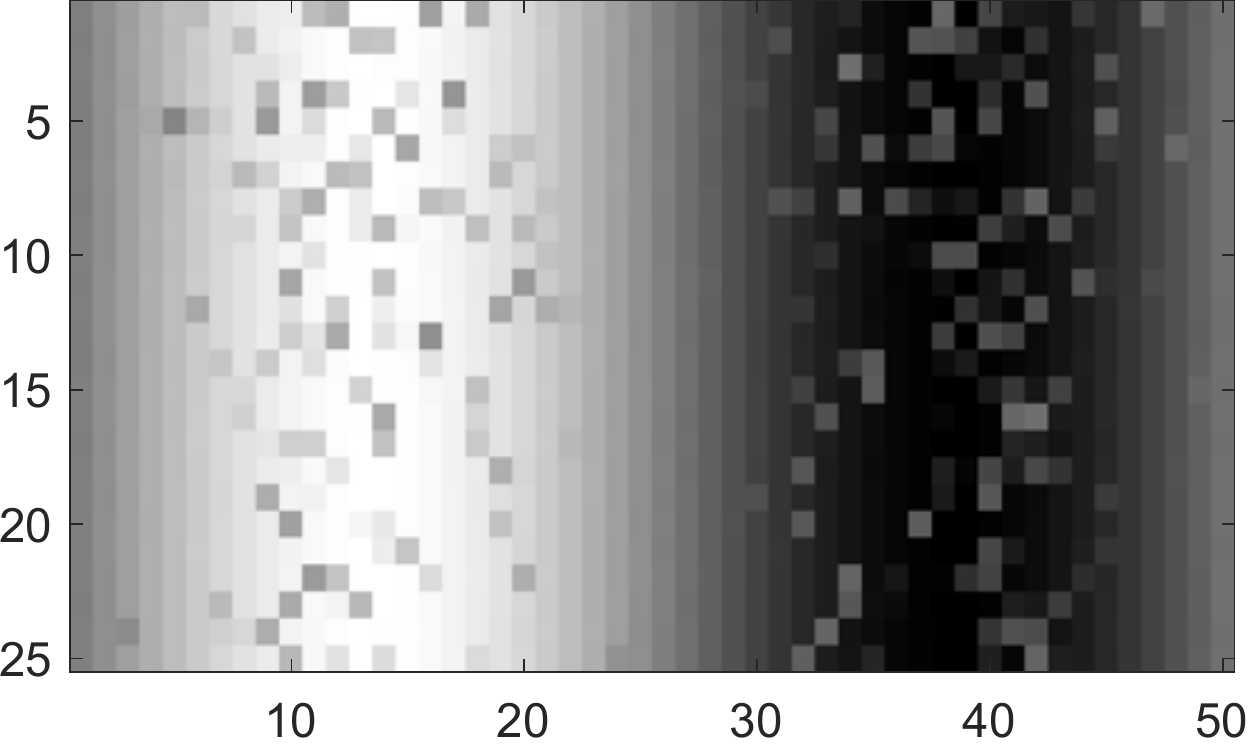}\label{fig:intervalsHillClimbingIllumination10000}}
\subfloat[]{\includegraphics[width=0.25\columnwidth, height=0.15\columnwidth]{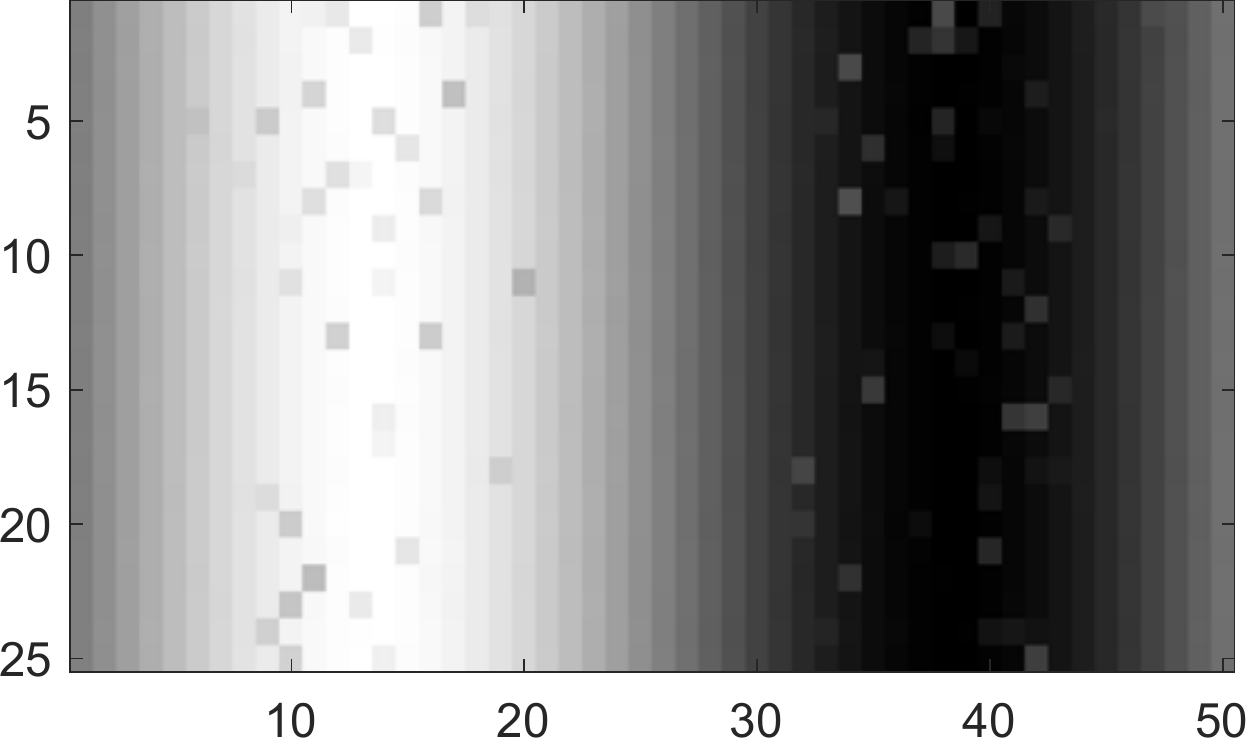}\label{fig:intervalsHillClimbingIllumination20000}}

\subfloat[]{\includegraphics[width=0.25\columnwidth, height=0.15\columnwidth]{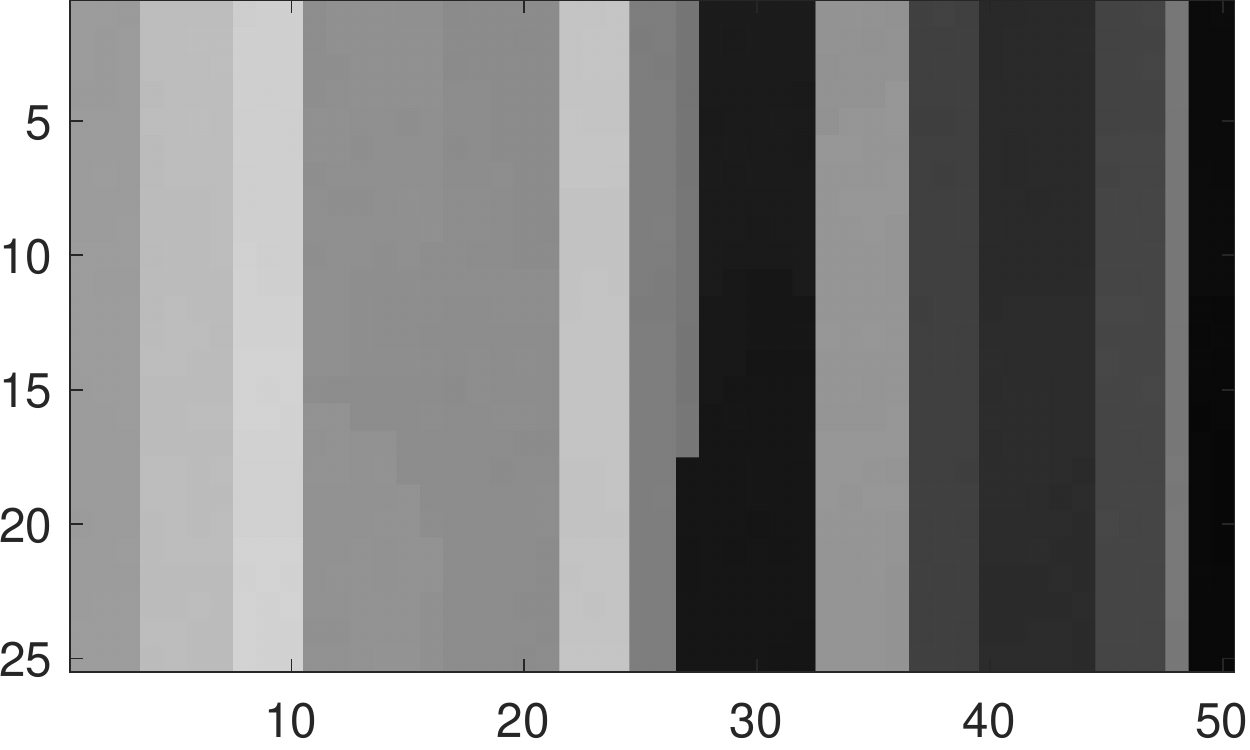}\label{fig:intervalsCopyBestIllumination101}}
\subfloat[]{\includegraphics[width=0.25\columnwidth, height=0.15\columnwidth]{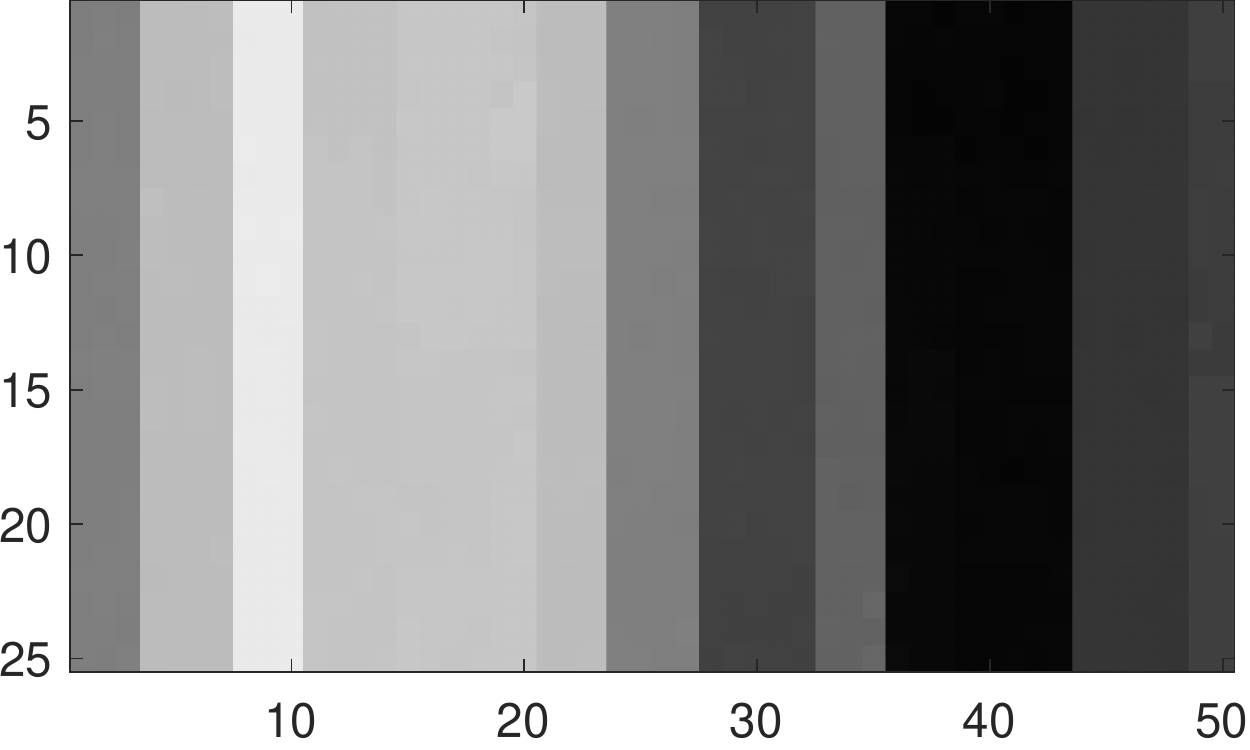}\label{fig:intervalsCopyBestIllumination1000}}
\subfloat[]{\includegraphics[width=0.25\columnwidth, height=0.15\columnwidth]{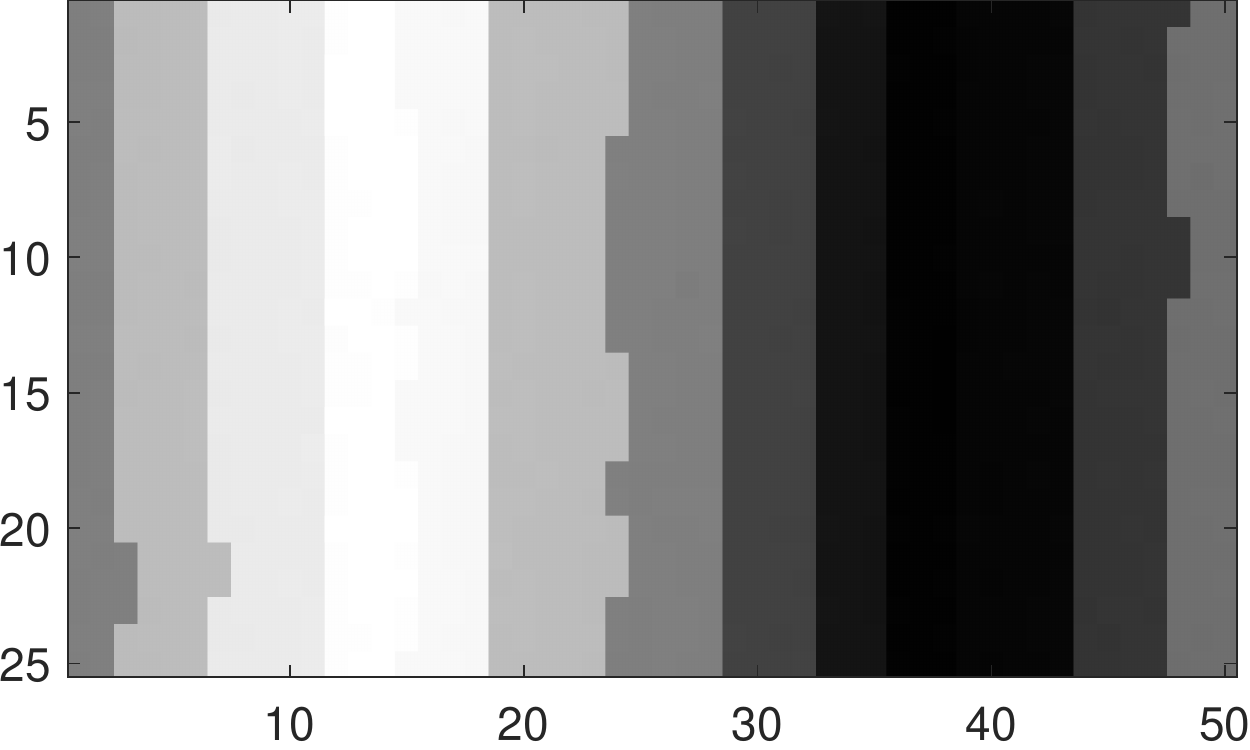}\label{fig:intervalsCopyBestIllumination10000}}
\subfloat[]{\includegraphics[width=0.25\columnwidth, height=0.15\columnwidth]{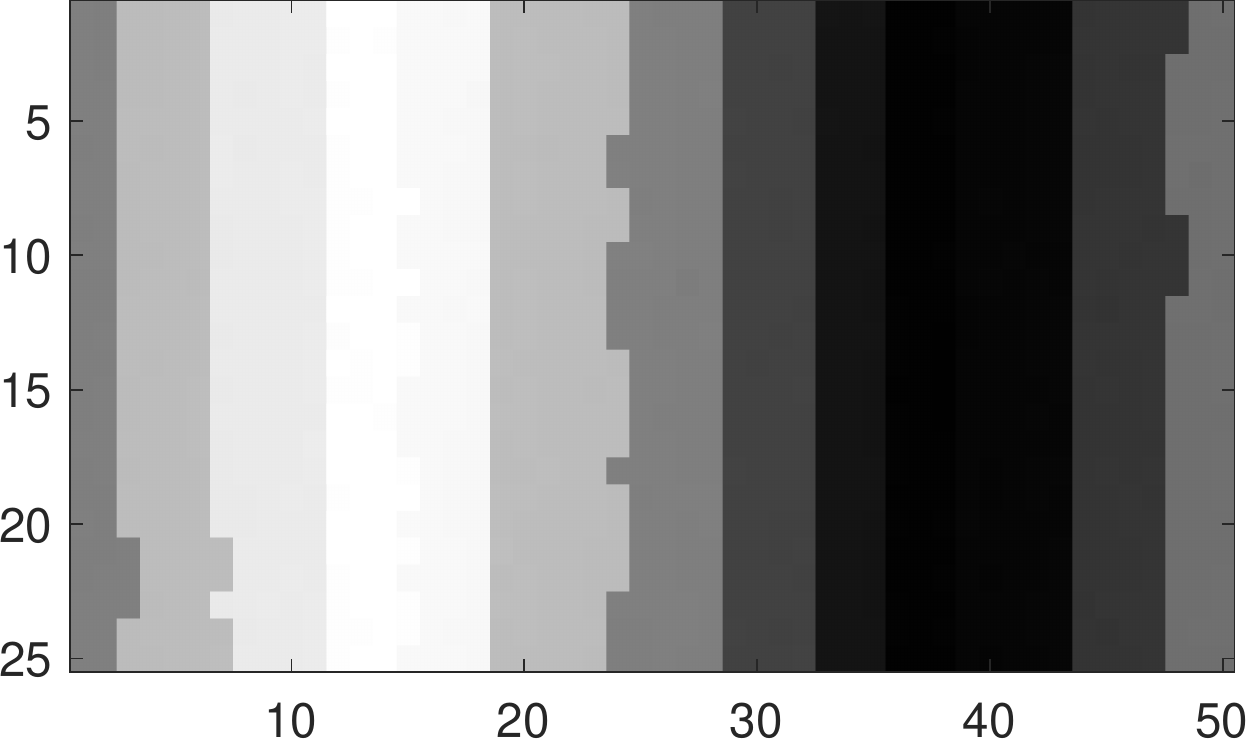}\label{fig:intervalsCopyBestIllumination20000}}

\subfloat[]{\includegraphics[width=0.25\columnwidth, height=0.15\columnwidth]{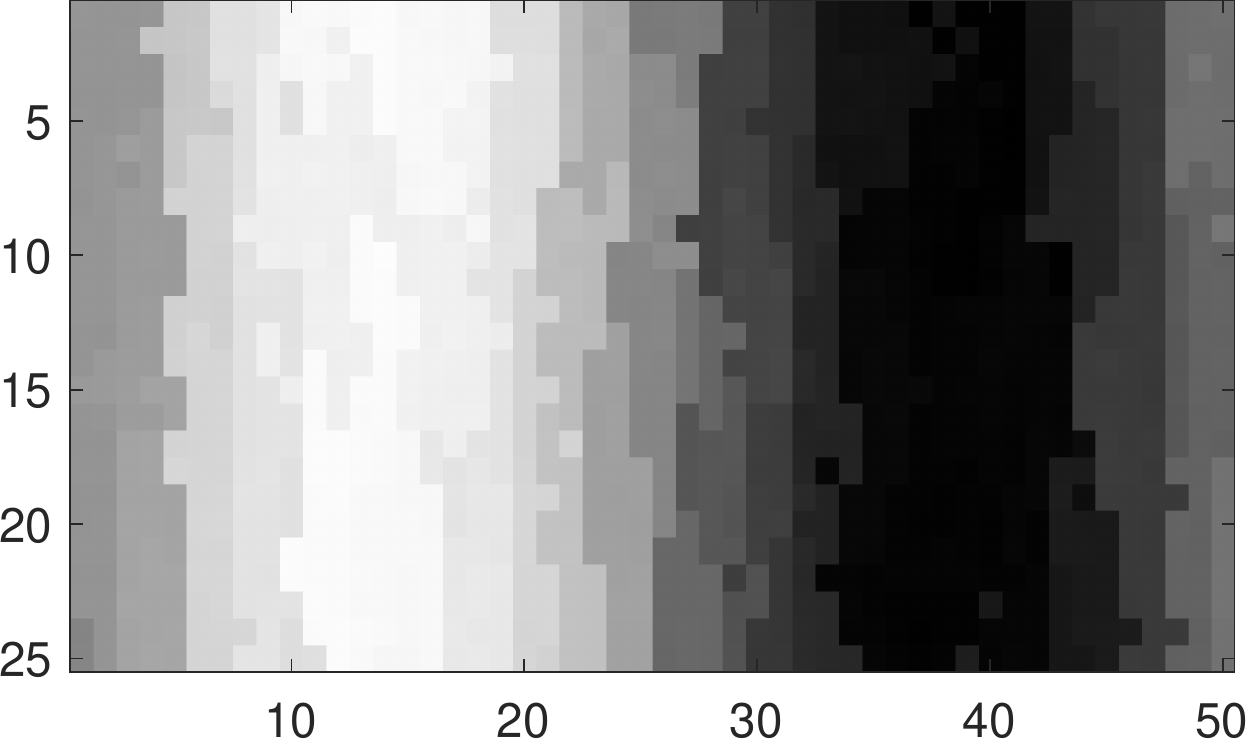}\label{fig:intervalsXoverRandIllumination101}}
\subfloat[]{\includegraphics[width=0.25\columnwidth, height=0.15\columnwidth]{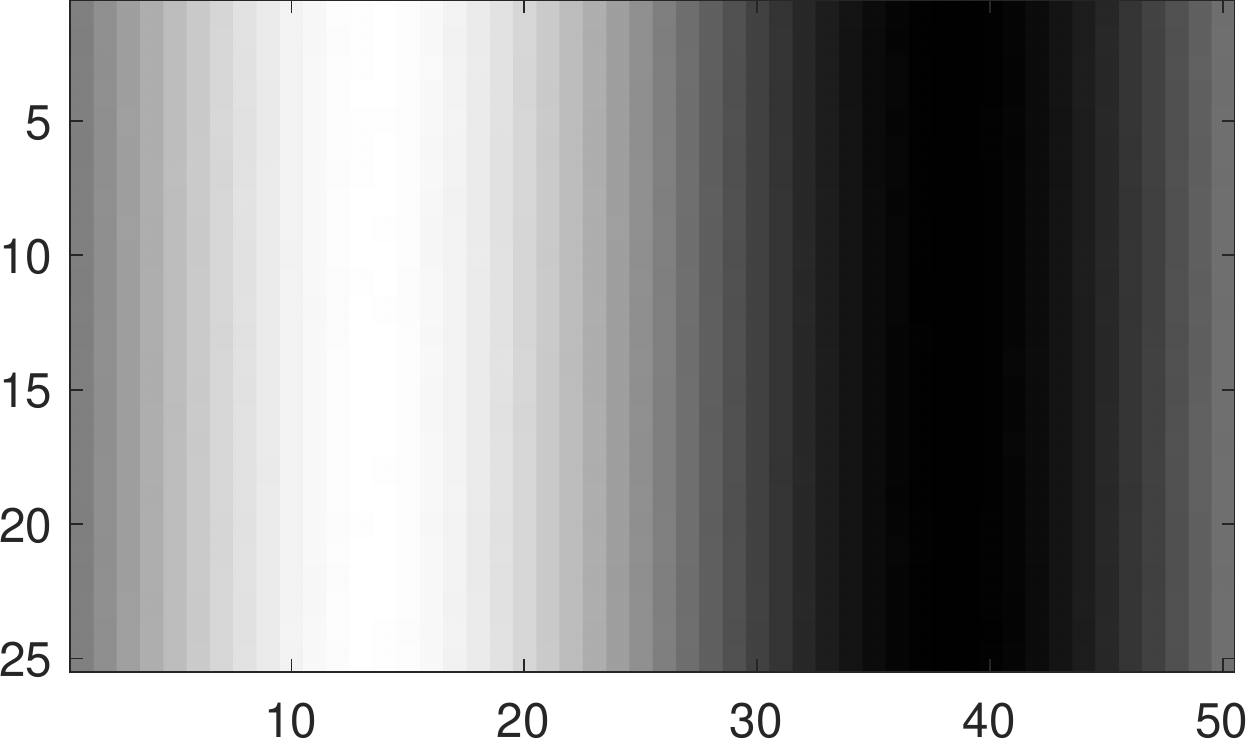}\label{fig:intervalsXoverRandIllumination1000}}
\subfloat[]{\includegraphics[width=0.25\columnwidth, height=0.15\columnwidth]{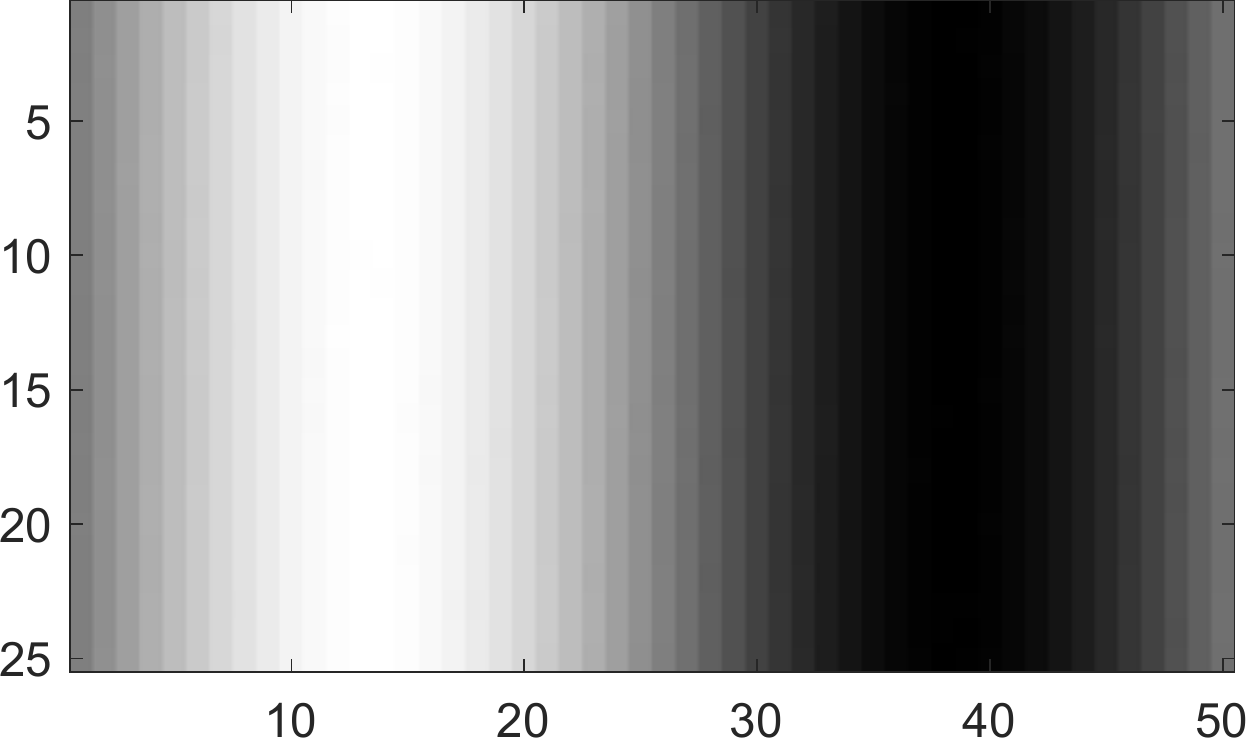}\label{fig:intervalsXoverRandIllumination10000}}
\subfloat[]{\includegraphics[width=0.25\columnwidth, height=0.15\columnwidth]{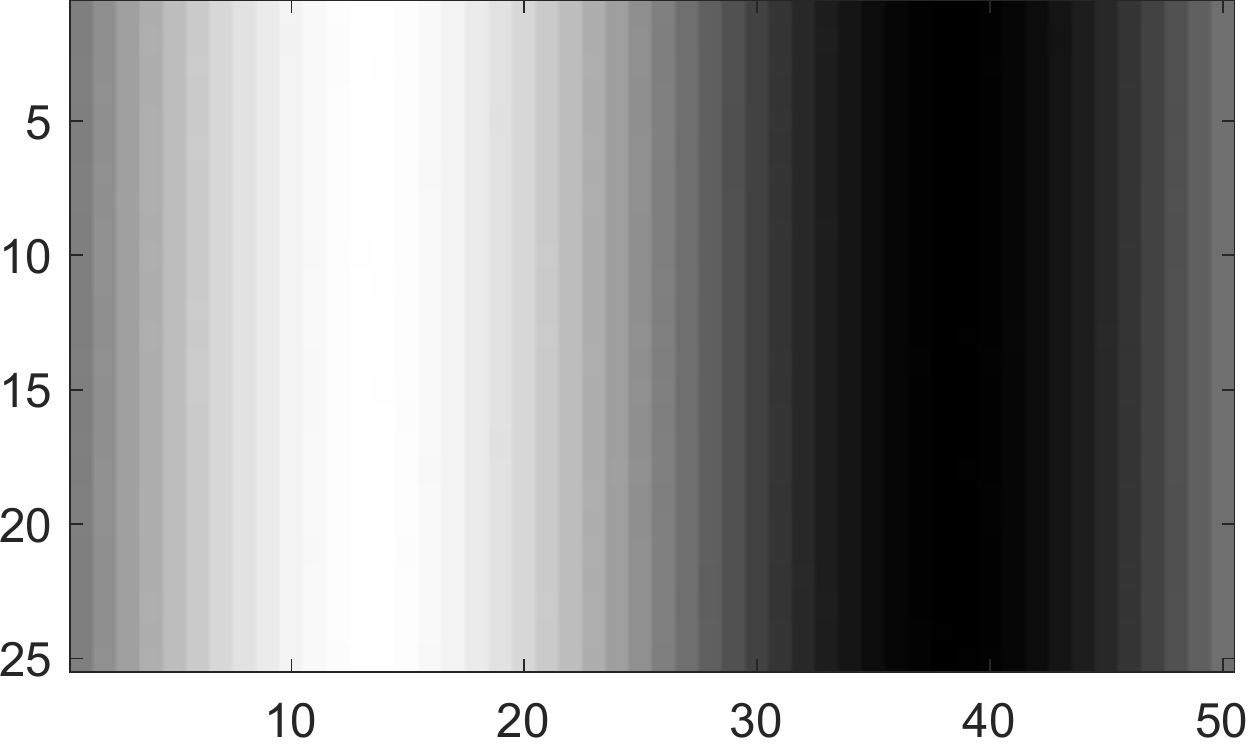}\label{fig:intervalsXoverRandIllumination20000}}
\end{subfigures}
\caption{Illumination problem (vector representation): Visualization of the evolutionary process at generations $g \in \{100, 1000, 10000, 20000\}$ for each row from left to right (ground truth shown in Figure~\ref{fig:illuminationGroundTruthT0}). The rows present the results of HillClimbing, CopyBest and XoverRand ($mr=0.001$) respectively.} \label{fig:illuminationVectorVisualizationProcess}
\end{figure}

Figure~\ref{fig:illuminationVectorVisualizationProcess} shows a visualization of the results obtained by HillClimbing, CopyBest and XoverRandCP02 ($mr=0.001$) during the evolutionary process\footnote[2]{Video of the evolutionary process available at: \url{https://youtu.be/5HZ-SMLyv_E}.}. The results of HillClimbing appear more noisy, with many isolated, non-optimal agents. Also, we observe vertical ``stripes'' in the results of CopyBest: this is due to the fact that the optimal parameters of the agents in each column are the same, therefore they tend to share the same parameters. Once again, we observe that crossover helps the optimization process even though the optimal parameters of neighboring agents are different.

%---------------------------------------------

\subsection{Distributed model of indoor human presence and activity}

\begin{figure}[ht!]
\begin{subfigures}
\subfloat[Human presence (room A)]{\includegraphics[width=0.45\columnwidth]{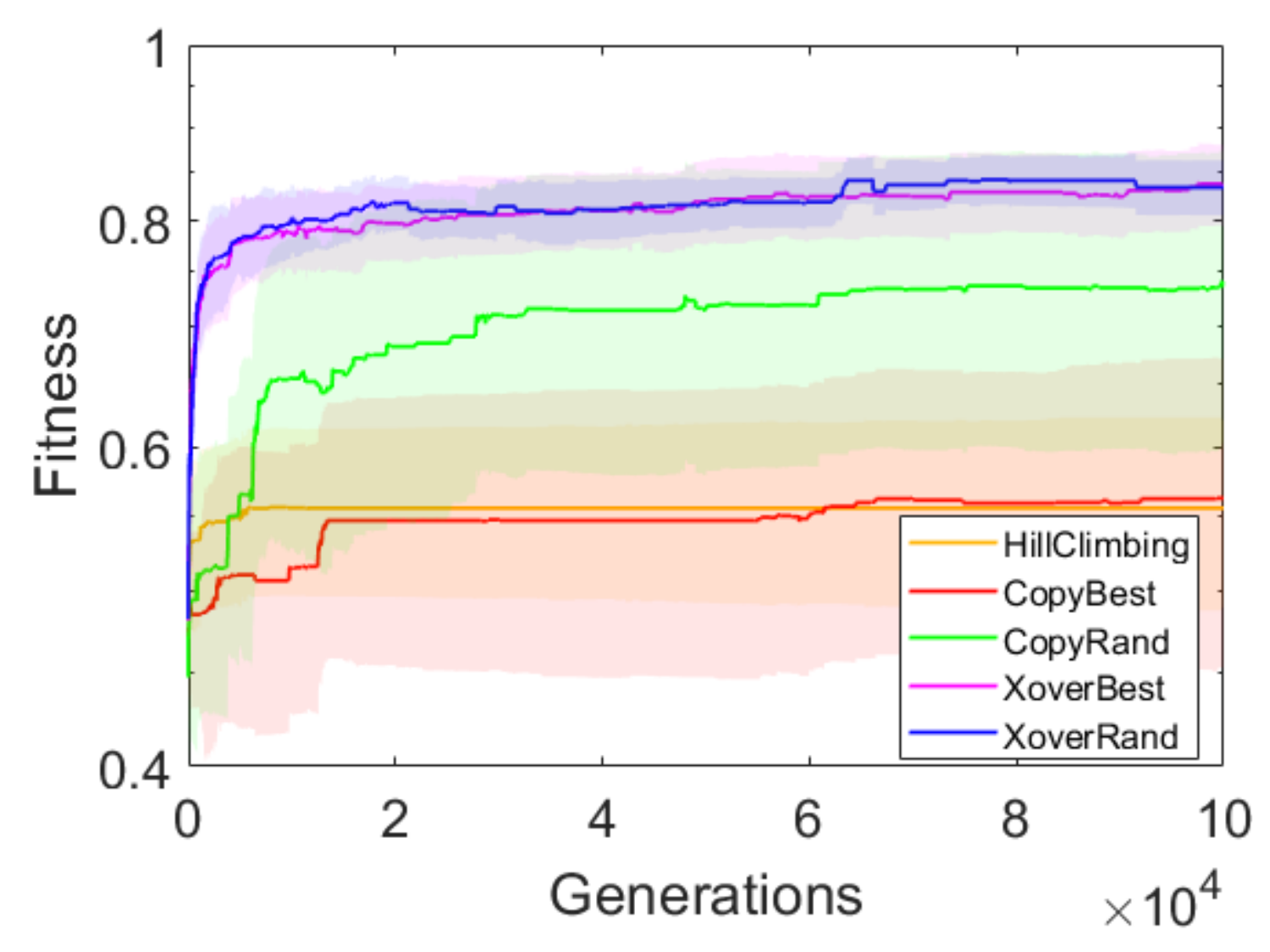}\label{fig:testLocAHOAll}}
\hfill
\subfloat[Human activity (room A)]{\includegraphics[width=0.45\columnwidth]{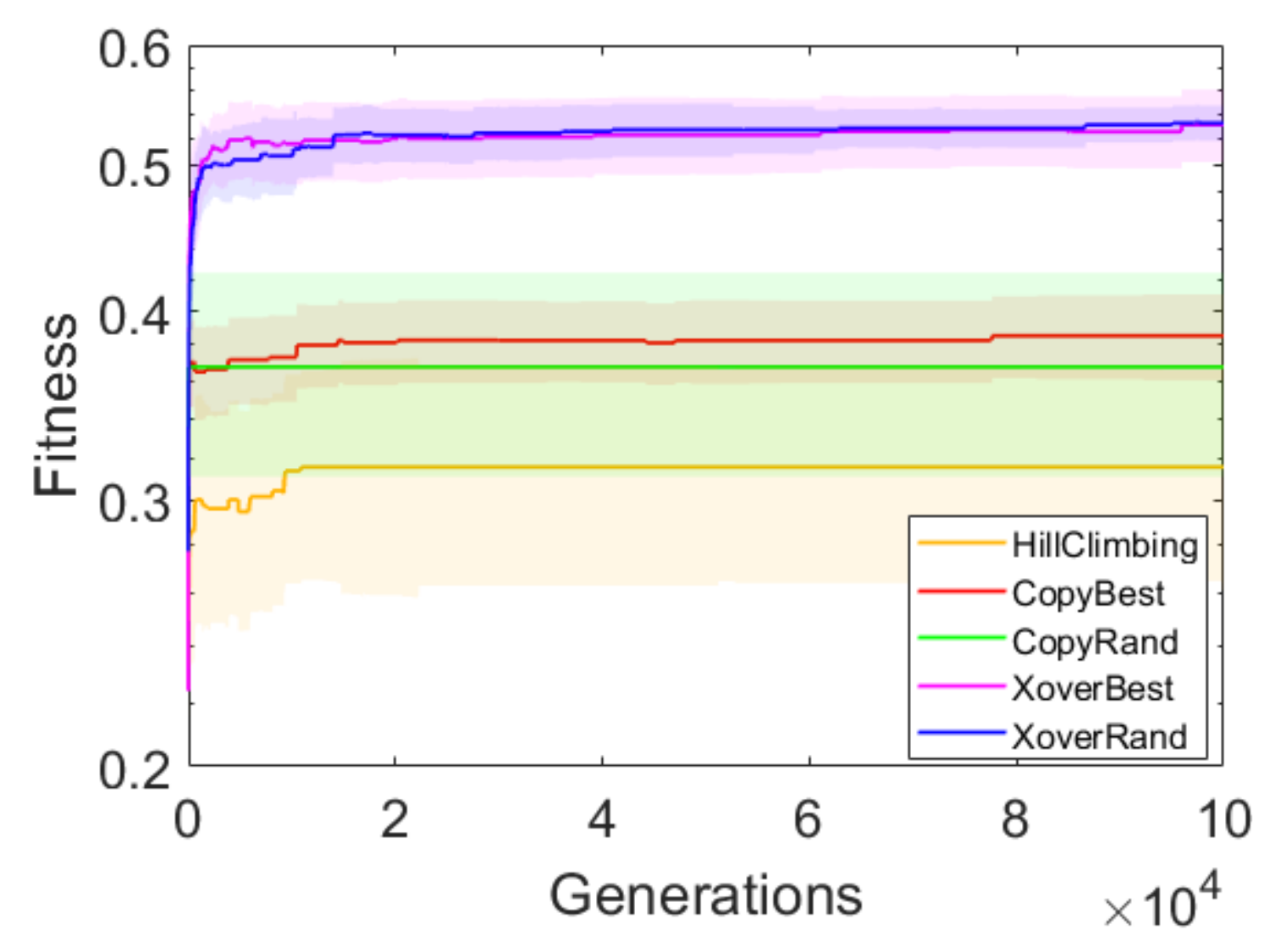}\label{fig:testLocAHAAll}}

\subfloat[Human presence (room B)]{\includegraphics[width=0.45\columnwidth]{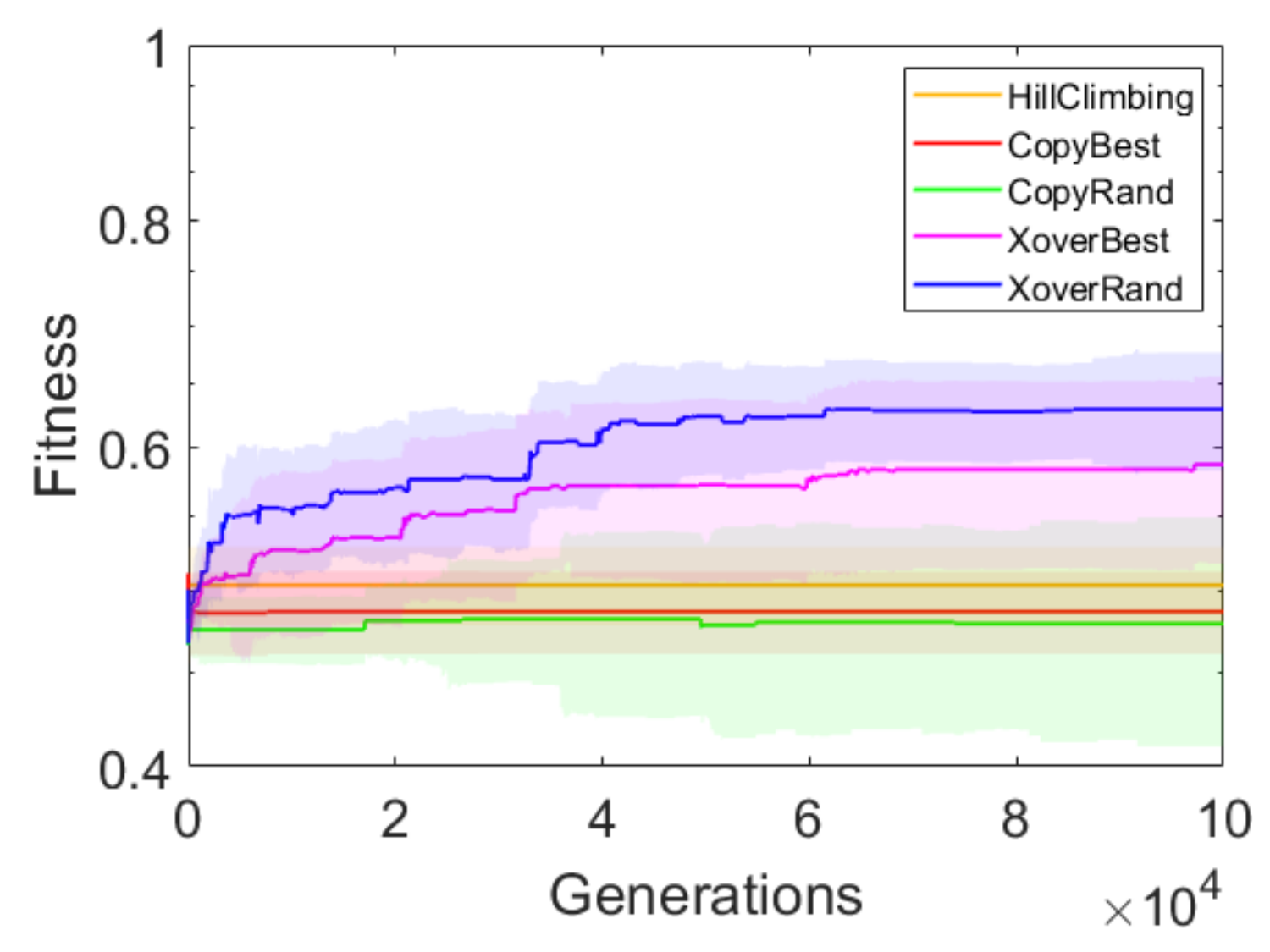}\label{fig:testLocBHOAll}}
\hfill
\subfloat[Human activity (room B)]{\includegraphics[width=0.45\columnwidth]{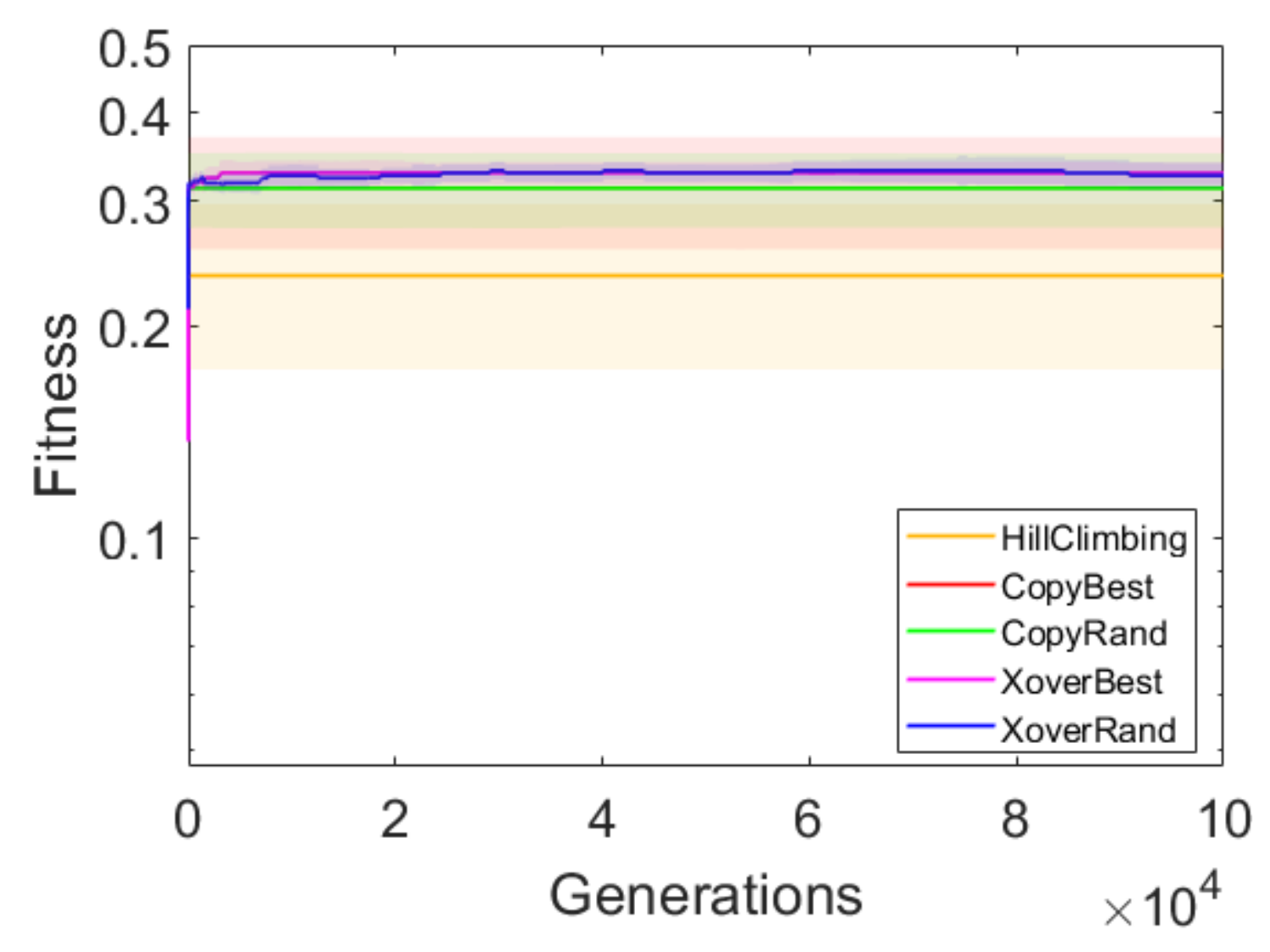}\label{fig:testLocBHAAll}}

\subfloat[Human presence (room C)]{\includegraphics[width=0.45\columnwidth]{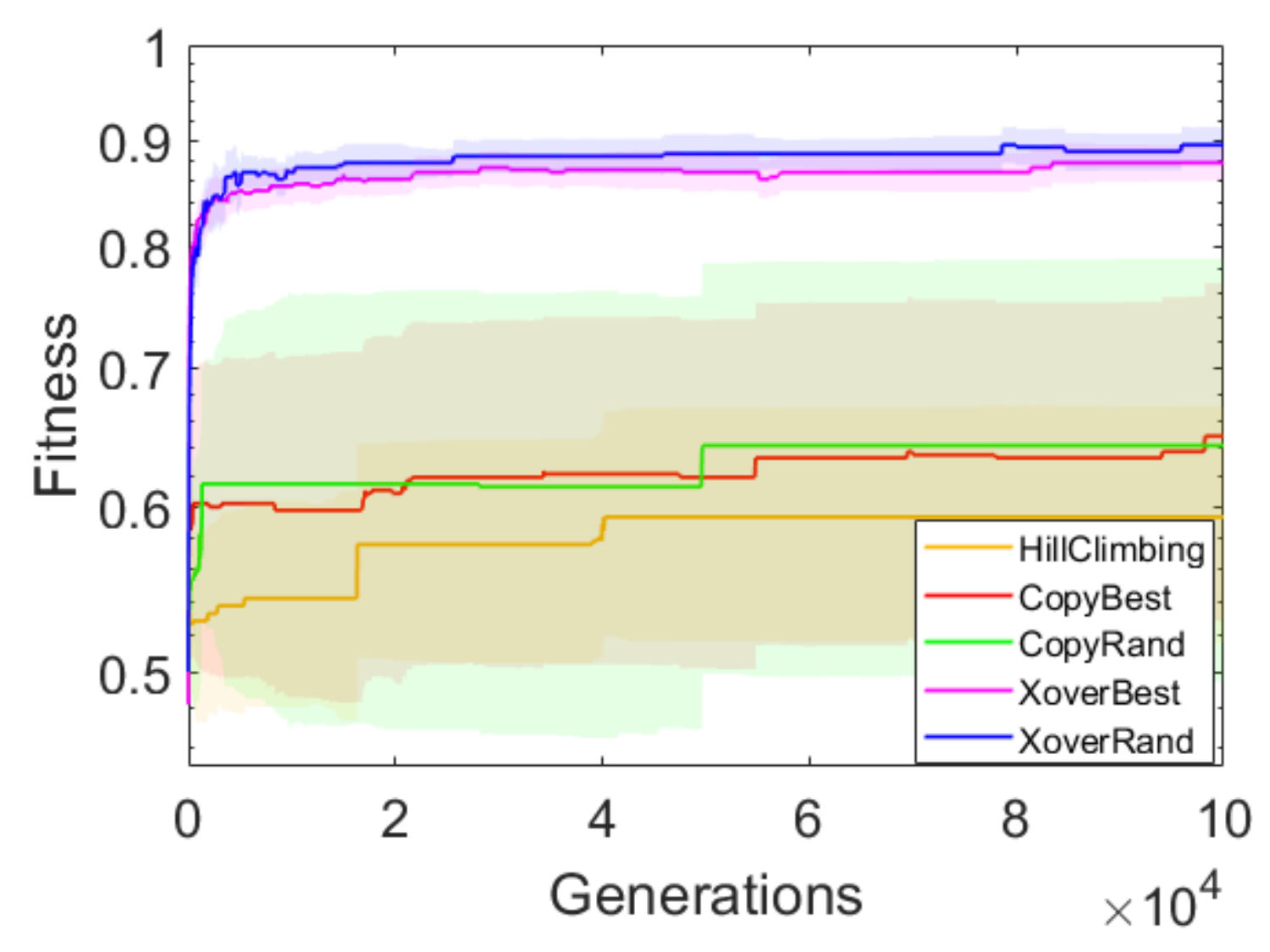}\label{fig:testLocCHOAll}}
\hfill
\subfloat[Human activity (room C)]{\includegraphics[width=0.45\columnwidth]{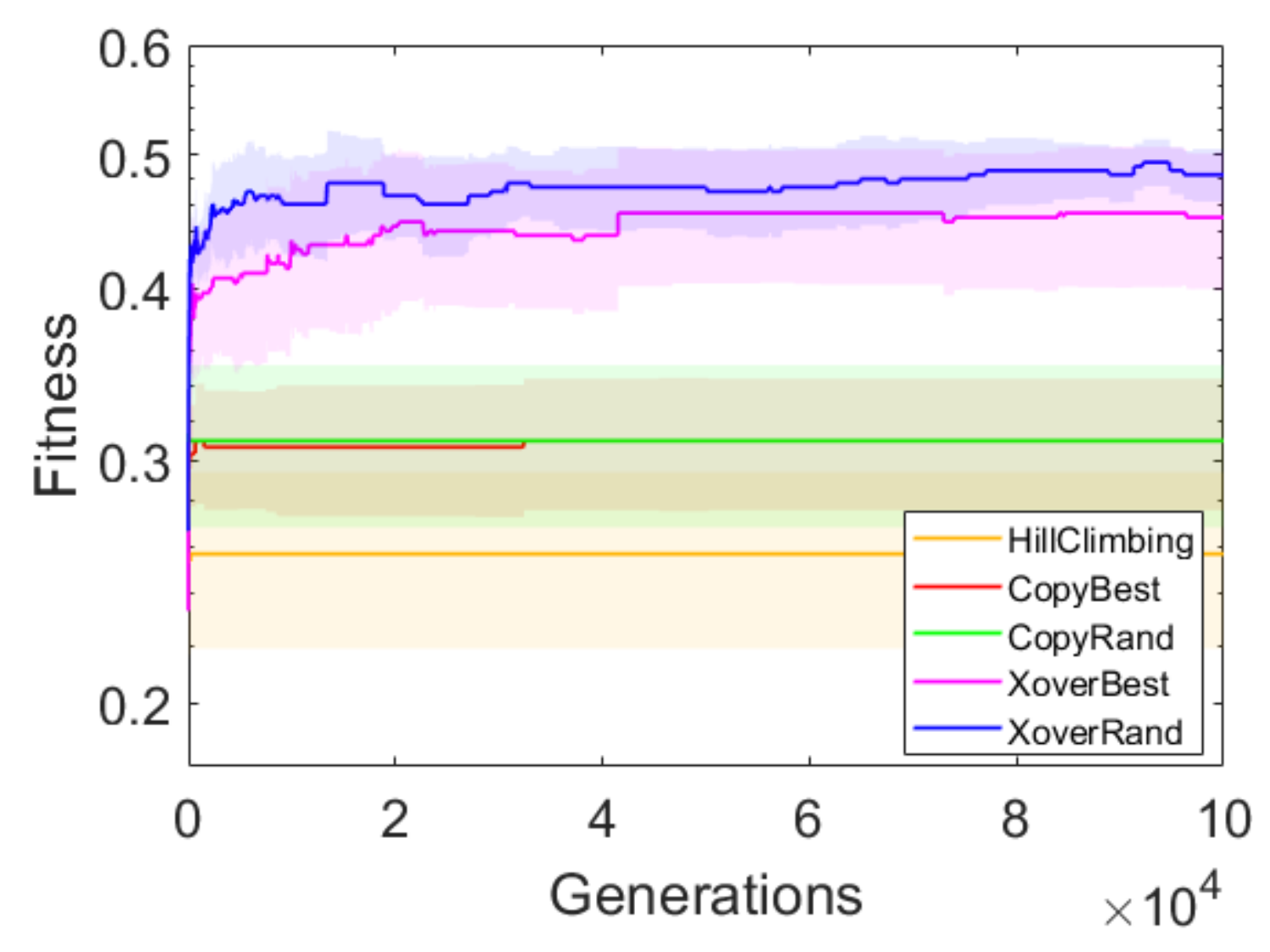}\label{fig:testLocCHAAll}}
\end{subfigures}
\caption{Distributed model: Test accuracy (mean $\pm$ std. dev. across $10$ runs) obtained with different EE algorithms during the evolutionary process. For CopyBest, CopyRand, XoverBest and XoverRand we show the fitness trend obtained with the best parameter setting.}\label{fig:roomClimateresults}
\end{figure}

Differently from the two previous problems, in this case many versions of the algorithm produced statistically equivalent results\footnote{See Tables~\ref{tab:wilcoxonLocAHO}-\ref{tab:wilcoxonLocCHA} in the Appendix for pairwise comparisons of $p$-values.}. We show in Figure~\ref{fig:roomClimateresults} the fitness trends (test accuracy) obtained with the Embodied Evolution algorithms for each room and task, averaged across $10$ runs per algorithm, with the corresponding std. dev. As we did for the previous two problems, in the case of CopyBest, CopyRand, XoverBest and XoverRand we show the fitness trend obtained with the best parameter setting according to the Nemenyi test, see \ref{app:stats} for details. Overall, also on this problem the algorithms that use crossover perform better. In particular, XoverRand performs best in all rooms on the human presence task, and in room A and C on the human activity task. In the remaining case (human activity task in room B), XoverBest performs slightly better than XoverRand. However, as we discuss below this task appears intrinsically harder than the others, and the difference in performance across different algorithms is less clear. In all cases, HillClimbing shows the worst results, followed by CopyBest and CopyRand which are statistically equivalent in most cases except for the human presence task in room A, where CopyRand performs much better than CopyBest. These results confirm then the general trends that we observed on the imitation and illumination problems.

%TODO or median?

For completeness, in Tables~\ref{tab:results-LocA}-\ref{tab:results-LocC}, we report the model results (in terms of training and test accuracy) obtained with the best Embodied Evolution algorithm (according to the Nemenyi test) on each node and each room for both tasks (human presence and human activity), in comparison with those reported in \cite{morgner2017privacy}. Please note that concerning the original results on the human activity task, except for the nodes A1 and C1 (i.e., the nodes showing the best and worst accuracy value, respectively) all the other results are reported only graphically, therefore the exact accuracy values are not available. Also, in \cite{morgner2017privacy} only the test accuracy results are reported.

From the results, it can be seen that the Embodied Evolution algorithm (in particular the XoverRand/XoverBest versions, depending on the room and task) is quite effective especially on the human activity task, although with different performances depending on the room and node. Limiting our analysis to the test accuracy, it can be noted that:
\begin{itemize}[leftmargin=*]
    \item On the human presence task, the results are --in general-- slightly worse than those reported in \cite{morgner2017privacy}. This is likely due to the different Machine Learning model used (in \cite{morgner2017privacy} the model was a Na\"{i}ve Bayes classifier, while in our case it is a FFNN), and to the fact that in \cite{morgner2017privacy} the sensor data are pre-processed (with statistical analysis and feature selection), while in our case we simply use the raw data, properly scaled to feed the Neural Network. Nevertheless, the accuracy values are high in all cases (better than the probability of random guessing, $50\%$). Furthermore, in the case of room C our method is able to find an accuracy value on nodes C3 and C4 that is even higher than that reported in \cite{morgner2017privacy} on node C1. Finally, as in \cite{morgner2017privacy} we observe that the task on room B seems to be more difficult (showing lower accuracy values) w.r.t. the other two rooms. This is probably due to the comparatively larger size of this room and the layout of the sensor network.
    \item On the human activity task, the results are much better than those reported in \cite{morgner2017privacy} in the case of room A and C, while in the case of room B they are quite worse. This finding is especially interesting because this multi-class classification problem is supposed to be harder than the human presence problem, that is instead binary (either a human is present in the room or not). In particular, we can observe that our method produces consistently higher accuracy values on all nodes in room A and C. In the case of room B, it produces better accuracy only on node B3, while in the other two nodes the accuracy is much lower. Again, this is likely due to the layout of room B and its corresponding sensor network, and to the fact that this network is composed of only three nodes, such that the parameter exchange is probably too limited to be effective.
\end{itemize}

To conclude, this IoT problem shows how Embodied Evolution can be easily applied to real-world network cases, where leveraging the parameter exchange can lead to improved performances. On the other hand, the case of room B shows one possible limitation of our scheme, whose performance might be affected by the number as well as the position of the nodes, although these aspects might also depend on cost and logistic constraints of the specific application. As a side note, this observation is consistent with the literature on evolutionary algorithms, where various works, such as \cite{yaman2019comparison}, have shown that ``micro-population'' algorithms (i.e., algorithms with a very small population size) tend to get stuck at local optima more easily.

\begin{table}[ht!]
\centering
\caption{Distributed model (room A): Training/test accuracy at each node found by the best performing Embodied Evolution algorithm. The boldface indicates the best accuracy value (across the nodes) obtained on each task.}\label{tab:results-LocA}
% \begin{tabular}{|c|P{1.5cm}|P{1.5cm}|P{1.5cm}|P{1.5cm}|}
% \hline
% \multirow{2}{*}{\textbf{Node}} & \multicolumn{2}{c|}{\textbf{Human presence}} & \multicolumn{2}{c|}{\textbf{Human activity}} \\ \cline{2-5}
% & \textbf{Training} & \textbf{Test} & \textbf{Training} & \textbf{Test} \\ \hline
% \textbf{A1} & 85.46 & \textbf{85.56} & 55.93 & \textbf{57.47} \\ \hline
% \textbf{A2} & \textbf{87.73} & 85.51 & \textbf{59.09} & 51.32 \\ \hline
% \textbf{A3} & 83.87 & 80.47 & 57.94 & 52.04 \\ \hline
% \textbf{A4} & 83.44 & 81.45 & 54.76 & 51.32 \\ \hline
% \end{tabular}
\begin{tabular}{|c|P{1.5cm}|P{1.5cm}|P{1.5cm}|}
\hline
 \multirow{3}{*}{\textbf{Node}} & \multicolumn{3}{c|}{\textbf{Human presence}} \\ \cline{2-4}
 & \textbf{Training} [ours] & \textbf{Test} [ours] & \textbf{~Test}
 \newline
 \cite{morgner2017privacy} \\ \hline
\textbf{A1} & 85.46 & \textbf{85.56} & \textbf{93.5}\\ \hline
\textbf{A2} & \textbf{87.73} & 85.51 & 89.7\\ \hline
\textbf{A3} & 83.87 & 80.47 & 84.2\\ \hline
\textbf{A4} & 83.44 & 81.45 & 78.4\\ \hline
\end{tabular}
\begin{tabular}{|c|P{1.5cm}|P{1.5cm}|P{1.5cm}|}
\hline
 \multirow{3}{*}{\textbf{Node}} & \multicolumn{3}{c|}{\textbf{Human activity}} \\ \cline{2-4}
 & \textbf{Training} [ours] & \textbf{Test} [ours] & \textbf{~Test}
 \newline
 \cite{morgner2017privacy} \\ \hline
\textbf{A1} & 55.93 & \textbf{57.47} & \textbf{56.8}\\ \hline
\textbf{A2} & \textbf{59.09} & 51.32 & $\sim$ 35\\ \hline
\textbf{A3} & 57.94 & 52.04 & $\sim$ 30\\ \hline
\textbf{A4} & 54.76 & 51.32 & $\sim$ 24\\ \hline
\end{tabular}
\end{table}

\begin{table}[ht!]
\centering
\caption{Distributed model (room B): Training/test accuracy at each node found by the best performing Embodied Evolution algorithm. The boldface indicates the best accuracy value (across the nodes) obtained on each task.}\label{tab:results-LocB}
% \begin{tabular}{|c|P{1.5cm}|P{1.5cm}|P{1.5cm}|P{1.5cm}|}
% \hline
% \multirow{2}{*}{\textbf{Node}} & \multicolumn{2}{c|}{\textbf{Human presence}} & \multicolumn{2}{c|}{\textbf{Human activity}} \\ \cline{2-5}
% & \textbf{Training} & \textbf{Test} & \textbf{Training} & \textbf{Test} \\ \hline
% \textbf{B1} & \textbf{74.55} & \textbf{62.78} & \textbf{35.59} & 30.65\\ \hline
% \textbf{B2} & 72.00 & 60.61 & 33.48 & 28.85 \\ \hline
% \textbf{B3} & 73.11 & 62.56 & 33.43 & \textbf{39.35}\\ \hline
% \end{tabular}
\begin{tabular}{|c|P{1.5cm}|P{1.5cm}|P{1.5cm}|}
\hline
 \multirow{3}{*}{\textbf{Node}} & \multicolumn{3}{c|}{\textbf{Human presence}} \\ \cline{2-4}
 & \textbf{Training} [ours] & \textbf{Test} [ours] & \textbf{~Test}
 \newline
 \cite{morgner2017privacy} \\ \hline
\textbf{B1} & \textbf{74.55} & \textbf{62.78} & \textbf{88.5}\\ \hline
\textbf{B2} & 72.00 & 60.61 & 81.3\\ \hline
\textbf{B3} & 73.11 & 62.56 & 66.8\\ \hline
\end{tabular}
\begin{tabular}{|c|P{1.5cm}|P{1.5cm}|P{1.5cm}|}
\hline
 \multirow{3}{*}{\textbf{Node}} & \multicolumn{3}{c|}{\textbf{Human activity}} \\ \cline{2-4}
 & \textbf{Training} [ours] & \textbf{Test} [ours] & \textbf{~Test}
 \newline
 \cite{morgner2017privacy} \\ \hline
\textbf{B1} & \textbf{35.59} & 30.65 & \textbf{$\sim$ 54}\\ \hline
\textbf{B2} & 33.48 & 28.85 & $\sim$ 35 \\ \hline
\textbf{B3} & 33.43 & \textbf{39.35} & $\sim$ 30\\ \hline
\end{tabular}
\end{table}

\begin{table}[ht!]
\centering
\caption{Distributed model (room C): Training/test accuracy at each node found by the best performing Embodied Evolution algorithm. The boldface indicates the best accuracy value (across the nodes) obtained on each task.}\label{tab:results-LocC}
% \begin{tabular}{|c|P{1.5cm}|P{1.5cm}|P{1.5cm}|P{1.5cm}|}
% \hline
%  \multirow{2}{*}{\textbf{Node}} & \multicolumn{2}{c|}{\textbf{Human presence}} & \multicolumn{2}{c|}{\textbf{Human activity}} \\ \cline{2-5}
%  & \textbf{Training} & \textbf{Test} & \textbf{Training} & \textbf{Test} \\ \hline
% \textbf{C1} & 92.13 & 86.67 & 52.97 & 45.00\\ \hline
% \textbf{C2} & 92.73 & 91.11 & 48.31 & \textbf{56.67}\\ \hline
% \textbf{C3} & \textbf{93.59} & \textbf{92.22} & 55.60 & 48.33\\ \hline
% \textbf{C4} & 90.76 & \textbf{92.22} & 50.43 & 45.00\\ \hline
% \textbf{C5} & 86.59 & 86.67 & \textbf{58.97} & 48.33\\ \hline
% \end{tabular}
\begin{tabular}{|c|P{1.5cm}|P{1.5cm}|P{1.5cm}|}
\hline
 \multirow{3}{*}{\textbf{Node}} & \multicolumn{3}{c|}{\textbf{Human presence}} \\ \cline{2-4}
 & \textbf{Training} [ours] & \textbf{Test} [ours] & \textbf{~Test}
 \newline
 \cite{morgner2017privacy} \\ \hline
\textbf{C1} & 92.13 & 86.67 & \textbf{91.0} \\ \hline
\textbf{C2} & 92.73 & 91.11 & 88.6 \\ \hline
\textbf{C3} & \textbf{93.59} & \textbf{92.22} & 86.0 \\ \hline
\textbf{C4} & 90.76 & \textbf{92.22} & 88.7 \\ \hline
\textbf{C5} & 86.59 & 86.67 & 90.7 \\ \hline
\end{tabular}
\begin{tabular}{|c|P{1.5cm}|P{1.5cm}|P{1.5cm}|}
\hline
 \multirow{3}{*}{\textbf{Node}} & \multicolumn{3}{c|}{\textbf{Human activity}} \\ \cline{2-4}
 & \textbf{Training} [ours] & \textbf{Test} [ours] & \textbf{~Test}
 \newline
 \cite{morgner2017privacy} \\ \hline
\textbf{C1} & 52.97 & 45.00 & 23.9\\ \hline
\textbf{C2} & 48.31 & \textbf{56.67} & $\sim$ 24\\ \hline
\textbf{C3} & 55.60 & 48.33 & \textbf{$\sim$ 34}\\ \hline
\textbf{C4} & 50.43 & 45.00 & $\sim$ 24\\ \hline
\textbf{C5} & \textbf{58.97} & 48.33 & $\sim$ 31\\ \hline
\end{tabular}
\end{table}

\clearpage

%% file: 5_conclusions.tex
\section{Conclusions}
\label{sec:conclusions}

In many network applications, the optimal behavior of the agents is not known before deployment, and the agents often need to adapt to environment changes. Performing the optimization process offline, by collecting each agent to change its behavior parameters whenever it is needed, would be costly and ineffective. In this work, we used a distributed Embodied Evolution approach to optimize the behavior of networks of agents at runtime.

We assumed a collection of spatially distributed agents that can communicate locally. We used the local communication capability to exchange the agents' behavior parameters, so to copy the neighbors' parameters and/or use them for local perturbations and crossover. Intuitively, the agents with close proximity may be required to perform similarly (thus may have similar optimal parameters) since they are likely to share similar environmental conditions. In this case, it would make sense to share their parameters to learn from one another. Nevertheless, even when the optimal parameters are drastically different within a neighborhood, parameter exchange may still be helpful in the optimization process. To test these two cases, we devised a number of experimental scenarios with various levels of differences between the optimal parameters of any given agent and those of its neighbors. We further compared these results with the case where each agent optimizes its behavior without any parameter exchange.

We found that parameter exchange (through crossover) works the best in all cases -except with arithmetic crossover and small mutation rates- including when the differences of the optimal parameters of the agents with their neighbors are large. We also observed a certain variation of the performance depending on the parameter settings of the algorithm (i.e., $mr$ and $cp$), therefore in the future it might be useful to evaluate (self-)adaptive evolutionary approaches. Among the various versions of the algorithm we tested, XoverRand demonstrated the best performance in general. Since this version does not require the fitness of the neighbors (due to random selection), in practical scenarios it might possible to broadcast only some randomly selected components of the agents' genotypes. In terms of limitations, the performance of the proposed EE scheme seems to be affected by the number as well as the position of the nodes, as we have observed in the IoT case. %Therefore, applying this approach in contexts with very few nodes (i.e. 3-4) might not provide meaningful advantages.

This work was mainly aimed at demonstrating the effectiveness of the Embodied Evolution approach over networks. Therefore, we made some simplifying assumptions, in particular we considered fixed topologies with perfect communication. In future works, we will extend this approach to more realistic cases, i.e. with dynamic connections and possibly noisy communication. Another assumption of this work was that the local fitness functions are in general different (in terms of optimum) across agents, but still comparable. However, this might not be the case of all applications: therefore, it may be interesting to see if parameter exchange is beneficial also in those cases where the local fitness functions are not directly comparable, and how embodied evolution could be applied (for instance, function similarity could be used to define the neighborhood function, to create niches/communities, or techniques from multi-factorial optimization \cite{7464295,7161358} could be applied). Another possibility would be to hybridize Embodied Evolution schemes with distributed gradient-based optimization \cite{rabbat2004distributed} and other forms of diffusion strategies \cite{chen2012diffusion}. Finally, we plan to validate our approach with physical networks.

% NOTE: Research questions:
% - Which settings would improve learning and adaptation to the environment? (i.e. connectivity or not, or keeping one's own function or not)
% - Different regions and speciation? if the environment requires drastically different functions/agents how the system will respond to this? how the agents live in the neighbor would respond to this in different settings?

%% file: 6_appendix_stats.tex
\section{Statistical analysis}
\label{app:stats}

In the following, we report the results of the pairwise comparisons (based on Wilcoxon Rank-Sum test \cite{wilcoxon1992individual}, $\alpha=0.05$), visually represented in a symmetric matrix format, and a post-hoc multiple-comparisons analysis (based on Nemenyi test \cite{nemenyi1962distribution}, $\alpha=0.05$), visually represented with Critical Difference plots. We performed both tests on the five versions of the Embodied Evolution algorithm described in the main text, with different parameter settings, as we tested them on the imitation problem (both $28 \times 28$ scenario and $7 \times 7$ scenarios), the illumination problem (both single parameter and vector representation scenarios), and the distributed model of indoor human presence and activity (for the three rooms and both tasks).

\subsection{Wilcoxon Rank-Sum test}
As for the imitation and illumination problems, in Tables \ref{tab:wilcoxonImitation28}-\ref{tab:wilcoxonIlluminationVector} it can be noted that, apart from a few occasional cases (slightly more frequent in the case of the imitation problem, $7 \times 7$ scenario), the null-hypothesis $H_0$ (statistical equivalence) is rejected in most cases, highlighting the fact that in general the different parameter settings and algorithm versions are indeed different (pairwise) in statistically significant terms. 

Concerning the distributed model problem, in Tables \ref{tab:wilcoxonLocAHO}-\ref{tab:wilcoxonLocCHA} it can be seen that there are ``clusters'' of algorithm versions that are pairwise equivalent (in particular the various parameter settings of XoverRand and XoverBest). It is also confirmed that the two tasks (human presence and human activity) appear more difficult in the case of room B, where most algorithms result equivalent.

\subsection{Nemenyi test}
\begin{figure}[ht!]
 \centering
 \includegraphics[width=\columnwidth]{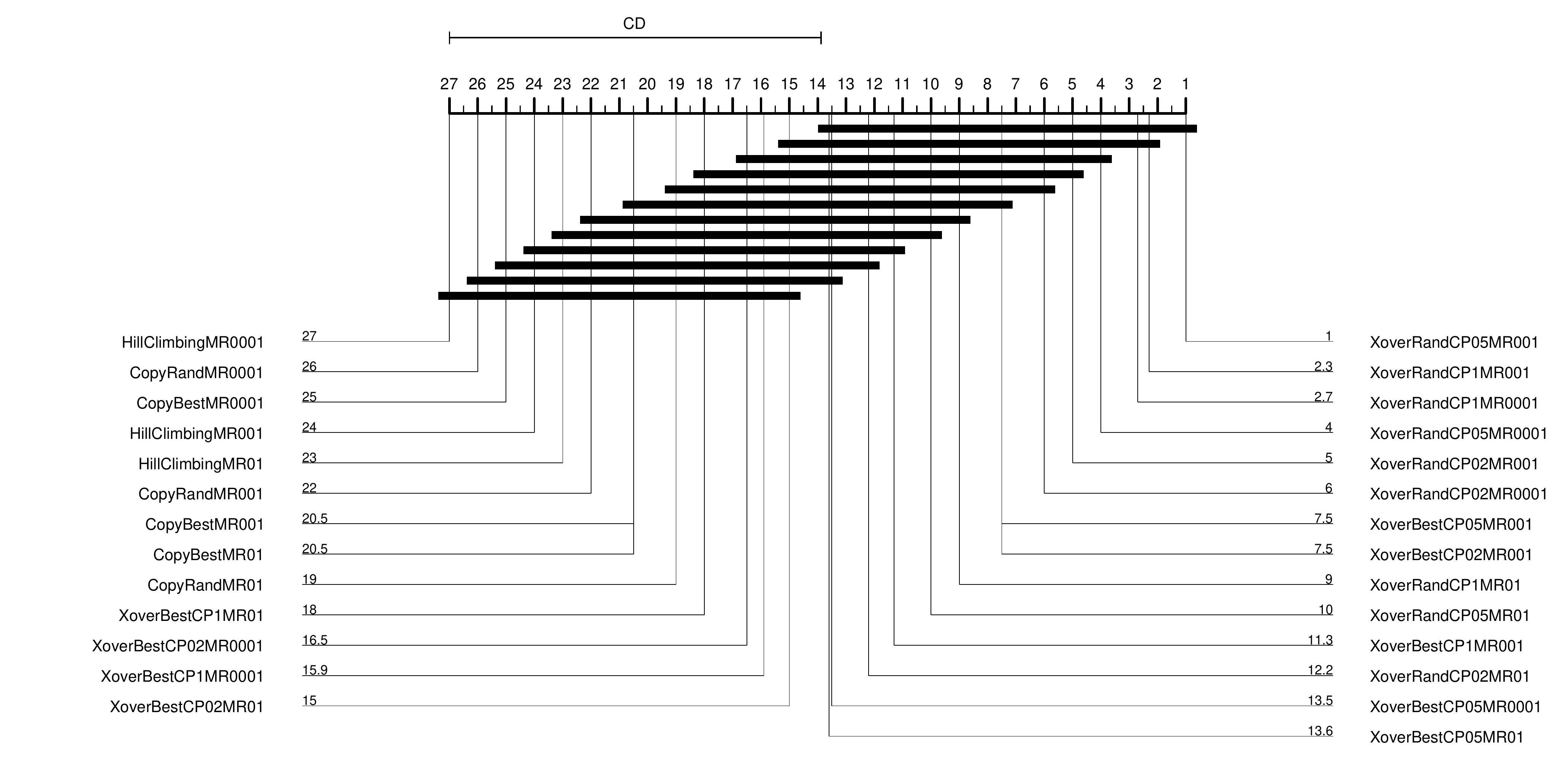}
 \caption{Critical Difference plot based on the Nemenyi post-hoc test ($\alpha = 0.05$) on the algorithm versions tested on the imitation problem ($28 \times 28$ scenario).}
 \label{fig:nemenyiImitation28}
\end{figure}

\begin{figure}[ht!]
 \centering
 \includegraphics[width=\columnwidth]{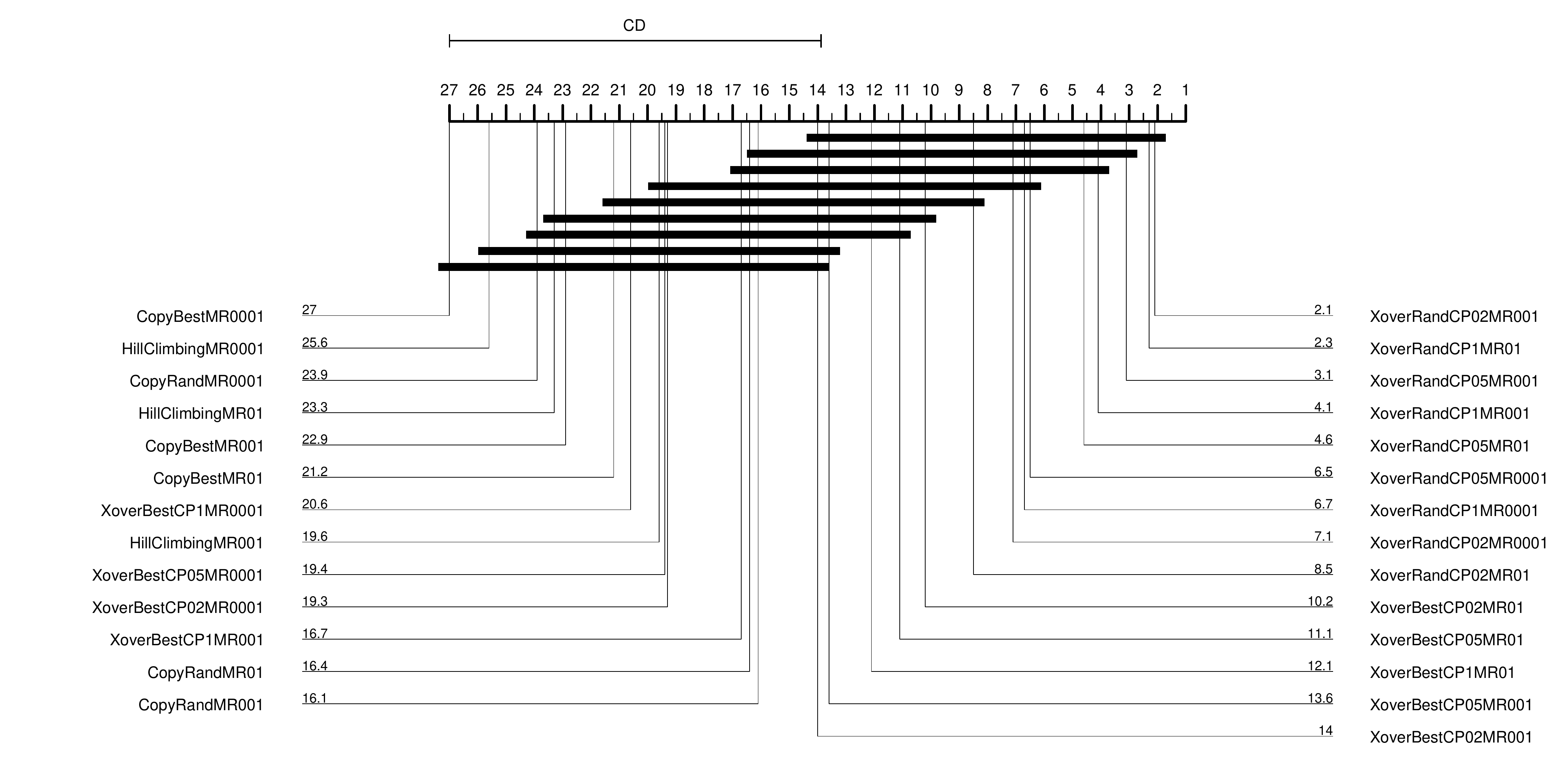}
 \caption{Critical Difference plot based on the Nemenyi post-hoc test ($\alpha = 0.05$) on the algorithm versions tested on the imitation problem ($7 \times 7$ scenario).}
 \label{fig:nemenyiImitation7}
\end{figure}

\begin{figure}[ht!]
 \centering
 \includegraphics[width=\columnwidth]{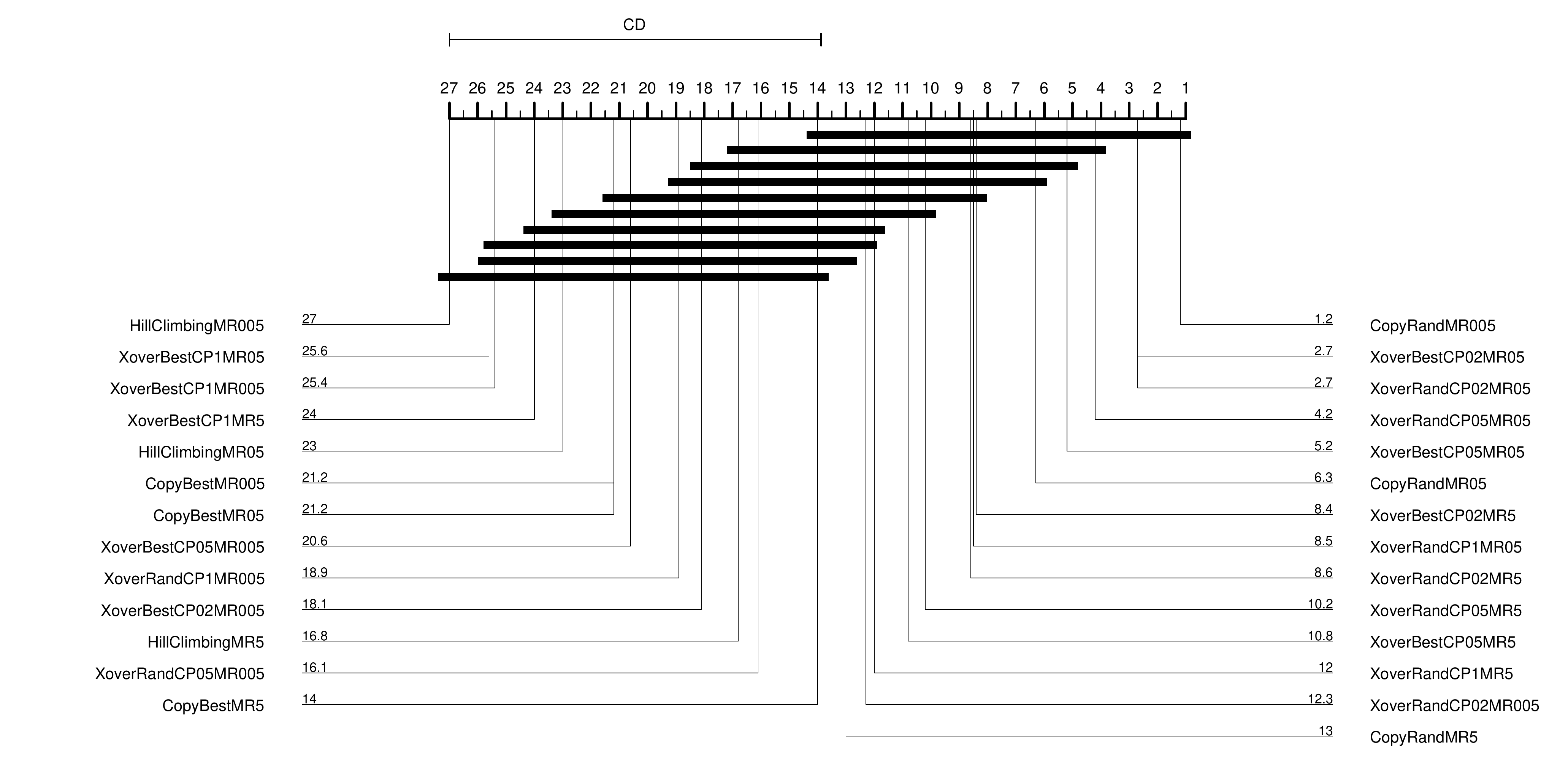}
 \caption{Critical Difference plot based on the Nemenyi post-hoc test ($\alpha = 0.05$) on the algorithm versions tested on the illumination problem (single parameter scenario).}
 \label{fig:nemenyiIlluminationSingle}
\end{figure}

\begin{figure}[ht!]
 \centering
 \includegraphics[width=\columnwidth]{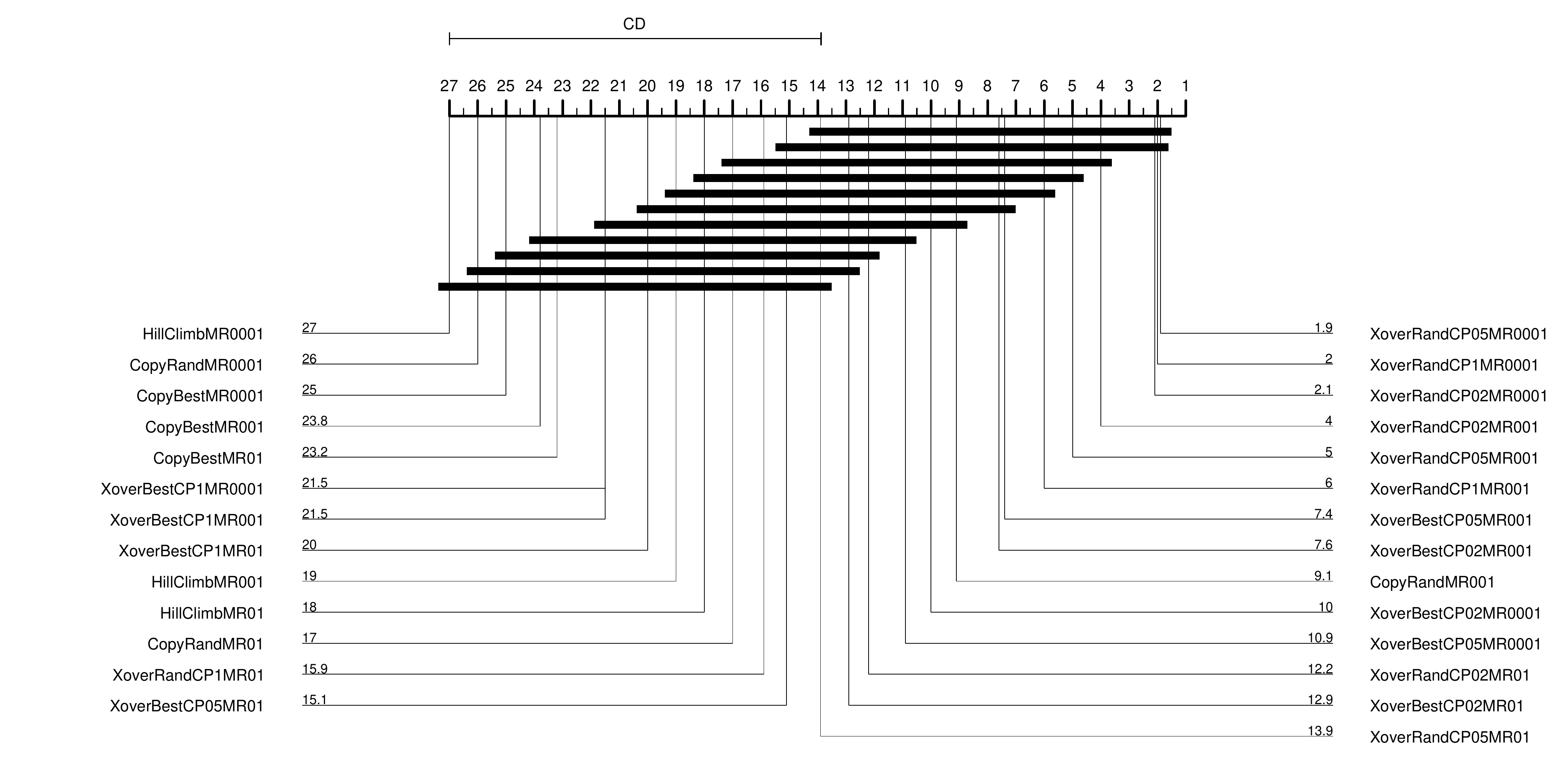}
 \caption{Critical Difference plot based on the Nemenyi post-hoc test ($\alpha = 0.05$) on the algorithm versions tested on the illumination problem (vector representation scenario).}
 \label{fig:nemenyiIlluminationVector}
\end{figure}

\begin{figure}[ht!]
 \centering
 \includegraphics[width=\columnwidth]{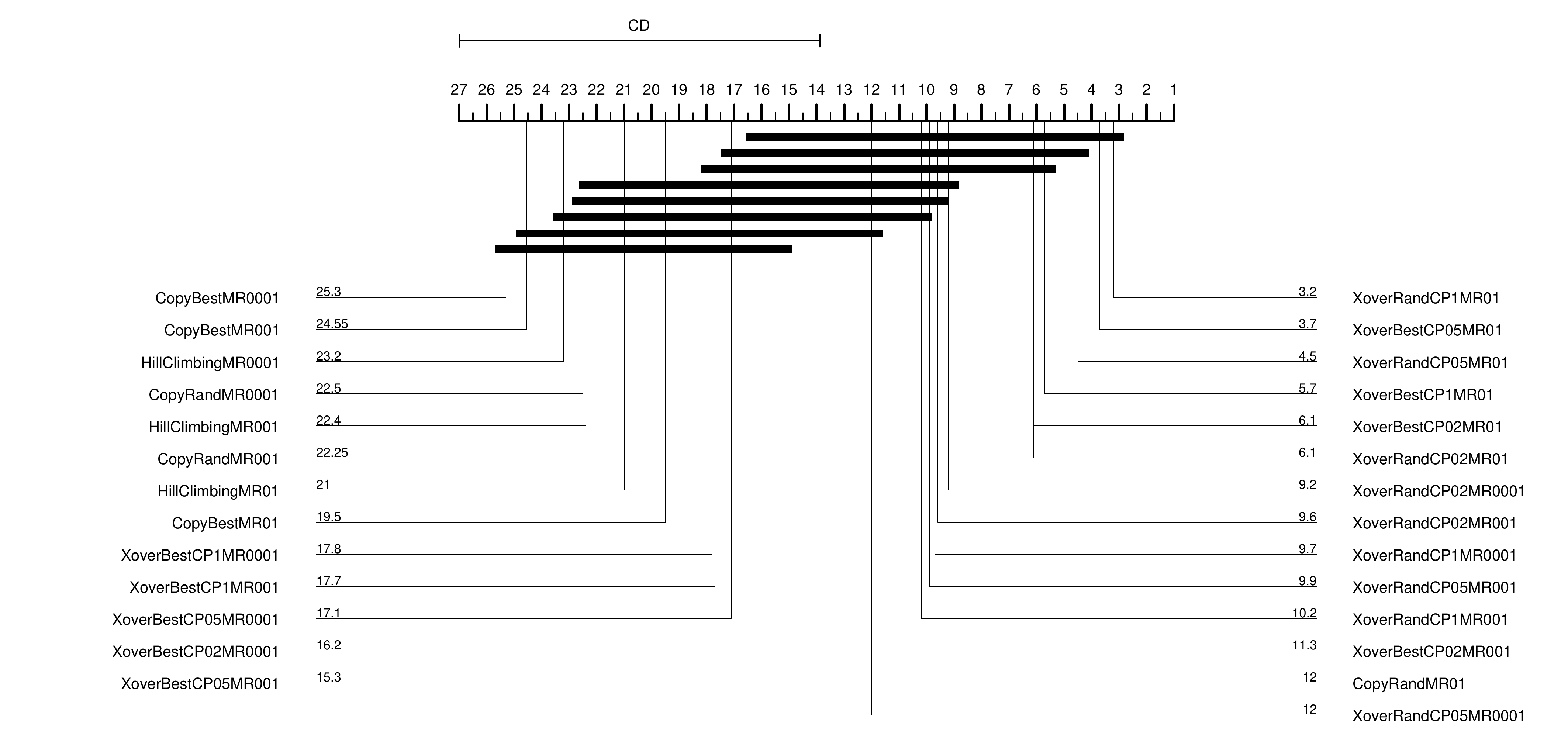}
 \caption{Critical Difference plot based on the Nemenyi post-hoc test ($\alpha = 0.05$) on the algorithm versions tested on the distributed model (room A, human presence task).}
 \label{fig:nemenyiLocAHO}
\end{figure}

\begin{figure}[ht!]
 \centering
 \includegraphics[width=\columnwidth]{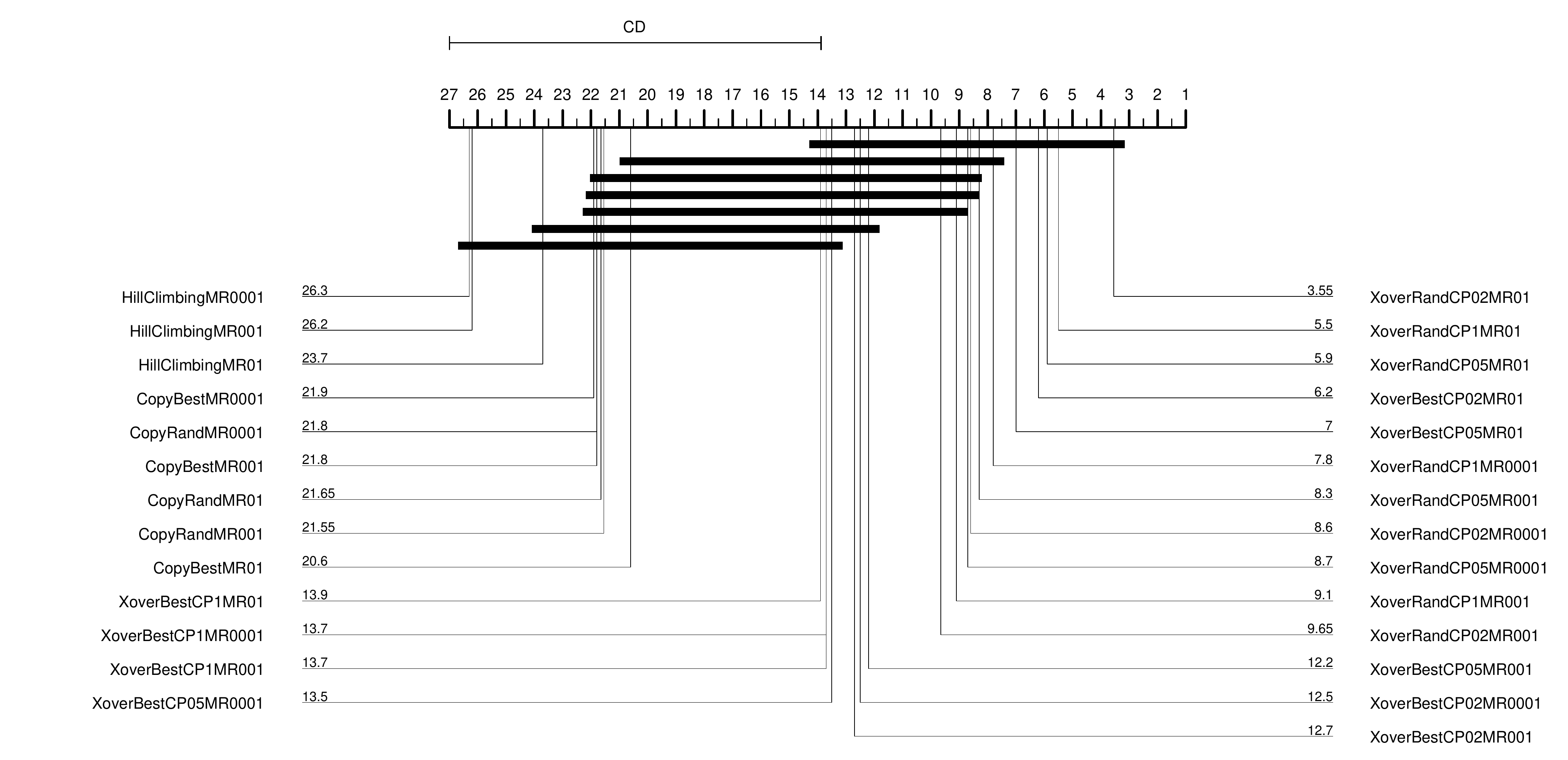}
 \caption{Critical Difference plot based on the Nemenyi post-hoc test ($\alpha = 0.05$) on the algorithm versions tested on the distributed model (room A, human activity task).}
 \label{fig:nemenyiLocAHA}
\end{figure}

\begin{figure}[ht!]
 \centering
 \includegraphics[width=\columnwidth]{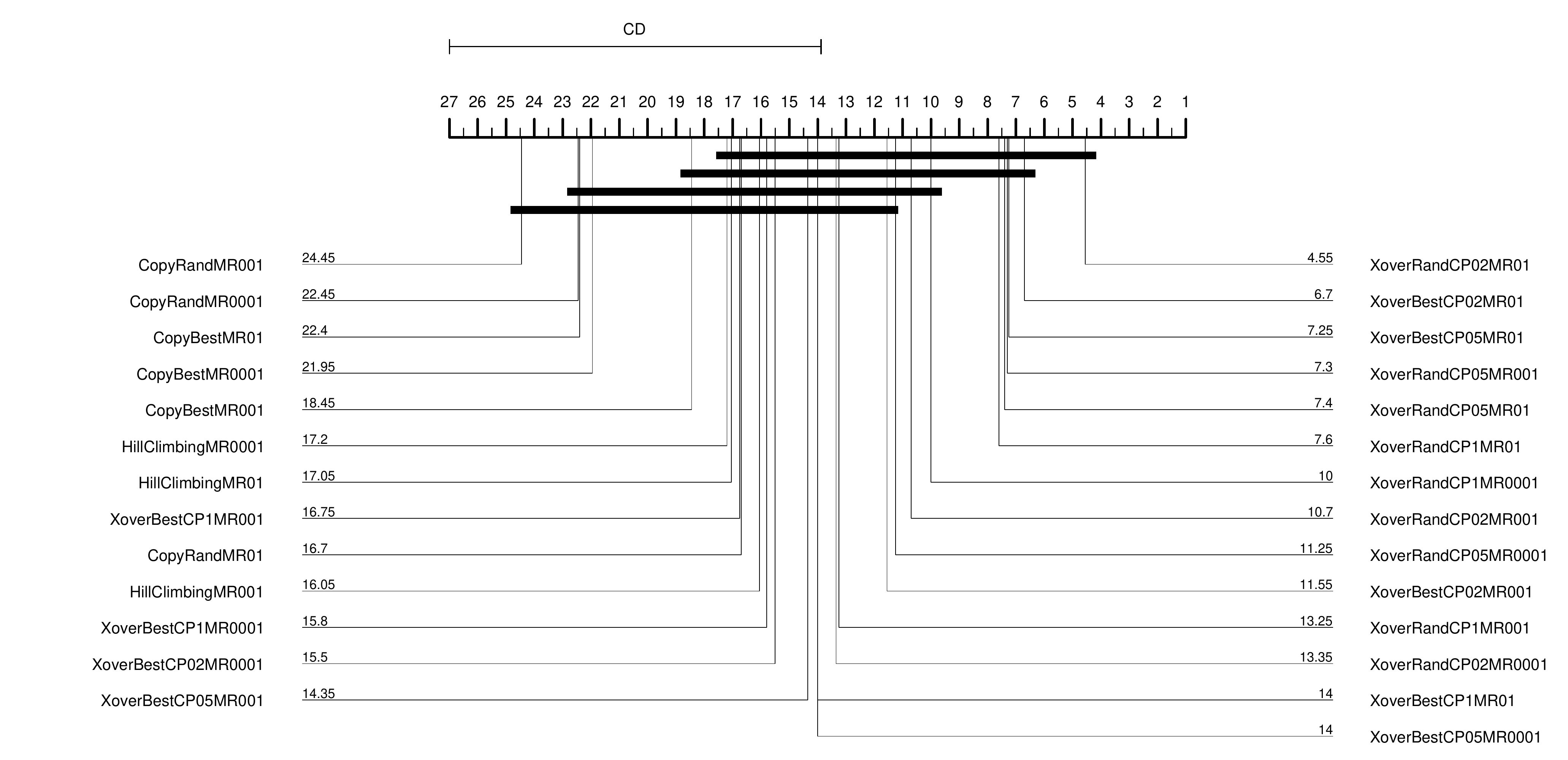}
 \caption{Critical Difference plot based on the Nemenyi post-hoc test ($\alpha = 0.05$) on the algorithm versions tested on the distributed model (room B, human presence task).}
 \label{fig:nemenyiLocBHO}
\end{figure}

\begin{figure}[ht!]
 \centering
 \includegraphics[width=\columnwidth]{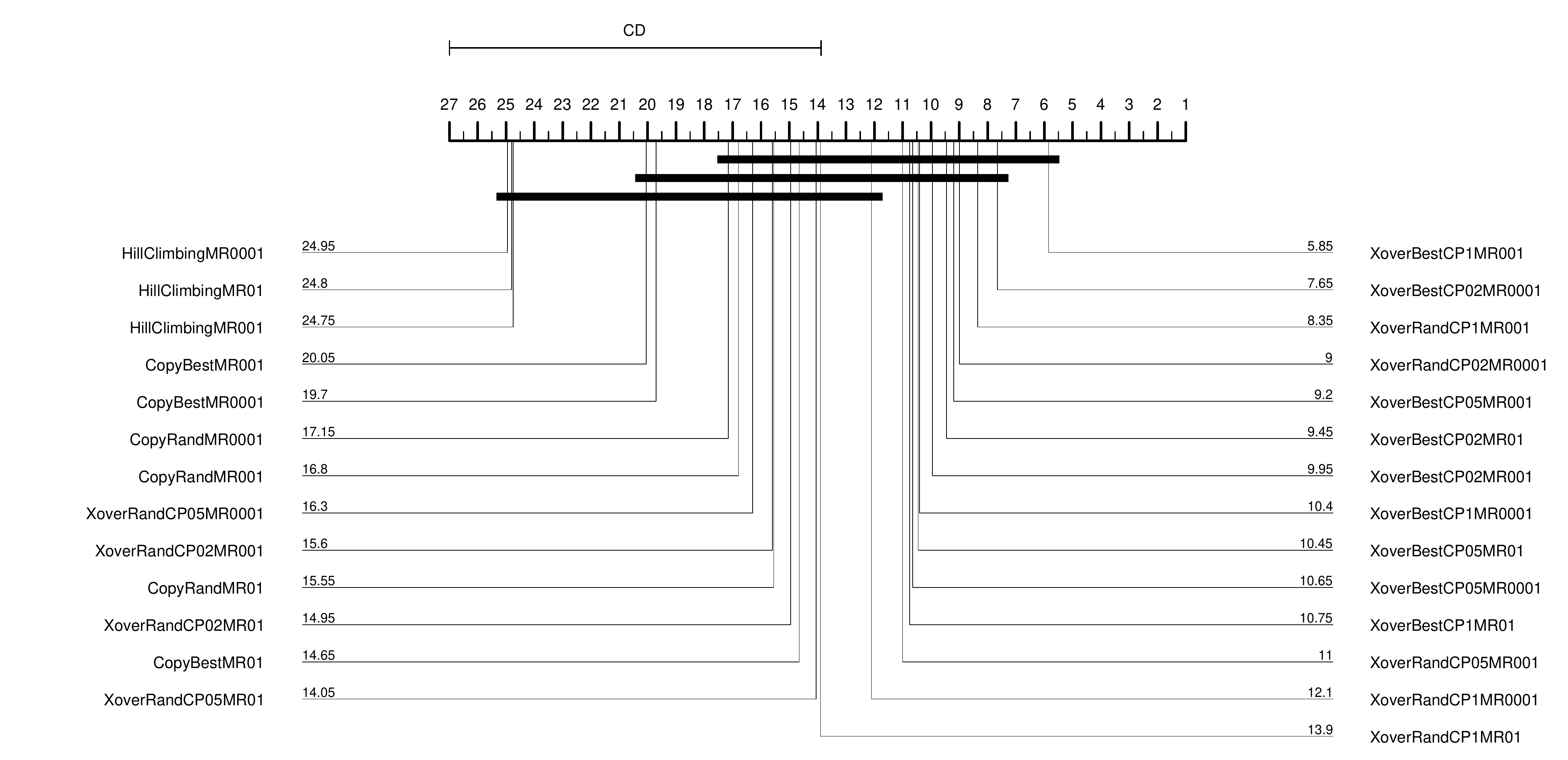}
 \caption{Critical Difference plot based on the Nemenyi post-hoc test ($\alpha = 0.05$) on the algorithm versions tested on the distributed model (room B, human activity task).}
 \label{fig:nemenyiLocBHA}
\end{figure}

\begin{figure}[ht!]
 \centering
 \includegraphics[width=\columnwidth]{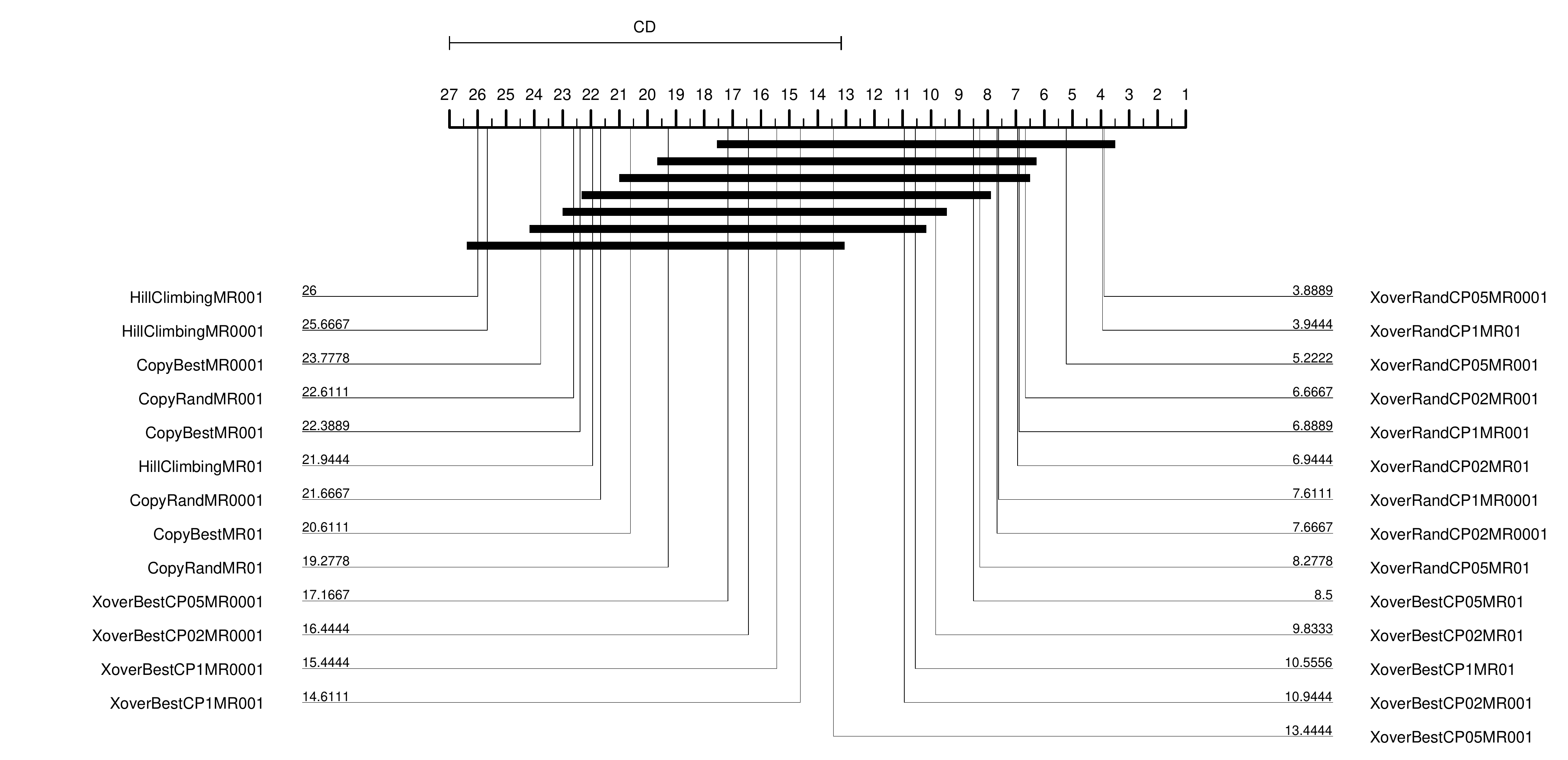}
 \caption{Critical Difference plot based on the Nemenyi post-hoc test ($\alpha = 0.05$) on the algorithm versions tested on the distributed model (room C, human presence task).}
 \label{fig:nemenyiLocCHO}
\end{figure}

\begin{figure}[ht!]
 \centering
 \includegraphics[width=\columnwidth]{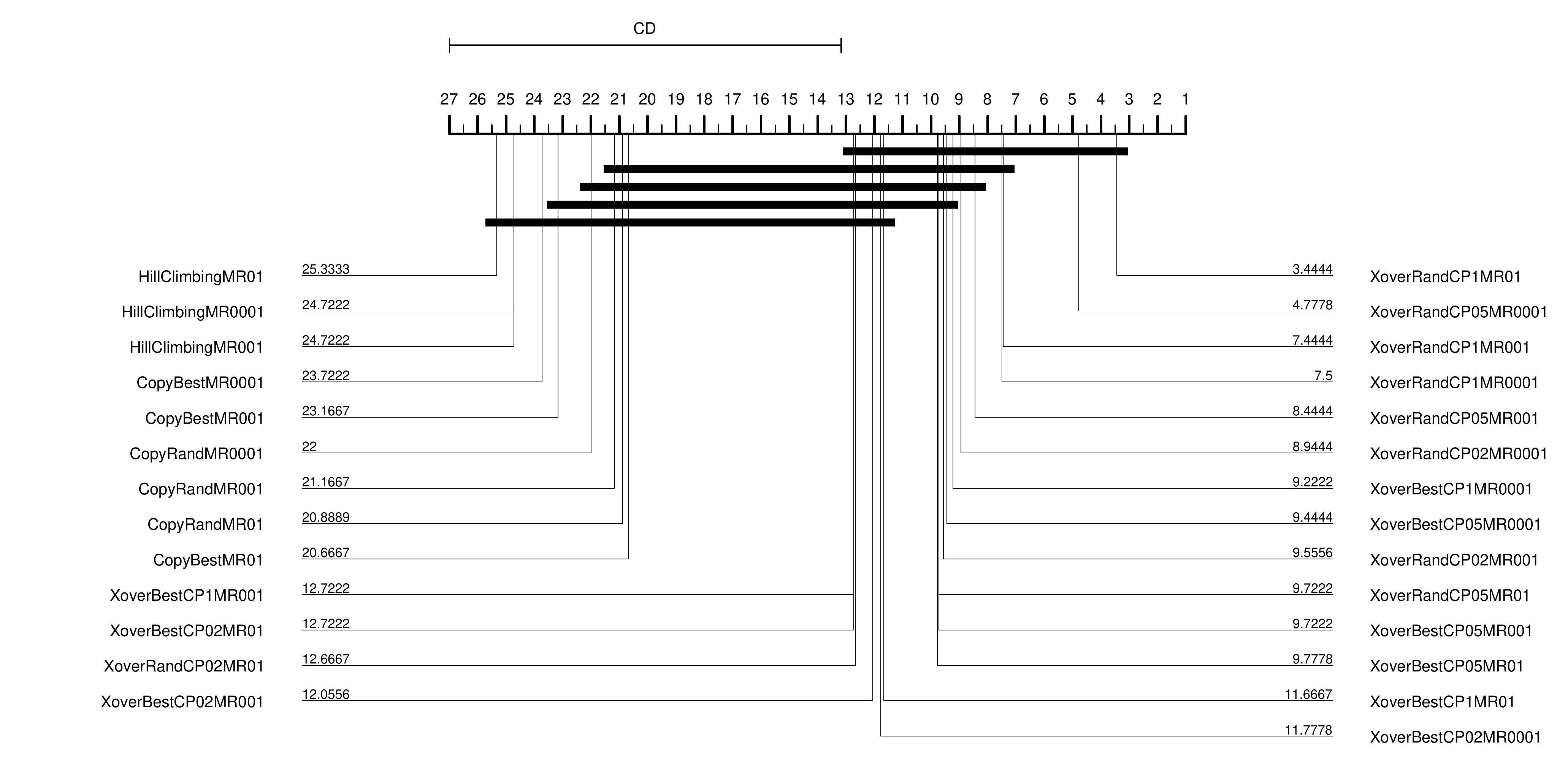}
 \caption{Critical Difference plot based on the Nemenyi post-hoc test ($\alpha = 0.05$) on the algorithm versions tested on the distributed model (room C, human activity task).}
 \label{fig:nemenyiLocCHA}
\end{figure}

In Figures \ref{fig:nemenyiImitation28}-\ref{fig:nemenyiLocCHA}
algorithms that are considered statistically equivalent w.r.t. the Critical Difference (CD) calculated by the Nemenyi test (depicted as a segment on top of the plot) are graphically connected by a thick black line. Better algorithms get a lower rank. With this notation in mind, the main results of the Nemenyi test can be summarized as follows.

\noindent{}As for the imitation and illumination problems, it results that:
\begin{itemize}[leftmargin=*]
\item In Figure~\ref{fig:nemenyiImitation28} (imitation problem, $28 \times 28$ scenario), XoverRand versions consistently rank before XoverBest, CopyRand/CopyBest and HillClimbing versions, in this order. Furthermore, all XoverRand versions and two XoverBest versions result statistically equivalent w.r.t. the CD. The versions without crossover (CopyRand/CopyBest and HillClimbing) perform better (i.e. they get lower ranks) with higher $mr$ values, while the contrary happens on the versions with crossover.
\item In Figure~\ref{fig:nemenyiImitation7} (imitation problem, $7 \times 7$ scenario), the trend is similar to the $28 \times 28$ scenario, with XoverRand versions resulting statistically equivalent w.r.t. their parametrization, and ranking before the other versions (being CopyBest with $mr=0.001$ the worst-performing algorithm). Similar observations on the effect of $mr$ on the versions with/without crossover hold true also in this case.
\item In Figure \ref{fig:nemenyiIlluminationSingle} (illumination problem, single parameter scenario), CopyRand with $mr=0.005$ ranks first, but it results statistically equivalent (w.r.t. the CD) to CopyRand with $mr=0.05$ and $mr=0.5$, as well as XoverBest and XoverRand versions with $mr$ larger than $0.005$. Also in this case, HillClimbing (with the lowest $mr$ value) obtains the worst rank.
\item In Figure \ref{fig:nemenyiIlluminationVector} (illumination problem, vector representation scenario), most of the XoverRand versions rank before XoverBest and the versions without crossover (although the general trend is somehow less clear than the previous cases).
\end{itemize}

\noindent{}As for the distributed model problem, from Figures \ref{fig:nemenyiLocAHO}-\ref{fig:nemenyiLocCHA} it results that, in all cases (except the activity detection task in room B), XoverRand is assigned better ranks w.r.t. the other algorithms. In the case of the human activity detection task in room B, most of the algorithms appear to produce similar results (which, as we have seen in the main text, correspond to low accuracy values).
% \ref{fig:nemenyiLocAHO}-\ref{fig:nemenyiLocAHA}-\ref{fig:nemenyiLocBHO}-\ref{fig:nemenyiLocBHA}-\ref{fig:nemenyiLocCHO}-\ref{fig:nemenyiLocCHA} 

Overall, the above observations obtained from the multiple-comparisons analysis are in line what we discussed in Section \ref{sec:results} of the main text.

\clearpage

%%------------------- Imitation 28x28 case -----------------------

\begin{table}[ht!]
\caption{Wilcoxon Rank-sum test ($\alpha = 0.05$) on the pairwise comparisons between the algorithm versions tested on the imitation problem ($28 \times 28$ scenario). ``='' indicates that the null-hypothesis $H_0$ (statistical equivalence) is accepted (omitted on the diagonal cells). Empty cells indicate that the null-hypothesis is rejected.}
\label{tab:wilcoxonImitation28}
\resizebox{\textwidth}{!}{
\begin{tabular}{|l|P{0.4cm}|P{0.4cm}|P{0.4cm}|P{0.4cm}|P{0.4cm}|P{0.4cm}|P{0.4cm}|P{0.4cm}|P{0.4cm}|P{0.4cm}|P{0.4cm}|P{0.4cm}|P{0.4cm}|P{0.4cm}|P{0.4cm}|P{0.4cm}|P{0.4cm}|P{0.4cm}|P{0.4cm}|P{0.4cm}|P{0.4cm}|P{0.4cm}|P{0.4cm}|P{0.4cm}|P{0.4cm}|P{0.4cm}|P{0.4cm}|P{0.4cm}|}
\hline
 & & \textbf{1} & \textbf{2} & \textbf{3} & \textbf{4} & \textbf{5} & \textbf{6} & \textbf{7} & \textbf{8} & \textbf{9} & \textbf{10} & \textbf{11} & \textbf{12} & \textbf{13} & \textbf{14} & \textbf{15} & \textbf{16} & \textbf{17} & \textbf{18} & \textbf{19} & \textbf{20} & \textbf{21} & \textbf{22} & \textbf{23} & \textbf{24} & \textbf{25} & \textbf{26} & \textbf{27} \\ \hline
\textbf{HillClimbingMR01} & \textbf{1} & & & & & & & & & & & & & & & & & & & & & & & & & & & \\ \hline
\textbf{CopyBestMR01} & \textbf{2} & & & & & & & & & & & = & & & & & & & & & & & & & & & & \\ \hline
\textbf{CopyRandMR01} & \textbf{3} & & & & & & & & & & & & & & & & & & & & & & & & & & & \\ \hline
\textbf{XoverBestCP1MR01} & \textbf{4} & & & & & & & & & & & & & & & & & & & & & & & & & & & \\ \hline
\textbf{XoverBestCP05MR01} & \textbf{5} & & & & & & & & & & & & & & & & & & & & & & & = & & & & \\ \hline
\textbf{XoverBestCP02MR01} & \textbf{6} & & & & & & & & & & & & & & & & & & & & & & & = & & & & \\ \hline
\textbf{XoverRandCP1MR01} & \textbf{7} & & & & & & & & & & & & & & & & & & & & & & & & & & & \\ \hline
\textbf{XoverRandCP05MR01} & \textbf{8} & & & & & & & & & & & & & & & & & & & & & & & & & & & \\ \hline
\textbf{XoverRandCP02MR01} & \textbf{9} & & & & & & & & & & & & & & & & & & & & & & & = & & & & \\ \hline
\textbf{HillClimbingMR001} & \textbf{10} & & & & & & & & & & & & & & & & & & & & & & & & & & & \\ \hline
\textbf{CopyBestMR001} & \textbf{11} & & = & & & & & & & & & & & & & & & & & & & & & & & & & \\ \hline
\textbf{CopyRandMR001} & \textbf{12} & & & & & & & & & & & & & & & & & & & & & & & & & & & \\ \hline
\textbf{XoverBestCP1MR001} & \textbf{13} & & & & & & & & & & & & & & & & & & & & & & & & & & & \\ \hline
\textbf{XoverBestCP05MR001} & \textbf{14} & & & & & & & & & & & & & & & = & & & & & & & & & & & & \\ \hline
\textbf{XoverBestCP02MR001} & \textbf{15} & & & & & & & & & & & & & & = & & & & & & & & & & & & & \\ \hline
\textbf{XoverRandCP1MR001} & \textbf{16} & & & & & & & & & & & & & & & & & & & & & & & & & = & & \\ \hline
\textbf{XoverRandCP05MR001} & \textbf{17} & & & & & & & & & & & & & & & & & & & & & & & & & & & \\ \hline
\textbf{XoverRandCP02MR001} & \textbf{18} & & & & & & & & & & & & & & & & & & & & & & & & & & & \\ \hline
\textbf{HillClimbingMR0001} & \textbf{19} & & & & & & & & & & & & & & & & & & & & & & & & & & & \\ \hline
\textbf{CopyBestMR0001} & \textbf{20} & & & & & & & & & & & & & & & & & & & & & & & & & & & \\ \hline
\textbf{CopyRandMR0001} & \textbf{21} & & & & & & & & & & & & & & & & & & & & & & & & & & & \\ \hline
\textbf{XoverBestCP1MR0001} & \textbf{22} & & & & & & & & & & & & & & & & & & & & & & & & = & & & \\ \hline
\textbf{XoverBestCP05MR0001} & \textbf{23} & & & & & = & = & & & = & & & & & & & & & & & & & & & & & & \\ \hline
\textbf{XoverBestCP02MR0001} & \textbf{24} & & & & & & & & & & & & & & & & & & & & & & = & & & & & \\ \hline
\textbf{XoverRandCP1MR0001} & \textbf{25} & & & & & & & & & & & & & & & & = & & & & & & & & & & & \\ \hline
\textbf{XoverRandCP05MR0001} & \textbf{26} & & & & & & & & & & & & & & & & & & & & & & & & & & & \\ \hline
\textbf{XoverRandCP02MR0001} & \textbf{27} & & & & & & & & & & & & & & & & & & & & & & & & & & & \\ \hline
\end{tabular}
}
\end{table}

%%------------------- Imitation 7x7 case -------------------------

\begin{table}[ht!]
\caption{Wilcoxon Rank-sum test ($\alpha = 0.05$) on the pairwise comparisons between the algorithm versions tested on the imitation problem ($7 \times 7$ scenario). ``='' indicates that the null-hypothesis $H_0$ (statistical equivalence) is accepted (omitted on the diagonal cells). Empty cells indicate that the null-hypothesis is rejected.}
\label{tab:wilcoxonImitation7}
\resizebox{\textwidth}{!}{
\begin{tabular}{|l|P{0.4cm}|P{0.4cm}|P{0.4cm}|P{0.4cm}|P{0.4cm}|P{0.4cm}|P{0.4cm}|P{0.4cm}|P{0.4cm}|P{0.4cm}|P{0.4cm}|P{0.4cm}|P{0.4cm}|P{0.4cm}|P{0.4cm}|P{0.4cm}|P{0.4cm}|P{0.4cm}|P{0.4cm}|P{0.4cm}|P{0.4cm}|P{0.4cm}|P{0.4cm}|P{0.4cm}|P{0.4cm}|P{0.4cm}|P{0.4cm}|P{0.4cm}|}
\hline
 & & \textbf{1} & \textbf{2} & \textbf{3} & \textbf{4} & \textbf{5} & \textbf{6} & \textbf{7} & \textbf{8} & \textbf{9} & \textbf{10} & \textbf{11} & \textbf{12} & \textbf{13} & \textbf{14} & \textbf{15} & \textbf{16} & \textbf{17} & \textbf{18} & \textbf{19} & \textbf{20} & \textbf{21} & \textbf{22} & \textbf{23} & \textbf{24} & \textbf{25} & \textbf{26} & \textbf{27} \\ \hline
\textbf{HillClimbingMR01} & \textbf{1} & & & & & & & & & & & = & & & & & & & & & & = & & & & & & \\ \hline
\textbf{CopyBestMR01} & \textbf{2} & & & & & & & & & & = & & & & & & & & & & & & = & = & = & & & \\ \hline
\textbf{CopyRandMR01} & \textbf{3} & & & & & & & & & & & & = & = & & & & & & & & & & = & = & & & \\ \hline
\textbf{XoverBestCP1MR01} & \textbf{4} & & & & & & & & & & & & & & & & & & & & & & & & & & & \\ \hline
\textbf{XoverBestCP05MR01} & \textbf{5} & & & & & & = & & & & & & & & & & & & & & & & & & & & & \\ \hline
\textbf{XoverBestCP02MR01} & \textbf{6} & & & & & = & & & & & & & & & & & & & & & & & & & & & & \\ \hline
\textbf{XoverRandCP1MR01} & \textbf{7} & & & & & & & & & & & & & & & & = & = & = & & & & & & & & & \\ \hline
\textbf{XoverRandCP05MR01} & \textbf{8} & & & & & & & & & & & & & & & & = & & & & & & & & & & & \\ \hline
\textbf{XoverRandCP02MR01} & \textbf{9} & & & & & & & & & & & & & & & & & & & & & & & & & & & = \\ \hline
\textbf{HillClimbingMR001} & \textbf{10} & & = & & & & & & & & & & & & & & & & & & & & = & = & = & & & \\ \hline
\textbf{CopyBestMR001} & \textbf{11} & = & & & & & & & & & & & & & & & & & & & & = & & & & & & \\ \hline
\textbf{CopyRandMR001} & \textbf{12} & & & = & & & & & & & & & & = & & & & & & & & & & & & & & \\ \hline
\textbf{XoverBestCP1MR001} & \textbf{13} & & & = & & & & & & & & & = & & & & & & & & & & & = & = & & & \\ \hline
\textbf{XoverBestCP05MR001} & \textbf{14} & & & & & & & & & & & & & & & = & & & & & & & & & & & & \\ \hline
\textbf{XoverBestCP02MR001} & \textbf{15} & & & & & & & & & & & & & & = & & & & & & & & & & & & & \\ \hline
\textbf{XoverRandCP1MR001} & \textbf{16} & & & & & & & = & = & & & & & & & & & = & = & & & & & & & & & \\ \hline
\textbf{XoverRandCP05MR001} & \textbf{17} & & & & & & & = & & & & & & & & & = & & = & & & & & & & & & \\ \hline
\textbf{XoverRandCP02MR001} & \textbf{18} & & & & & & & = & & & & & & & & & = & = & & & & & & & & & & \\ \hline
\textbf{HillClimbingMR0001} & \textbf{19} & & & & & & & & & & & & & & & & & & & & & & & & & & & \\ \hline
\textbf{CopyBestMR0001} & \textbf{20} & & & & & & & & & & & & & & & & & & & & & & & & & & & \\ \hline
\textbf{CopyRandMR0001} & \textbf{21} & = & & & & & & & & & & = & & & & & & & & & & & & & & & & \\ \hline
\textbf{XoverBestCP1MR0001} & \textbf{22} & & = & & & & & & & & = & & & & & & & & & & & & & = & = & & & \\ \hline
\textbf{XoverBestCP05MR0001} & \textbf{23} & & = & = & & & & & & & = & & & = & & & & & & & & & = & & = & & & \\ \hline
\textbf{XoverBestCP02MR0001} & \textbf{24} & & = & = & & & & & & & = & & & = & & & & & & & & & = & = & & & & \\ \hline
\textbf{XoverRandCP1MR0001} & \textbf{25} & & & & & & & & & & & & & & & & & & & & & & & & & & = & = \\ \hline
\textbf{XoverRandCP05MR0001} & \textbf{26} & & & & & & & & & & & & & & & & & & & & & & & & & = & & = \\ \hline
\textbf{XoverRandCP02MR0001} & \textbf{27} & & & & & & & & & = & & & & & & & & & & & & & & & & = & = & \\ \hline

\end{tabular}
}
\end{table}

%%------------------- Illumination Single parameter case --------- 

\begin{table}[ht!]
\caption{Wilcoxon Rank-sum test ($\alpha = 0.05$) on the pairwise comparisons between the algorithm versions tested on the illumination problem (single parameter scenario). ``='' indicates that the null-hypothesis $H_0$ (statistical equivalence) is accepted (omitted on the diagonal cells). Empty cells indicate that the null-hypothesis is rejected.}
\label{tab:wilcoxonIlluminationSingle}
\resizebox{\textwidth}{!}{
\begin{tabular}{|l|P{0.4cm}|P{0.4cm}|P{0.4cm}|P{0.4cm}|P{0.4cm}|P{0.4cm}|P{0.4cm}|P{0.4cm}|P{0.4cm}|P{0.4cm}|P{0.4cm}|P{0.4cm}|P{0.4cm}|P{0.4cm}|P{0.4cm}|P{0.4cm}|P{0.4cm}|P{0.4cm}|P{0.4cm}|P{0.4cm}|P{0.4cm}|P{0.4cm}|P{0.4cm}|P{0.4cm}|P{0.4cm}|P{0.4cm}|P{0.4cm}|P{0.4cm}|}
\hline
 & & \textbf{1} & \textbf{2} & \textbf{3} & \textbf{4} & \textbf{5} & \textbf{6} & \textbf{7} & \textbf{8} & \textbf{9} & \textbf{10} & \textbf{11} & \textbf{12} & \textbf{13} & \textbf{14} & \textbf{15} & \textbf{16} & \textbf{17} & \textbf{18} & \textbf{19} & \textbf{20} & \textbf{21} & \textbf{22} & \textbf{23} & \textbf{24} & \textbf{25} & \textbf{26} & \textbf{27} \\ \hline
\textbf{HillClimbingMR05} & \textbf{1} & & & & & & & & & & & & & & & & & & & & & & & & & & & \\ \hline
\textbf{CopyBestMR05} & \textbf{2} & & & & & & & & & & & & & & & & & & & & & & & & & & & \\ \hline
\textbf{CopyRandMR05} & \textbf{3} & & & & & & & & & & & & & & & & & & & & & & & & & & & \\ \hline
\textbf{XoverBestCP1MR05} & \textbf{4} & & & & & & & & & & & & & & & & & & & & & & & & & & & \\ \hline
\textbf{XoverBestCP05MR05} & \textbf{5} & & & & & & & & & & & & & & & & & & & & & & & & & & & \\ \hline
\textbf{XoverBestCP02MR05} & \textbf{6} & & & & & & & & & = & & & & & & & & & & & & & & & & & & \\ \hline
\textbf{XoverRandCP1MR05} & \textbf{7} & & & & & & & & & & & & & & & & & & & & & & & & & & & \\ \hline
\textbf{XoverRandCP05MR05} & \textbf{8} & & & & & & & & & & & & & & & & & & & & & & & & & & & \\ \hline
\textbf{XoverRandCP02MR05} & \textbf{9} & & & & & & = & & & & & & & & & & & & & & & & & & & & & \\ \hline
\textbf{HillClimbingMR005} & \textbf{10} & & & & & & & & & & & & & & & & & & & & & & & & & & & \\ \hline
\textbf{CopyBestMR005} & \textbf{11} & & & & & & & & & & & & & & & & & & & & = & & & = & & & & \\ \hline
\textbf{CopyRandMR005} & \textbf{12} & & & & & & & & & & & & & & & & & & & & & & & & & & & \\ \hline
\textbf{XoverBestCP1MR005} & \textbf{13} & & & & & & & & & & & & & & & & & & & & & & = & & & & & \\ \hline
\textbf{XoverBestCP05MR005} & \textbf{14} & & & & & & & & & & & & & & & & & & & & & & & & & & & \\ \hline
\textbf{XoverBestCP02MR005} & \textbf{15} & & & & & & & & & & & & & & & & & & = & & & & & & & & & \\ \hline
\textbf{XoverRandCP1MR005} & \textbf{16} & & & & & & & & & & & & & & & & & & & & & & & & & & & \\ \hline
\textbf{XoverRandCP05MR005} & \textbf{17} & & & & & & & & & & & & & & & & & & & & & & & & & & & \\ \hline
\textbf{XoverRandCP02MR005} & \textbf{18} & & & & & & & & & & & & & & & = & & & & & & & & & & & & \\ \hline
\textbf{HillClimbingMR0005} & \textbf{19} & & & & & & & & & & & & & & & & & & & & & & & & & & & \\ \hline
\textbf{CopyBestMR0005} & \textbf{20} & & & & & & & & & & & = & & & & & & & & & & & & = & & & & \\ \hline
\textbf{CopyRandMR0005} & \textbf{21} & & & & & & & & & & & & & & & & & & & & & & & & & & & \\ \hline
\textbf{XoverBestCP1MR0005} & \textbf{22} & & & & & & & & & & & & & = & & & & & & & & & & & & & & \\ \hline
\textbf{XoverBestCP05MR0005} & \textbf{23} & & & & & & & & & & & = & & & & & & & & & = & & & & & & & \\ \hline
\textbf{XoverBestCP02MR0005} & \textbf{24} & & & & & & & & & & & & & & & & & & & & & & & & & & & \\ \hline
\textbf{XoverRandCP1MR0005} & \textbf{25} & & & & & & & & & & & & & & & & & & & & & & & & & & & \\ \hline
\textbf{XoverRandCP05MR0005} & \textbf{26} & & & & & & & & & & & & & & & & & & & & & & & & & & & \\ \hline
\textbf{XoverRandCP02MR0005} & \textbf{27} & & & & & & & & & & & & & & & & & & & & & & & & & & & \\ \hline
\end{tabular}
}
\end{table}

%%------------------- Illumination Vector representation case ---- 

\begin{table}[ht!]
\caption{Wilcoxon Rank-sum test ($\alpha = 0.05$) on the pairwise comparisons between the algorithm versions tested on the illumination problem (vector representation scenario). ``='' indicates that the null-hypothesis $H_0$ (statistical equivalence) is accepted (omitted on the diagonal cells). Empty cells indicate that the null-hypothesis is rejected.}
\label{tab:wilcoxonIlluminationVector}
\resizebox{\textwidth}{!}{
\begin{tabular}{|l|P{0.4cm}|P{0.4cm}|P{0.4cm}|P{0.4cm}|P{0.4cm}|P{0.4cm}|P{0.4cm}|P{0.4cm}|P{0.4cm}|P{0.4cm}|P{0.4cm}|P{0.4cm}|P{0.4cm}|P{0.4cm}|P{0.4cm}|P{0.4cm}|P{0.4cm}|P{0.4cm}|P{0.4cm}|P{0.4cm}|P{0.4cm}|P{0.4cm}|P{0.4cm}|P{0.4cm}|P{0.4cm}|P{0.4cm}|P{0.4cm}|P{0.4cm}|}
\hline
 & & \textbf{1} & \textbf{2} & \textbf{3} & \textbf{4} & \textbf{5} & \textbf{6} & \textbf{7} & \textbf{8} & \textbf{9} & \textbf{10} & \textbf{11} & \textbf{12} & \textbf{13} & \textbf{14} & \textbf{15} & \textbf{16} & \textbf{17} & \textbf{18} & \textbf{19} & \textbf{20} & \textbf{21} & \textbf{22} & \textbf{23} & \textbf{24} & \textbf{25} & \textbf{26} & \textbf{27} \\ \hline
\textbf{HillClimbingMR01} & \textbf{1} & & & & & & & & & & & & & & & & & & & & & & & & & & & \\ \hline
\textbf{CopyBestMR01} & \textbf{2} & & & & & & & & & & & & & & & & & & & & & & & & & & & \\ \hline
\textbf{CopyRandMR01} & \textbf{3} & & & & & & & & & & & & & & & & & & & & & & & & & & & \\ \hline
\textbf{XoverBestCP1MR01} & \textbf{4} & & & & & & & & & & & & & & & & & & & & & & & & & & & \\ \hline
\textbf{XoverBestCP05MR01} & \textbf{5} & & & & & & & & & & & & & & & & & & & & & & & & & & & \\ \hline
\textbf{XoverBestCP02MR01} & \textbf{6} & & & & & & & & & & & & & & & & & & & & & & & & & & & \\ \hline
\textbf{XoverRandCP1MR01} & \textbf{7} & & & & & & & & & & & & & & & & & & & & & & & & & & & \\ \hline
\textbf{XoverRandCP05MR01} & \textbf{8} & & & & & & & & & & & & & & & & & & & & & & & & & & & \\ \hline
\textbf{XoverRandCP02MR01} & \textbf{9} & & & & & & & & & & & & & & & & & & & & & & & & & & & \\ \hline
\textbf{HillClimbingMR001} & \textbf{10} & & & & & & & & & & & & & & & & & & & & & & & & & & & \\ \hline
\textbf{CopyBestMR001} & \textbf{11} & & & & & & & & & & & & & & & & & & & & & & & & & & & \\ \hline
\textbf{CopyRandMR001} & \textbf{12} & & & & & & & & & & & & & & & & & & & & & & & & = & & & \\ \hline
\textbf{XoverBestCP1MR001} & \textbf{13} & & & & & & & & & & & & & & & & & & & & & & = & & & & & \\ \hline
\textbf{XoverBestCP05MR001} & \textbf{14} & & & & & & & & & & & & & & & = & & & & & & & & & & & & \\ \hline
\textbf{XoverBestCP02MR001} & \textbf{15} & & & & & & & & & & & & & & = & & & & & & & & & & & & & \\ \hline
\textbf{XoverRandCP1MR001} & \textbf{16} & & & & & & & & & & & & & & & & & & & & & & & & & & & \\ \hline
\textbf{XoverRandCP05MR001} & \textbf{17} & & & & & & & & & & & & & & & & & & & & & & & & & & & \\ \hline
\textbf{XoverRandCP02MR001} & \textbf{18} & & & & & & & & & & & & & & & & & & & & & & & & & & & \\ \hline
\textbf{HillClimbingMR0001} & \textbf{19} & & & & & & & & & & & & & & & & & & & & & & & & & & & \\ \hline
\textbf{CopyBestMR0001} & \textbf{20} & & & & & & & & & & & & & & & & & & & & & & & & & & & \\ \hline
\textbf{CopyRandMR0001} & \textbf{21} & & & & & & & & & & & & & & & & & & & & & & & & & & & \\ \hline
\textbf{XoverBestCP1MR0001} & \textbf{22} & & & & & & & & & & & & & = & & & & & & & & & & & & & & \\ \hline
\textbf{XoverBestCP05MR0001} & \textbf{23} & & & & & & & & & & & & & & & & & & & & & & & & & & & \\ \hline
\textbf{XoverBestCP02MR0001} & \textbf{24} & & & & & & & & & & & & = & & & & & & & & & & & & & & & \\ \hline
\textbf{XoverRandCP1MR0001} & \textbf{25} & & & & & & & & & & & & & & & & & & & & & & & & & & = & = \\ \hline
\textbf{XoverRandCP05MR0001} & \textbf{26} & & & & & & & & & & & & & & & & & & & & & & & & & = & & = \\ \hline
\textbf{XoverRandCP02MR0001} & \textbf{27} & & & & & & & & & & & & & & & & & & & & & & & & & = & = & \\ \hline
\end{tabular}
}
\end{table}

%%------------------- LOCATION A Human Presence detection --------

\begin{table}[ht!]
\caption{Wilcoxon Rank-sum test ($\alpha = 0.05$) on the pairwise comparisons between the algorithm versions tested on the distributed model (room A, human presence task). ``='' indicates that the null-hypothesis $H_0$ (statistical equivalence) is accepted (omitted on the diagonal cells). Empty cells indicate that the null-hypothesis is rejected.}
\label{tab:wilcoxonLocAHO}
\resizebox{\textwidth}{!}{
\begin{tabular}{|l|P{0.4cm}|P{0.4cm}|P{0.4cm}|P{0.4cm}|P{0.4cm}|P{0.4cm}|P{0.4cm}|P{0.4cm}|P{0.4cm}|P{0.4cm}|P{0.4cm}|P{0.4cm}|P{0.4cm}|P{0.4cm}|P{0.4cm}|P{0.4cm}|P{0.4cm}|P{0.4cm}|P{0.4cm}|P{0.4cm}|P{0.4cm}|P{0.4cm}|P{0.4cm}|P{0.4cm}|P{0.4cm}|P{0.4cm}|P{0.4cm}|P{0.4cm}|}
\hline
 & & \textbf{1} & \textbf{2} & \textbf{3} & \textbf{4} & \textbf{5} & \textbf{6} & \textbf{7} & \textbf{8} & \textbf{9} & \textbf{10} & \textbf{11} & \textbf{12} & \textbf{13} & \textbf{14} & \textbf{15} & \textbf{16} & \textbf{17} & \textbf{18} & \textbf{19} & \textbf{20} & \textbf{21} & \textbf{22} & \textbf{23} & \textbf{24} & \textbf{25} & \textbf{26} & \textbf{27} \\ \hline
\textbf{HillClimbingMR01} & \textbf{1} & & = & & & & & & & & = & & = & & & & & & & = & & = & & & & & & \\ \hline
\textbf{CopyBestMR01} & \textbf{2} & = & & & & & & & & & = & = & = & = & = & & & & & = & & = & = & = & = & & & \\ \hline
\textbf{CopyRandMR01} & \textbf{3} & & & & & & & & & & & & & & = & = & = & = & = & & & & & = & & = & = & = \\ \hline
\textbf{XoverBestCP1MR01} & \textbf{4} & & & & & = & = & = & = & = & & & & & & & & & = & & & & & & & & & = \\ \hline
\textbf{XoverBestCP05MR01} & \textbf{5} & & & & = & & = & = & = & = & & & & & & & & & = & & & & & & & & & \\ \hline
\textbf{XoverBestCP02MR01} & \textbf{6} & & & & = & = & & = & = & = & & & & & & & = & & = & & & & & & & = & & = \\ \hline
\textbf{XoverRandCP1MR01} & \textbf{7} & & & & = & = & = & & = & & & & & & & & & & & & & & & & & & & \\ \hline
\textbf{XoverRandCP05MR01} & \textbf{8} & & & & = & = & = & = & & = & & & & & & & & & = & & & & & & & & & \\ \hline
\textbf{XoverRandCP02MR01} & \textbf{9} & & & & = & = & = & & = & & & & & & & & = & = & = & & & & & & & = & & = \\ \hline
\textbf{HillClimbingMR001} & \textbf{10} & = & = & & & & & & & & & & = & & & & & & & = & & = & & & & & & \\ \hline
\textbf{CopyBestMR001} & \textbf{11} & & = & & & & & & & & & & = & & & & & & & = & = & = & & & & & & \\ \hline
\textbf{CopyRandMR001} & \textbf{12} & = & = & & & & & & & & = & = & & & & & & & & = & & = & & & & & & \\ \hline
\textbf{XoverBestCP1MR001} & \textbf{13} & & = & & & & & & & & & & & & = & & & & & & & & = & = & = & & & \\ \hline
\textbf{XoverBestCP05MR001} & \textbf{14} & & = & = & & & & & & & & & & = & & = & & & = & & & & = & = & = & & = & \\ \hline
\textbf{XoverBestCP02MR001} & \textbf{15} & & & = & & & & & & & & & & & = & & = & = & = & & & & & & & = & = & = \\ \hline
\textbf{XoverRandCP1MR001} & \textbf{16} & & & = & & & = & & & = & & & & & & = & & = & = & & & & & & & = & = & = \\ \hline
\textbf{XoverRandCP05MR001} & \textbf{17} & & & = & & & & & & = & & & & & & = & = & & = & & & & & & & = & = & = \\ \hline
\textbf{XoverRandCP02MR001} & \textbf{18} & & & = & = & = & = & & = & = & & & & & = & = & = & = & & & & & & & & = & = & = \\ \hline
\textbf{HillClimbingMR0001} & \textbf{19} & = & = & & & & & & & & = & = & = & & & & & & & & & = & & & & & & \\ \hline
\textbf{CopyBestMR0001} & \textbf{20} & & & & & & & & & & & = & & & & & & & & & & & & & & & & \\ \hline
\textbf{CopyRandMR0001} & \textbf{21} & = & = & & & & & & & & = & = & = & & & & & & & = & & & & & & & & \\ \hline
\textbf{XoverBestCP1MR0001} & \textbf{22} & & = & & & & & & & & & & & = & = & & & & & & & & & = & = & & & \\ \hline
\textbf{XoverBestCP05MR0001} & \textbf{23} & & = & = & & & & & & & & & & = & = & & & & & & & & = & & = & & & \\ \hline
\textbf{XoverBestCP02MR0001} & \textbf{24} & & = & & & & & & & & & & & = & = & & & & & & & & = & = & & & & \\ \hline
\textbf{XoverRandCP1MR0001} & \textbf{25} & & & = & & & = & & & = & & & & & & = & = & = & = & & & & & & & & = & = \\ \hline
\textbf{XoverRandCP05MR0001} & \textbf{26} & & & = & & & & & & & & & & & = & = & = & = & = & & & & & & & = & & = \\ \hline
\textbf{XoverRandCP02MR0001} & \textbf{27} & & & = & = & & = & & & = & & & & & & = & = & = & = & & & & & & & = & = & \\ \hline
\end{tabular}
}
\end{table}

%%------------------- LOCATION A Human Activity detection --------

\begin{table}[ht!]
\caption{Wilcoxon Rank-sum test ($\alpha = 0.05$) on the pairwise comparisons between the algorithm versions tested on the distributed model (room A, human activity task). ``='' indicates that the null-hypothesis $H_0$ (statistical equivalence) is accepted (omitted on the diagonal cells). Empty cells indicate that the null-hypothesis is rejected.}
\label{tab:wilcoxonLocAHA}
\resizebox{\textwidth}{!}{
\begin{tabular}{|l|P{0.4cm}|P{0.4cm}|P{0.4cm}|P{0.4cm}|P{0.4cm}|P{0.4cm}|P{0.4cm}|P{0.4cm}|P{0.4cm}|P{0.4cm}|P{0.4cm}|P{0.4cm}|P{0.4cm}|P{0.4cm}|P{0.4cm}|P{0.4cm}|P{0.4cm}|P{0.4cm}|P{0.4cm}|P{0.4cm}|P{0.4cm}|P{0.4cm}|P{0.4cm}|P{0.4cm}|P{0.4cm}|P{0.4cm}|P{0.4cm}|P{0.4cm}|}
\hline
 & & \textbf{1} & \textbf{2} & \textbf{3} & \textbf{4} & \textbf{5} & \textbf{6} & \textbf{7} & \textbf{8} & \textbf{9} & \textbf{10} & \textbf{11} & \textbf{12} & \textbf{13} & \textbf{14} & \textbf{15} & \textbf{16} & \textbf{17} & \textbf{18} & \textbf{19} & \textbf{20} & \textbf{21} & \textbf{22} & \textbf{23} & \textbf{24} & \textbf{25} & \textbf{26} & \textbf{27} \\ \hline
\textbf{HillClimbingMR01} & \textbf{1} & & & = & & & & & & & & = & = & & & & & & & & & = & & & & & & \\ \hline
\textbf{CopyBestMR01} & \textbf{2} & & & = & & & & & & & & = & = & & & & & & & & = & = & & & & & & \\ \hline
\textbf{CopyRandMR01} & \textbf{3} & = & = & & & & & & & & & = & = & & & & & & & & = & = & & & & & & \\ \hline
\textbf{XoverBestCP1MR01} & \textbf{4} & & & & & & & & & & & & & = & = & = & & & = & & & & = & = & = & & & \\ \hline
\textbf{XoverBestCP05MR01} & \textbf{5} & & & & & & = & = & = & = & & & & & & & = & = & = & & & & & & & = & = & = \\ \hline
\textbf{XoverBestCP02MR01} & \textbf{6} & & & & & = & & = & = & = & & & & & & & = & = & = & & & & & & & = & = & = \\ \hline
\textbf{XoverRandCP1MR01} & \textbf{7} & & & & & = & = & & = & = & & & & & & & = & = & = & & & & & & & = & = & = \\ \hline
\textbf{XoverRandCP05MR01} & \textbf{8} & & & & & = & = & = & & = & & & & & & & = & = & = & & & & & & & = & = & = \\ \hline
\textbf{XoverRandCP02MR01} & \textbf{9} & & & & & = & = & = & = & & & & & & & & & & & & & & & & & = & & \\ \hline
\textbf{HillClimbingMR001} & \textbf{10} & & & & & & & & & & & & & & & & & & & = & & & & & & & & \\ \hline
\textbf{CopyBestMR001} & \textbf{11} & = & = & = & & & & & & & & & = & & & & & & & & = & = & & & & & & \\ \hline
\textbf{CopyRandMR001} & \textbf{12} & = & = & = & & & & & & & & = & & & & & & & & & = & = & & & & & & \\ \hline
\textbf{XoverBestCP1MR001} & \textbf{13} & & & & = & & & & & & & & & & = & = & = & & = & & & & = & = & = & & = & \\ \hline
\textbf{XoverBestCP05MR001} & \textbf{14} & & & & = & & & & & & & & & = & & = & = & = & = & & & & = & = & = & = & = & = \\ \hline
\textbf{XoverBestCP02MR001} & \textbf{15} & & & & = & & & & & & & & & = & = & & = & & = & & & & = & = & = & = & = & \\ \hline
\textbf{XoverRandCP1MR001} & \textbf{16} & & & & & = & = & = & = & & & & & = & = & = & & = & = & & & & = & = & = & = & = & = \\ \hline
\textbf{XoverRandCP05MR001} & \textbf{17} & & & & & = & = & = & = & & & & & & = & & = & & = & & & & & & & = & = & = \\ \hline
\textbf{XoverRandCP02MR001} & \textbf{18} & & & & = & = & = & = & = & & & & & = & = & = & = & = & & & & & = & = & = & = & = & = \\ \hline
\textbf{HillClimbingMR0001} & \textbf{19} & & & & & & & & & & = & & & & & & & & & & & & & & & & & \\ \hline
\textbf{CopyBestMR0001} & \textbf{20} & & = & = & & & & & & & & = & = & & & & & & & & & = & & & & & & \\ \hline
\textbf{CopyRandMR0001} & \textbf{21} & = & = & = & & & & & & & & = & = & & & & & & & & = & & & & & & & \\ \hline
\textbf{XoverBestCP1MR0001} & \textbf{22} & & & & = & & & & & & & & & = & = & = & = & & = & & & & & = & = & & = & \\ \hline
\textbf{XoverBestCP05MR0001} & \textbf{23} & & & & = & & & & & & & & & = & = & = & = & & = & & & & = & & = & & = & \\ \hline
\textbf{XoverBestCP02MR0001} & \textbf{24} & & & & = & & & & & & & & & = & = & = & = & & = & & & & = & = & & & = & \\ \hline
\textbf{XoverRandCP1MR0001} & \textbf{25} & & & & & = & = & = & = & = & & & & & = & = & = & = & = & & & & & & & & = & = \\ \hline
\textbf{XoverRandCP05MR0001} & \textbf{26} & & & & & = & = & = & = & & & & & = & = & = & = & = & = & & & & = & = & = & = & & = \\ \hline
\textbf{XoverRandCP02MR0001} & \textbf{27} & & & & & = & = & = & = & & & & & & = & & = & = & = & & & & & & & = & = & \\ \hline
\end{tabular}
}
\end{table}

%%------------------- LOCATION B Human Presence detection --------

\begin{table}[ht!]
\caption{Wilcoxon Rank-sum test ($\alpha = 0.05$) on the pairwise comparisons between the algorithm versions tested on the distributed model (room B, human presence task). ``='' indicates that the null-hypothesis $H_0$ (statistical equivalence) is accepted (omitted on the diagonal cells). Empty cells indicate that the null-hypothesis is rejected.}
\label{tab:wilcoxonLocBHO}
\resizebox{\textwidth}{!}{
\begin{tabular}{|l|P{0.4cm}|P{0.4cm}|P{0.4cm}|P{0.4cm}|P{0.4cm}|P{0.4cm}|P{0.4cm}|P{0.4cm}|P{0.4cm}|P{0.4cm}|P{0.4cm}|P{0.4cm}|P{0.4cm}|P{0.4cm}|P{0.4cm}|P{0.4cm}|P{0.4cm}|P{0.4cm}|P{0.4cm}|P{0.4cm}|P{0.4cm}|P{0.4cm}|P{0.4cm}|P{0.4cm}|P{0.4cm}|P{0.4cm}|P{0.4cm}|P{0.4cm}|}
\hline
 & & \textbf{1} & \textbf{2} & \textbf{3} & \textbf{4} & \textbf{5} & \textbf{6} & \textbf{7} & \textbf{8} & \textbf{9} & \textbf{10} & \textbf{11} & \textbf{12} & \textbf{13} & \textbf{14} & \textbf{15} & \textbf{16} & \textbf{17} & \textbf{18} & \textbf{19} & \textbf{20} & \textbf{21} & \textbf{22} & \textbf{23} & \textbf{24} & \textbf{25} & \textbf{26} & \textbf{27} \\ \hline
\textbf{HillClimbingMR01} & \textbf{1} & & & = & = & & & & & & = & = & & = & = & = & = & & = & = & & & = & = & = & & = & = \\ \hline
\textbf{CopyBestMR01} & \textbf{2} & & & = & & & & & & & & = & = & & & & & & & = & = & = & & & & & & \\ \hline
\textbf{CopyRandMR01} & \textbf{3} & = & = & & = & & & & & & = & = & & = & = & = & = & & = & = & = & = & = & = & = & = & = & = \\ \hline
\textbf{XoverBestCP1MR01} & \textbf{4} & = & & = & & = & = & = & = & & = & = & & = & = & = & = & = & = & = & & & = & = & = & = & = & = \\ \hline
\textbf{XoverBestCP05MR01} & \textbf{5} & & & & = & & = & = & = & = & & & & & & = & = & = & = & & & & & & = & = & = & = \\ \hline
\textbf{XoverBestCP02MR01} & \textbf{6} & & & & = & = & & = & = & = & & & & & & = & & = & = & & & & & & & = & = & = \\ \hline
\textbf{XoverRandCP1MR01} & \textbf{7} & & & & = & = & = & & = & = & & & & & & = & = & = & = & & & & & & & = & = & = \\ \hline
\textbf{XoverRandCP05MR01} & \textbf{8} & & & & = & = & = & = & & = & & & & & & = & & = & = & & & & & & = & = & = & = \\ \hline
\textbf{XoverRandCP02MR01} & \textbf{9} & & & & & = & = & = & = & & & & & & & & & = & = & & & & & & & & & \\ \hline
\textbf{HillClimbingMR001} & \textbf{10} & = & & = & = & & & & & & & = & & = & = & = & = & & = & = & & & = & = & = & & = & = \\ \hline
\textbf{CopyBestMR001} & \textbf{11} & = & = & = & = & & & & & & = & & & = & = & & = & & & = & = & = & = & = & = & & & = \\ \hline
\textbf{CopyRandMR001} & \textbf{12} & & = & & & & & & & & & & & & & & & & & & = & = & & & & & & \\ \hline
\textbf{XoverBestCP1MR001} & \textbf{13} & = & & = & = & & & & & & = & = & & & = & = & = & & = & = & = & & = & = & = & & = & = \\ \hline
\textbf{XoverBestCP05MR001} & \textbf{14} & = & & = & = & & & & & & = & = & & = & & = & = & & = & = & & & = & = & = & = & = & = \\ \hline
\textbf{XoverBestCP02MR001} & \textbf{15} & = & & = & = & = & = & = & = & & = & & & = & = & & = & = & = & = & & & = & = & = & = & = & = \\ \hline
\textbf{XoverRandCP1MR001} & \textbf{16} & = & & = & = & = & & = & & & = & = & & = & = & = & & & = & = & & & = & = & = & = & = & = \\ \hline
\textbf{XoverRandCP05MR001} & \textbf{17} & & & & = & = & = & = & = & = & & & & & & = & & & = & & & & & & & = & = & = \\ \hline
\textbf{XoverRandCP02MR001} & \textbf{18} & = & & = & = & = & = & = & = & = & = & & & = & = & = & = & = & & = & & & = & = & = & = & = & = \\ \hline
\textbf{HillClimbingMR0001} & \textbf{19} & = & = & = & = & & & & & & = & = & & = & = & = & = & & = & & = & = & = & = & = & & = & = \\ \hline
\textbf{CopyBestMR0001} & \textbf{20} & & = & = & & & & & & & & = & = & = & & & & & & = & & = & & & & & & \\ \hline
\textbf{CopyRandMR0001} & \textbf{21} & & = & = & & & & & & & & = & = & & & & & & & = & = & & & & & & & \\ \hline
\textbf{XoverBestCP1MR0001} & \textbf{22} & = & & = & = & & & & & & = & = & & = & = & = & = & & = & = & & & & = & = & & = & = \\ \hline
\textbf{XoverBestCP05MR0001} & \textbf{23} & = & & = & = & & & & & & = & = & & = & = & = & = & & = & = & & & = & & = & = & = & = \\ \hline
\textbf{XoverBestCP02MR0001} & \textbf{24} & = & & = & = & = & & & = & & = & = & & = & = & = & = & & = & = & & & = & = & & = & = & = \\ \hline
\textbf{XoverRandCP1MR0001} & \textbf{25} & & & = & = & = & = & = & = & & & & & & = & = & = & = & = & & & & & = & = & & = & = \\ \hline
\textbf{XoverRandCP05MR0001} & \textbf{26} & = & & = & = & = & = & = & = & & = & & & = & = & = & = & = & = & = & & & = & = & = & = & & = \\ \hline
\textbf{XoverRandCP02MR0001} & \textbf{27} & = & & = & = & = & = & = & = & & = & = & & = & = & = & = & = & = & = & & & = & = & = & = & = & \\ \hline

\end{tabular}
}
\end{table}

%%------------------- LOCATION B Human Activity detection --------

\begin{table}[ht!]
\caption{Wilcoxon Rank-sum test ($\alpha = 0.05$) on the pairwise comparisons between the algorithm versions tested on the distributed model (room B, human activity task). ``='' indicates that the null-hypothesis $H_0$ (statistical equivalence) is accepted (omitted on the diagonal cells). Empty cells indicate that the null-hypothesis is rejected.}
\label{tab:wilcoxonLocBHA}
\resizebox{\textwidth}{!}{
\begin{tabular}{|l|P{0.4cm}|P{0.4cm}|P{0.4cm}|P{0.4cm}|P{0.4cm}|P{0.4cm}|P{0.4cm}|P{0.4cm}|P{0.4cm}|P{0.4cm}|P{0.4cm}|P{0.4cm}|P{0.4cm}|P{0.4cm}|P{0.4cm}|P{0.4cm}|P{0.4cm}|P{0.4cm}|P{0.4cm}|P{0.4cm}|P{0.4cm}|P{0.4cm}|P{0.4cm}|P{0.4cm}|P{0.4cm}|P{0.4cm}|P{0.4cm}|P{0.4cm}|}
\hline
 & & \textbf{1} & \textbf{2} & \textbf{3} & \textbf{4} & \textbf{5} & \textbf{6} & \textbf{7} & \textbf{8} & \textbf{9} & \textbf{10} & \textbf{11} & \textbf{12} & \textbf{13} & \textbf{14} & \textbf{15} & \textbf{16} & \textbf{17} & \textbf{18} & \textbf{19} & \textbf{20} & \textbf{21} & \textbf{22} & \textbf{23} & \textbf{24} & \textbf{25} & \textbf{26} & \textbf{27} \\ \hline
\textbf{HillClimbingMR01} & \textbf{1} & & & & & & & & & & = & & & & & & & & & = & = & & & & & & & \\ \hline
\textbf{CopyBestMR01} & \textbf{2} & & & = & = & = & = & = & = & = & & & = & & & = & = & = & = & & = & = & = & = & & = & = & \\ \hline
\textbf{CopyRandMR01} & \textbf{3} & & = & & = & = & = & = & = & = & & = & = & & & = & & = & = & & = & = & = & = & & = & = & = \\ \hline
\textbf{XoverBestCP1MR01} & \textbf{4} & & = & = & & = & = & = & = & = & & & = & = & = & = & = & = & = & & & = & = & = & = & = & = & = \\ \hline
\textbf{XoverBestCP05MR01} & \textbf{5} & & = & = & = & & = & = & = & = & & & & = & = & = & = & = & = & & & & = & = & = & = & = & = \\ \hline
\textbf{XoverBestCP02MR01} & \textbf{6} & & = & = & = & = & & = & = & & & & = & = & = & = & = & = & & & & = & = & = & = & = & & = \\ \hline
\textbf{XoverRandCP1MR01} & \textbf{7} & & = & = & = & = & = & & = & = & & = & = & & = & = & = & = & = & & = & = & = & = & = & = & = & = \\ \hline
\textbf{XoverRandCP05MR01} & \textbf{8} & & = & = & = & = & = & = & & = & & & = & & = & = & = & = & = & & = & = & = & = & = & = & = & = \\ \hline
\textbf{XoverRandCP02MR01} & \textbf{9} & & = & = & = & = & & = & = & & & = & = & & & = & = & = & = & & = & = & = & = & & = & = & = \\ \hline
\textbf{HillClimbingMR001} & \textbf{10} & = & & & & & & & & & & & & & & & & & & = & = & & & & & & & \\ \hline
\textbf{CopyBestMR001} & \textbf{11} & & & = & & & & = & & = & & & = & & & & & & = & & = & = & & & & & = & \\ \hline
\textbf{CopyRandMR001} & \textbf{12} & & = & = & = & & = & = & = & = & & = & & & & & & & = & & = & = & & = & & = & = & \\ \hline
\textbf{XoverBestCP1MR001} & \textbf{13} & & & & = & = & = & & & & & & & & = & & = & & & & & & = & = & = & & & = \\ \hline
\textbf{XoverBestCP05MR001} & \textbf{14} & & & & = & = & = & = & = & & & & & = & & = & = & = & & & & & = & = & = & = & & = \\ \hline
\textbf{XoverBestCP02MR001} & \textbf{15} & & = & = & = & = & = & = & = & = & & & & & = & & = & = & & & & & = & = & = & = & & = \\ \hline
\textbf{XoverRandCP1MR001} & \textbf{16} & & = & & = & = & = & = & = & = & & & & = & = & = & & = & = & & & & = & = & = & = & = & = \\ \hline
\textbf{XoverRandCP05MR001} & \textbf{17} & & = & = & = & = & = & = & = & = & & & & & = & = & = & & = & & & & = & = & = & = & = & = \\ \hline
\textbf{XoverRandCP02MR001} & \textbf{18} & & = & = & = & = & & = & = & = & & = & = & & & & = & = & & & = & = & & = & & = & = & = \\ \hline
\textbf{HillClimbingMR0001} & \textbf{19} & = & & & & & & & & & = & & & & & & & & & & = & & & & & & & \\ \hline
\textbf{CopyBestMR0001} & \textbf{20} & = & = & = & & & & = & = & = & = & = & = & & & & & & = & = & & = & & & & & = & \\ \hline
\textbf{CopyRandMR0001} & \textbf{21} & & = & = & = & & = & = & = & = & & = & = & & & & & & = & & = & & & = & & = & = & \\ \hline
\textbf{XoverBestCP1MR0001} & \textbf{22} & & = & = & = & = & = & = & = & = & & & & = & = & = & = & = & & & & & & = & = & = & = & = \\ \hline
\textbf{XoverBestCP05MR0001} & \textbf{23} & & = & = & = & = & = & = & = & = & & & = & = & = & = & = & = & = & & & = & = & & = & = & = & = \\ \hline
\textbf{XoverBestCP02MR0001} & \textbf{24} & & & & = & = & = & = & = & & & & & = & = & = & = & = & & & & & = & = & & = & & = \\ \hline
\textbf{XoverRandCP1MR0001} & \textbf{25} & & = & = & = & = & = & = & = & = & & & = & & = & = & = & = & = & & & = & = & = & = & & = & = \\ \hline
\textbf{XoverRandCP05MR0001} & \textbf{26} & & = & = & = & = & & = & = & = & & = & = & & & & = & = & = & & = & = & = & = & & = & & = \\ \hline
\textbf{XoverRandCP02MR0001} & \textbf{27} & & & = & = & = & = & = & = & = & & & & = & = & = & = & = & = & & & & = & = & = & = & = & \\ \hline
\end{tabular}
}
\end{table}

%%------------------- LOCATION C Human Presence detection --------

\begin{table}[ht!]
\caption{Wilcoxon Rank-sum test ($\alpha = 0.05$) on the pairwise comparisons between the algorithm versions tested on the distributed model (room C, human presence task). ``='' indicates that the null-hypothesis $H_0$ (statistical equivalence) is accepted (omitted on the diagonal cells). Empty cells indicate that the null-hypothesis is rejected.}
\label{tab:wilcoxonLocCHO}
\resizebox{\textwidth}{!}{
\begin{tabular}{|l|P{0.4cm}|P{0.4cm}|P{0.4cm}|P{0.4cm}|P{0.4cm}|P{0.4cm}|P{0.4cm}|P{0.4cm}|P{0.4cm}|P{0.4cm}|P{0.4cm}|P{0.4cm}|P{0.4cm}|P{0.4cm}|P{0.4cm}|P{0.4cm}|P{0.4cm}|P{0.4cm}|P{0.4cm}|P{0.4cm}|P{0.4cm}|P{0.4cm}|P{0.4cm}|P{0.4cm}|P{0.4cm}|P{0.4cm}|P{0.4cm}|P{0.4cm}|}
\hline
 & & \textbf{1} & \textbf{2} & \textbf{3} & \textbf{4} & \textbf{5} & \textbf{6} & \textbf{7} & \textbf{8} & \textbf{9} & \textbf{10} & \textbf{11} & \textbf{12} & \textbf{13} & \textbf{14} & \textbf{15} & \textbf{16} & \textbf{17} & \textbf{18} & \textbf{19} & \textbf{20} & \textbf{21} & \textbf{22} & \textbf{23} & \textbf{24} & \textbf{25} & \textbf{26} & \textbf{27} \\ \hline
\textbf{HillClimbingMR01} & \textbf{1} & & = & = & & & & & & & & = & = & & & & & & & & & = & & & & & & \\ \hline
\textbf{CopyBestMR01} & \textbf{2} & = & & = & & & & & & & & = & = & & & & & & & & = & = & & & & & & \\ \hline
\textbf{CopyRandMR01} & \textbf{3} & = & = & & & & & & & & & = & = & = & = & & & & & & & = & = & = & = & & & \\ \hline
\textbf{XoverBestCP1MR01} & \textbf{4} & & & & & = & = & & = & = & & & & & = & = & & & & & & & = & & & = & & = \\ \hline
\textbf{XoverBestCP05MR01} & \textbf{5} & & & & = & & = & & = & = & & & & & & = & = & = & = & & & & & & & = & & = \\ \hline
\textbf{XoverBestCP02MR01} & \textbf{6} & & & & = & = & & & = & = & & & & & & = & = & & = & & & & & & & = & & = \\ \hline
\textbf{XoverRandCP1MR01} & \textbf{7} & & & & & & & & & = & & & & & & & = & = & = & & & & & & & = & = & = \\ \hline
\textbf{XoverRandCP05MR01} & \textbf{8} & & & & = & = & = & & & = & & & & & & = & = & = & = & & & & & & & = & & = \\ \hline
\textbf{XoverRandCP02MR01} & \textbf{9} & & & & = & = & = & = & = & & & & & & & = & = & = & = & & & & & & & = & = & = \\ \hline
\textbf{HillClimbingMR001} & \textbf{10} & & & & & & & & & & & & & & & & & & & = & & & & & & & & \\ \hline
\textbf{CopyBestMR001} & \textbf{11} & = & = & = & & & & & & & & & = & & & & & & & & = & = & & & & & & \\ \hline
\textbf{CopyRandMR001} & \textbf{12} & = & = & = & & & & & & & & = & & & & & & & & & = & = & & & & & & \\ \hline
\textbf{XoverBestCP1MR001} & \textbf{13} & & & = & & & & & & & & & & & = & = & & & & & & & = & = & = & & & \\ \hline
\textbf{XoverBestCP05MR001} & \textbf{14} & & & = & = & & & & & & & & & = & & = & & & & & & & = & & = & & & \\ \hline
\textbf{XoverBestCP02MR001} & \textbf{15} & & & & = & = & = & & = & = & & & & = & = & & = & & = & & & & & & & = & & = \\ \hline
\textbf{XoverRandCP1MR001} & \textbf{16} & & & & & = & = & = & = & = & & & & & & = & & = & = & & & & & & & = & = & = \\ \hline
\textbf{XoverRandCP05MR001} & \textbf{17} & & & & & = & & = & = & = & & & & & & & = & & = & & & & & & & = & = & = \\ \hline
\textbf{XoverRandCP02MR001} & \textbf{18} & & & & & = & = & = & = & = & & & & & & = & = & = & & & & & & & & = & = & = \\ \hline
\textbf{HillClimbingMR0001} & \textbf{19} & & & & & & & & & & = & & & & & & & & & & = & & & & & & & \\ \hline
\textbf{CopyBestMR0001} & \textbf{20} & & = & & & & & & & & & = & = & & & & & & & = & & & & & & & & \\ \hline
\textbf{CopyRandMR0001} & \textbf{21} & = & = & = & & & & & & & & = & = & & & & & & & & & & & & & & & \\ \hline
\textbf{XoverBestCP1MR0001} & \textbf{22} & & & = & = & & & & & & & & & = & = & & & & & & & & & = & = & & & \\ \hline
\textbf{XoverBestCP05MR0001} & \textbf{23} & & & = & & & & & & & & & & = & & & & & & & & & = & & = & & & \\ \hline
\textbf{XoverBestCP02MR0001} & \textbf{24} & & & = & & & & & & & & & & = & = & & & & & & & & = & = & & & & \\ \hline
\textbf{XoverRandCP1MR0001} & \textbf{25} & & & & = & = & = & = & = & = & & & & & & = & = & = & = & & & & & & & & = & = \\ \hline
\textbf{XoverRandCP05MR0001} & \textbf{26} & & & & & & & = & & = & & & & & & & = & = & = & & & & & & & = & & = \\ \hline
\textbf{XoverRandCP02MR0001} & \textbf{27} & & & & = & = & = & = & = & = & & & & & & = & = & = & = & & & & & & & = & = & \\ \hline
\end{tabular}
}
\end{table}

%%------------------- LOCATION C Human Activity detection --------

\begin{table}[ht!]
\caption{Wilcoxon Rank-sum test ($\alpha = 0.05$) on the pairwise comparisons between the algorithm versions tested on the distributed model (room C, human activity task). ``='' indicates that the null-hypothesis $H_0$ (statistical equivalence) is accepted (omitted on the diagonal cells). Empty cells indicate that the null-hypothesis is rejected.}
\label{tab:wilcoxonLocCHA}
\resizebox{\textwidth}{!}{
\begin{tabular}{|l|P{0.4cm}|P{0.4cm}|P{0.4cm}|P{0.4cm}|P{0.4cm}|P{0.4cm}|P{0.4cm}|P{0.4cm}|P{0.4cm}|P{0.4cm}|P{0.4cm}|P{0.4cm}|P{0.4cm}|P{0.4cm}|P{0.4cm}|P{0.4cm}|P{0.4cm}|P{0.4cm}|P{0.4cm}|P{0.4cm}|P{0.4cm}|P{0.4cm}|P{0.4cm}|P{0.4cm}|P{0.4cm}|P{0.4cm}|P{0.4cm}|P{0.4cm}|}
\hline
 & & \textbf{1} & \textbf{2} & \textbf{3} & \textbf{4} & \textbf{5} & \textbf{6} & \textbf{7} & \textbf{8} & \textbf{9} & \textbf{10} & \textbf{11} & \textbf{12} & \textbf{13} & \textbf{14} & \textbf{15} & \textbf{16} & \textbf{17} & \textbf{18} & \textbf{19} & \textbf{20} & \textbf{21} & \textbf{22} & \textbf{23} & \textbf{24} & \textbf{25} & \textbf{26} & \textbf{27} \\ \hline
\textbf{HillClimbingMR01} & \textbf{1} & & & & & & & & & & = & = & & & & & & & & = & = & = & & & & & & \\ \hline
\textbf{CopyBestMR01} & \textbf{2} & & & = & & & & & & & & = & = & & & & & & & & & = & & & & & & \\ \hline
\textbf{CopyRandMR01} & \textbf{3} & & = & & & & & & & & & = & = & & & & & & & & & = & & & & & & \\ \hline
\textbf{XoverBestCP1MR01} & \textbf{4} & & & & & = & = & & = & = & & & & = & = & = & = & = & = & & & & = & = & = & = & & = \\ \hline
\textbf{XoverBestCP05MR01} & \textbf{5} & & & & = & & = & = & = & = & & & & = & = & = & = & = & = & & & & = & = & = & = & = & = \\ \hline
\textbf{XoverBestCP02MR01} & \textbf{6} & & & & = & = & & & = & = & & & & = & = & = & & = & = & & & & = & = & = & = & & = \\ \hline
\textbf{XoverRandCP1MR01} & \textbf{7} & & & & & = & & & & & & & & & & & & = & & & & & & & & & = & \\ \hline
\textbf{XoverRandCP05MR01} & \textbf{8} & & & & = & = & = & & & = & & & & = & = & = & = & = & = & & & & = & = & = & = & & = \\ \hline
\textbf{XoverRandCP02MR01} & \textbf{9} & & & & = & = & = & & = & & & & & = & = & = & & = & = & & & & = & = & = & & & = \\ \hline
\textbf{HillClimbingMR001} & \textbf{10} & = & & & & & & & & & & = & & & & & & & & = & = & = & & & & & & \\ \hline
\textbf{CopyBestMR001} & \textbf{11} & = & = & = & & & & & & & = & & = & & & & & & & = & = & = & & & & & & \\ \hline
\textbf{CopyRandMR001} & \textbf{12} & & = & = & & & & & & & & = & & & & & & & & & = & = & & & & & & \\ \hline
\textbf{XoverBestCP1MR001} & \textbf{13} & & & & = & = & = & & = & = & & & & & = & = & & = & = & & & & = & = & = & & & \\ \hline
\textbf{XoverBestCP05MR001} & \textbf{14} & & & & = & = & = & & = & = & & & & = & & = & = & = & = & & & & = & = & = & = & & = \\ \hline
\textbf{XoverBestCP02MR001} & \textbf{15} & & & & = & = & = & & = & = & & & & = & = & & & = & = & & & & = & = & = & & & = \\ \hline
\textbf{XoverRandCP1MR001} & \textbf{16} & & & & = & = & & & = & & & & & & = & & & = & = & & & & = & = & = & = & = & = \\ \hline
\textbf{XoverRandCP05MR001} & \textbf{17} & & & & = & = & = & = & = & = & & & & = & = & = & = & & = & & & & = & = & = & = & = & = \\ \hline
\textbf{XoverRandCP02MR001} & \textbf{18} & & & & = & = & = & & = & = & & & & = & = & = & = & = & & & & & = & = & = & = & = & = \\ \hline
\textbf{HillClimbingMR0001} & \textbf{19} & = & & & & & & & & & = & = & & & & & & & & & = & = & & & & & & \\ \hline
\textbf{CopyBestMR0001} & \textbf{20} & = & & & & & & & & & = & = & = & & & & & & & = & & = & & & & & & \\ \hline
\textbf{CopyRandMR0001} & \textbf{21} & = & = & = & & & & & & & = & = & = & & & & & & & = & = & & & & & & & \\ \hline
\textbf{XoverBestCP1MR0001} & \textbf{22} & & & & = & = & = & & = & = & & & & = & = & = & = & = & = & & & & & = & = & = & = & = \\ \hline
\textbf{XoverBestCP05MR0001} & \textbf{23} & & & & = & = & = & & = & = & & & & = & = & = & = & = & = & & & & = & & = & = & & = \\ \hline
\textbf{XoverBestCP02MR0001} & \textbf{24} & & & & = & = & = & & = & = & & & & = & = & = & = & = & = & & & & = & = & & = & & = \\ \hline
\textbf{XoverRandCP1MR0001} & \textbf{25} & & & & = & = & = & & = & & & & & & = & & = & = & = & & & & = & = & = & & & = \\ \hline
\textbf{XoverRandCP05MR0001} & \textbf{26} & & & & & = & & = & & & & & & & & & = & = & = & & & & = & & & & & \\ \hline
\textbf{XoverRandCP02MR0001} & \textbf{27} & & & & = & = & = & & = & = & & & & & = & = & = & = & = & & & & = & = & = & = & & \\ \hline
\end{tabular}
}
\end{table}

